\documentclass[11pt]{article}
\usepackage[margin=1in]{geometry}
\usepackage[utf8]{inputenc}
\usepackage[T1]{fontenc}
\usepackage[hidelinks]{hyperref}
\usepackage{mathtools}
\usepackage{amsmath,amsthm,amssymb,amsfonts,mathrsfs}
\usepackage{dsfont,dirtytalk,adjustbox,url,nicefrac,booktabs,array,todonotes,longtable,tabu,todonotes,multirow,tabularx,caption}
\usepackage{tikz-cd}
\usepackage{graphicx}
\usepackage{epsfig}
\usepackage{comment}
\usepackage{bm}
\usepackage[english]{babel}
\usepackage[capitalise]{cleveref}
\usepackage{cite}
\def\rational#1#2{{\mathchoice{\textstyle{#1\over#2}}%
  {\scriptstyle{#1\over#2}}{\scriptscriptstyle{#1\over#2}}{#1/#2}}}
\def\half{\rational12}		% One half
\let\OLDthebibliography\thebibliography
\renewcommand\thebibliography[1]{
  \OLDthebibliography{#1}
  \setlength{\parskip}{0pt}
  \setlength{\itemsep}{0pt plus 0.3ex}
}

\newcommand{\targ}{\rho}
\newcommand{\targf}{V}
\newcommand{\dimM}{d}

\def\l{\mathrm{L}}

\newcommand{\R}{\mathbb R}

\newcommand{\M}{\mathcal M}  % base manifold
\newcommand{\X}{\mathcal{X}} % entire phase space manifold
\newcommand{\F}{\mathcal{F}}

\newcommand{\E}{\mathbb E}
\newcommand{\B}{B}
\newcommand{\bi}{B}
\renewcommand{\S}{S}

\newcommand{\A}{A}
\renewcommand{\H}{\mathcal{H}}
\newcommand{\T}{\mathcal{T}}

\newcommand{\defn}{\equiv}  
\newcommand{\dd}{d}
\newcommand{\kl}{\operatorname{KL}}
\newcommand{\ent}{\operatorname{S}}
\renewcommand{\det}{\operatorname{d e t}}

\newcommand{\tr}{\operatorname{T r}}

\newcommand{\dt}{\delta t}                 % Integrator stepsize

\newcommand{\order}{\mathcal{O}} % big O
\DeclareMathOperator*{\argmax}{arg\,max}
\DeclareMathOperator*{\argmin}{arg\,min}
\newcommand{\obj}{V} % objective function
\newcommand{\Poi}{\Pi} % Poisson tensor
\newcommand{\Hc}{K} % High dimensional Hamiltonian
\newcommand{\GG}{\mathcal{G}} % lie group
\newcommand{\Gg}{\mathfrak{g}} % lie algebra
\newcommand{\gmet}{g} % Riemannian metric
\newcommand{\mmd}{\mathrm{MMD}}  
\newcommand{\sm}{\mathrm{SM}}  
\newcommand{\ksd}{\mathrm{KSD}} 

\newcommand{\metric}[2]{\left< #1, #2 \right>} 
\def\div{\mathrm{div}}
\def\curl{\mathrm{curl}}
\def\steinop{\mathrm{S}}
\newcommand{\KL}{\mathrm{KL}}

\newcommand{\refsec}[1]{§\ref{#1}}  
\theoremstyle{plain}
\newtheorem{theorem}{Theorem}[section]

\theoremstyle{definition}
\newtheorem{definition}[theorem]{Definition}

\theoremstyle{remark}

\title{\bf Geometric Methods for Sampling, Optimisation, Inference and Adaptive Agents}

\author{Alessandro Barp$^{1,2,\dagger}$, Lancelot Da Costa$^{3,4,\dagger}$, Guilherme França$^{5,\dagger}$, Karl Friston$^{4}$, \\[.5em] Mark Girolami$^{1,2}$, Michael I. Jordan$^{5,6}$, and Grigorios A. Pavliotis$^{3}$}
\date{\small\it
    $^1$Department of Engineering, University of Cambridge, Cambridge, UK\\[.5em]
    $^2$The Alan Turing Institute, The British Library, London, UK\\[.5em]
    $^3$Department of Mathematics, Imperial College London, London%SW7 2AZ
    , UK\\[.5em]
    $^4$Wellcome Centre for Human Neuroimaging, University College London, London%WC1N 3AR
    , UK\\[.5em]
    $^5$Computer Science Division, University of California, Berkeley, USA\\[.5em]
    $^6$Department of Statistics, University of California, Berkeley, USA\\[.5em]
    $^\dagger$\textnormal{Equal contribution}
}

\begin{document}

\maketitle

\begin{abstract}
In this chapter, we identify fundamental geometric structures that underlie the problems of sampling, optimisation, inference and adaptive decision-making. Based on this identification, we derive algorithms that exploit these geometric structures to solve these problems efficiently. We show that a wide range of geometric theories emerge naturally in these fields, ranging from measure-preserving processes, information divergences, Poisson geometry, and geometric integration. Specifically, we explain how \emph{(i)} leveraging the symplectic geometry of Hamiltonian systems enable us
to construct (accelerated) sampling and optimisation methods, \emph{(ii)} the theory of Hilbertian subspaces and Stein operators provides a general methodology to obtain robust estimators, \emph{(iii)} preserving the information geometry of decision-making yields adaptive agents that perform active inference.
Throughout, we emphasise the rich connections between these fields; e.g., inference draws on sampling and optimisation, and adaptive decision-making assesses decisions by inferring their counterfactual consequences. Our exposition provides a conceptual overview of underlying ideas, rather than a technical discussion, which can be found in the references herein.
\end{abstract}

\vspace{2em}


\thispagestyle{empty}

\newpage

\tableofcontents

% Abstract 250 words max
% Approx 12.200 words in length
% Approx 430 per published page
% So 28.3 pages
% Deadline 1 Jan 2022

\section{Introduction}

Differential geometry plays a fundamental role in applied mathematics, statistics, and computer science, 
including numerical 
integration~\cite{mclachlan2002splitting,HairerBook,leimkuhler2004simulating,Celledoni:2014,marsden2001discrete}, 
optimisation~\cite{betancourt2018symplectic,bravetti2019optimization,Franca:2020,Franca:2021,FrancaBarp:2021,Alimisis21},
sampling~\cite{rousset2010free,Duane:1987,betancourt2017geometric,livingstone2019geometric,barp2019hamiltonian},
%,hartmann2005constrained,Kennedy:2012},
statistics on spaces with %intrinsic symmetries~\cite{pennec2004probabilities,pennec1998uniform,pennec2006intrinsic}, 
deep learning~\cite{Celledoni:2021,bronstein2021geometric}, 
medical imaging and shape  methods~\cite{vaillant2005surface, durrleman2009statistical},
interpolation~\cite{barpRiemannSteinKernelMethod2020}, and the study of random maps~\cite{harms2020geometry}, to name a few.
Of particular relevance to this chapter is information geometry, i.e., the differential geometric 
treatment of smooth statistical manifolds, 
whose origin stems from a seminal article by Rao~\cite{rao1992information} who introduced the Fisher metric tensor on parametrised statistical models, and thus a natural Riemannian geometry that was later observed to correspond to an infinitesimal distance with respect to the Kullback–Leibler (KL) divergence~\cite{jeffreys1946invariant}.
The geometric study of statistical models has had many successes~\cite{amari2016information,ay2017information,nielsen2020elementary}, ranging from statistical inference, where it was used to prove the optimality of the maximum likelihood estimator~\cite{amari2012differential}, to the construction of the category of mathematical statistics, generated by Markov morphisms~\cite{chentsov1965categories,jost2021probabilistic}.
Our goal in this chapter is to discuss the emergence of natural geometries within a few 
important areas of statistics and applied mathematics, namely \emph{optimisation}, \emph{sampling}, \emph{inference}, and
\emph{adaptive agents}. % which may be used to guide practitioner's methodologies.
We provide a conceptual introduction to the underlying ideas
%the algorithms, 
rather than a technical discussion,
%which can be found in the references provided.
highlighting connections with various fields of mathematics and physics.

%\paragraph{Accelerated optimisation}
The vast majority of statistics and machine learning applications involve 
solving \emph{optimisation} problems.
Accelerated gradient-based  methods~\cite{Polyak:1964,Nesterov:1983}, and 
several variations thereof, have became work\-hor\-ses in these fields. 
Recently, there has been  great
interest in studying such  methods from a continuous-time limiting perspective;
see, e.g., ~\cite{Candes:2016,Wibisono:2016,Wilson:2021,Franca:2018a,Franca:2018b,Franca:2021b,
muehlebach2021optimization,muehlebach2021constrained} and references therein. Such  methods
can  be seen as 1st order integrators to a classical \emph{Hamiltonian
system with dissipation}. This raises the question on how to discretise the system such that important properties are preserved, 
%assuming the continuum system is well-understood, i.e., 
assuming the system has fast convergence to 
critical points and desirable stability properties.  
It has been known for a long time that the class of \emph{symplectic integrators}  is the preferred choice for
simulating physical systems~\cite{Takahashi:1984,Suzuki:1990,Yoshida:1990,SanzSerna:1992,Benettin:1994,HairerBook,mclachlan2002splitting,McLachlan:2006,Forest:2006,Kennedy:2013}. These 
discretisation techniques, designed to  preserve the underlying (symplectic) geometry of Hamiltonian systems, also form the 
basis of \emph{Hamiltonian Monte Carlo} (HMC) (or hybrid Monte Carlo)
methods~\cite{Duane:1987,Neal:2011}. Originally, such a theory of geometric integration was developed
with conservative systems in mind while, in optimisation, the associated  
system is naturally a dissipative one. Nevertheless, symplectic integrators were exploited in this context~\cite{betancourt2018symplectic,Franca:2020,bravetti2019optimization}. 
More recently, it has been proved
that a generalisation of symplectic integrators to dissipative
Hamiltonian systems is indeed able to preserve rates of convergence
and stability~\cite{Franca:2021}, which are 
the main properties
of interest for optimisation. Followup work~\cite{FrancaBarp:2021} extended this approach,
%enabling the construction of optimisation methods
enabling optimisation on \emph{manifolds} and problems with \emph{constraints}.
There is also a tight connection between optimisation on the space of measures and sampling which dates back to~\cite{Otto:2001,Jordan:1998}; we will revisit these ideas in relation to dissipative Hamiltonian systems.
%Assuming minimal background,  here we will introduce the main ideas
%and emphasise the most important principles behind this new approach to
%construct optimisation methods based on dissipative Hamiltonian dynamics. 

% \paragraph{Markov Chain Monte Carlo methods} 
\emph{Sampling} methods are critical to the efficient implementation of many methodologies. Most modern samplers are based on Markov Chain Monte Carlo methods, which include slice samplers~\cite{neal2003slice,murray2010elliptical}, piecewise-deterministic Markov chains, such as bouncy particle and zig-zag samplers~\cite{davis1984piecewise,bouchard2018bouncy,vanetti2017piecewise,bierkens2019zig,bierkens2017piecewise,peters2012rejection}, Langevin algorithms~\cite{roberts1996exponential,durmus2017nonasymptotic,durmus2018efficient}, interacting particle systems~\cite{garbuno-inigoInteractingLangevinDiffusions2019} and 
the class of HMC
methods~\cite{Duane:1987,betancourt2017geometric,rousset2010free,betancourt2017conceptual,Neal:2011,barp2018geometry}.
%; we shall focus on geometric
%methods for HMC.
%will be the focus of the section.
The original HMC algorithm was introduced in physics to sample distributions on gauge groups for lattice quantum chromodynamics~\cite{Duane:1987}. 
It combined two approaches %to molecular simulations 
that emerged in previous decades, namely the Metropolis-Hastings algorithm and the  
Hamiltonian formulation of molecular dynamics~\cite{metropolis1953equation,hastings1970monte,alder1959studies}.
%As was the case in the context of accelerated optimisation, 
Modern HMC relies heavily on symplectic integrators to simulate a deterministic dynamic, responsible
for generating distant moves between samples and thus reduce their correlation, 
%by avoiding random-walk motion, 
while at the same time preserving important geometric properties.
This deterministic step is then usually combined with a corrective step (originally a Metropolis-Hastings acceptance step) to ensure preservation of the 
%the algorithm 
%samples the right 
correct target, and 
with a stochastic process,  employed to 
speed up convergence 
to the target distribution.
%
%Section 3 
We will first focus on the geometry of measure-preserving diffusions, which emerges from ideas 
formulated by Poincaré and Volterra, and form the building block of many samplers. 
In particular, we will discuss ways to ``accelerate'' sampling %convergence to the target distribution
using irreversibility and hypoellipticity.
We will then introduce HMC focusing % noting the importance of 
on its underlying Poisson geometry, the important role played by symmetries, and its connection to geometric integration. % in its implementation.

We then  %also
discuss the problem of \emph{statistical inference}, whose practical implementation usually relies upon sampling and optimisation.
Given observations from a target distribution, many estimators belong to the family of the so-called $M$ and $Z$ estimators~\cite{van2000asymptotic}, which are obtained by finding the parameters that maximises (or are zeros of) a parametrised set of functions.
These include the maximum likelihood  and minimum Hyv\"{a}rinen
score matching estimators~\cite{vapnik1999nature,hyvarinen2005estimation},
which are also particular instances of the minimum score estimators induced by scoring rules that quantify the discrepancy between a sample and a distribution~\cite{parry2012proper}.
The Monge–Kantorovich transportation problem~\cite{villani2009optimal}
 motivates another important class of estimators, namely the minimum Kantorovich and $p$-Wasserstein estimators, whose implementation use the Sinkhorn discrepancy~\cite{bassetti2006minimum,peyre2019computational,cuturi2013sinkhorn}.
 Our discussion of inference builds upon the theory of Hilbertian subspaces and, in particular, reproducing kernels.
 These inference schemes rely on the continuity of linear functionals, such as probability and Schwartz distributions, over a class of functions to geometrise the analysis of integral probability metrics which measure the worse case integration error. We shall explain how maximum mean, kernelised, and score matching discrepancies arise naturally from topological considerations.

Models of \emph{adaptive agents} are the basis of algorithmic-decision-making under uncertainty.
This is a difficult problem that spans multiple disciplines such as statistical decision theory~\cite{bergerStatisticalDecisionTheory1985}, game theory~\cite{vonneumannTheoryGamesEconomic1944}, control theory~\cite{bellmanAppliedDynamicProgramming2015}, reinforcement learning~\cite{bartoReinforcementLearningIntroduction1992}, and active inference~\cite{dacostaActiveInferenceDiscrete2020}. 
%Herein, we focus on active inference, an approach based on modelling brain function using tools from statistical physics~\cite{fristonFreeEnergyPrinciple2006,fristonActionBehaviorFreeenergy2010,fristonFreeenergyPrincipleUnified2010}.
%To illustrate the kind of optimisation problems—that yield to the sampling and inference schemes considered in preceding sections—we will consider active inference as a generic use case
%To illustrate the kind of approaches that yield to the inference schemes considered in the preceding section, we focus on active inference.
To illustrate a generic use case for the previous methodologies we consider active inference, a unifying formulation of behaviour---subsuming perception, planning and learning---as a process of inference~\cite{fristonActionBehaviorFreeenergy2010,fristonFreeenergyPrincipleUnified2010,fristonActiveInferenceEpistemic2015,dacostaActiveInferenceDiscrete2020}.
%Active inference emerged in the late 2000s as a unifying theory of brain function% based on ideas from statistical physics~\cite{fristonFreeEnergyPrinciple2006,fristonActionBehaviorFreeenergy2010,fristonFreeenergyPrincipleUnified2010}.
We describe decision-making under active inference using information geometry, revealing several special cases %that predominate
that are established notions %of optimal decision-making in
in statistics, cognitive science and engineering. We then show how preserving this information geometry in algorithms enables adaptive algorithmic decision-making,
%Active inference describes adaptive decision-making in terms of information geometry, with several special cases that predominate in statistics, cognitive science and engineering. Crucially, preserving this information geometry in algorithms leads to adaptive algorithmic decision-making.
endowing robots and artificial agents with useful capabilities, including robustness, generalisation and context-sensitivity~\cite{lanillosActiveInferenceRobotics2021,dacostaHowActiveInference2022a}.
%While engineering challenges remain, active inference is generally considered to endow robots with more human-like adaptive capabilities~\cite{lanillosActiveInferenceRobotics2021,dacostaHowActiveInference2022a}.
Active inference is an interesting use case because it has yet to be scaled---to tackle high dimensional problems---to the same extent as established approaches, such as reinforcement learning~\cite{silverMasteringGameGo2016}; however, numerical analyses generally show that active inference performs at least as well in simple environments~\cite{millidgeDeepActiveInference2020,vanderhimstDeepActiveInference2020,cullenActiveInferenceOpenAI2018,sajidActiveInferenceDemystified2021,markovicEmpiricalEvaluationActive2021a,paulActiveInferenceStochastic2021,mazzagliaContrastiveActiveInference2021}, and better in environments featuring volatility, ambiguity and context switches~\cite{sajidActiveInferenceDemystified2021,markovicEmpiricalEvaluationActive2021a}.

\begin{comment}

\subsection{Notation}

Time $\mathcal T$

Base space $\M$, with dimension $\dimM$. 

Variable living in base space is $q \in \M$

Target distribution on base space $\mathcal M$ is $\targ = e^{-\targf} \dd q$, and target function (i.e., potential) is $\targf$

Total space (e.g., bundle over $\mathcal M$) is $\mathcal X$ (which in practice may be $T\M$ or $T^*\M$)
Variable living in total space is $x \in \mathcal X$ (e.g., $x \defn (q,p)$)

Target distribution on total space $\mathcal X$:

\subsection{Technical background}

- Manifold
- (co)-Tangent bundle
- Riemannian manifold
- Riemannian metric

\textbf{Optimisation}
- vector field
- flow
- divergence of a vector field
- Hamiltonian vector field

\textbf{Sampling}
The space of probability measures

Statistical manifolds

Definition of divergence
- associated information tensor
- statistical divergence
- discrepancy

Definition of stochastic process:
- as random variable on path space
- as many random variables on state space

- Measure preserving process

- Smooth measure

- Markov process
- Generator of Markov process
- Fokker-Planck equation
-- FP equation as dynamic on the space of measures.
- Diffusion process
- Generator of diffusion process

\end{comment}

\section{Accelerated optimisation}
\label{sec:opt}

We shall be concerned with the problem of optimisation of a 
function  $\obj:\M \to \R$, i.e.,  finding a point that 
maximises $\obj(q)$, or minimises $-\obj(q)$, 
over a smooth manifold $\M$.
We will assume  this function is differentiable to 
construct  algorithms that rely on the flows of smooth vector fields guided by the derivatives of $\obj(q)$. 
%in order to
%efficiently explore $\M$, assumed to have at least one optimiser.
%(non-saddle) 
%critical point 
%of $\obj$.
%to apply the techniques of differential geometry.

Many algorithms in optimisation are given as a sequence of finite differences, represented by iterations of a mapping $\Psi_{\dt} : \M \to \M$, where $\dt>0$ is a step size. 
The analysis of such finite difference iterations is usually challenging,  
relying on painstaking algebra to obtain theoretical gua\-ran\-tees; such as  convergence to a critical point, stability,  and rates of convergence to a critical point.
Even when these algorithms are seen as discretisations of a continuum system, whose behaviour is presumably understood,
it is well-known that most discretisations break important properties of the system.
%approximation of suitable smooth flows,
%the properties such a flow are usually broken as we shift from the 
%continuous-time world, generated by smooth vector fields, to the discrete-time world of numerical trajectories, obtained by iterating a mapping.

\subsection{Principle of geometric integration}
\label{sec:principle}

Fortunately,  here comes into play one of the most fundamental ideas of \emph{geometric integration}: 
many numerical integrators are very close---exponentially in the step size---to a smooth dynamics generated by a \emph{shadow vector field} (a perturbation of the original vector
field).  This  allows us to analyse 
the discrete trajectory implemented by the algorithm
 using powerful tools from dynamical systems and differential geometry, which are a priori reserved to smooth systems.
 Crucially, while numerical integrators typically diverge 
 significantly from the dynamics they aim to simulate,
 \emph{geometric integrators}  respect the main properties of the system.
 In the context of optimisation this means respecting stability and rates of convergence. This  was first demonstrated in~\cite{Franca:2021} and further extended in~\cite{FrancaBarp:2021}; our following  discussion will be  
 based on these works.

The simplest way to construct  numerical methods to simulate the flow of a vector field $X$  arises when it is given by a sum,
$X = Y + Z$,
and the flows of  the individual vector fields $Y$ 
and $Z$ are---analytically or numerically---tractable.
In such a case, we can approximate the exact flow 
$\Phi^X_{\dt} = e^{ \dt X}$, for step size $\dt > 0$, 
by composing the individual 
flows $\Phi^{Y}_{\dt} = e^{\dt  Y}$ and $\Phi^{Z}_{\dt} = e^{\dt Z} $. The simplest composition  
is given by $\Psi^X_{\dt} \defn \Phi^Y_{\dt} \circ \Phi^Z_{\dt}$.
%, where $\dt > 0$ is a step size.
The \emph{Baker–Campbell–Hausdorff} (BCH) formula then yields
\begin{equation} \label{bch}
e^{\dt  Y} \circ e^{\dt Z} = e^{\dt \tilde{X}}, \qquad
\tilde{X} = (Y+Z) +\frac{1}{2} [Y,Z] \dt+
\dfrac{1}{12}\left( [Y,[Y,Z]] -   [Z,[Y,Z]] \right) \dt^2+ \cdots,
\end{equation}
where $[Y,Z] = Y Z - Z Y$ is the commutator between $Y$ and $Z$.
Thus, the numerical method itself can be seen as a smooth dynamical system 
with flow map $\Psi_{\dt}^X = e^{\dt \tilde{X}}$.
The goal of \emph{geometric integration} is to construct numerical methods for which
$\tilde{X}$ shares with $X$ the critical properties of interest; this is usually done
by requiring preservation of some geometric structure. 

%exactly the same properties as $X$, thererefore the 
%
%In some special cases, the entire right-hand of \eqref{bch} itself becomes a  flow $e^{\dt \tilde{X}}$ of some perturbed or \emph{shadow vector 
%field} $\tilde{X} = X + X_1 \dt + X_2 \dt^2 + \dotsm$ that shares the 
%same properties with the original vector field $X$\AB{We must state here that this is precisely the goal of geometric integrator, that $\tilde X$ share some of the properties of $X$}.
%\AB{In what sense does $\tilde X$ have the same properties? Any consistent integrator has a shadow vector field, but most integrators are not geometric so do not preserve any given property?}. 
%\textcolor{red}{This is what I meant by ``algebra'', like Poisson algebra, which makes each term in the expansion be itself a Hamiltonian or a function obeying the same algebra as $f$.}
%When this is true\AB{I am still confused about this, we always have a shadow vector field, the commutators do not need to satisfy any additional algebra, and most of the properties of $X$ will not be shared by $\tilde X$}, 
%$\tilde{X}$ have the same properties as $X$, and the numerical method
%itself can be seen as a (smooth) dynamical system associated to the vector field $\tilde{X}$. 
%Thus, 

Recall that a numerical map $\Psi_{\dt}^X$ is said to be of order 
$r\ge 1$ if $\big| \Psi_{\dt}^X - \Phi_{\dt}^X\big| = \order( \dt^{r+1})$;
we abuse notation slightly and let $|\cdot|$ denote a well-defined distance over manifolds (see~\cite{Hansen:2011} for details). Thus,  the  expansion  \eqref{bch} also shows that the error in the approximation is 
$\big| \Psi^X_{\dt } - \Phi^X_{\dt} \big| = \order(\dt^2)$, i.e., we have
an integrator of order $r = 1$.
%Importantly,  such an expansion ensures the numerical method has the same properties as the original vector field $X$, since
%it is a perturbation of it in terms of the step size (this will be clear shortly) \AB{Wait this is not true? Otherwise any integrator would have the same geometric properties as the continuous system }. 
One can also consider more elaborate
compositions, such as 
\begin{equation}\label{leap_comp}
\Psi_{\dt}^X \equiv \Phi^Y_{ \dt / 2} \circ \Phi^{Z}_{\dt} \circ \Phi^Y_{ \dt / 2},
\end{equation}
which is more accurate  since 
the first term in  \eqref{bch} cancels
out, yielding an integrator of order $r=2$.%\footnote{By looking for appropriate compositions, that cancel first terms in the BCH formula, is how higher-order methods are constructed~\cite{Yoshida:1990}. However, methods for $r > 2$ tend to be expensive numerically, with not so many benefits (if any) over methods of order $r=2$.}
\footnote{Higher-order methods
are constructed by looking for appropriate compositions that cancel first terms in the BCH formula~\cite{Yoshida:1990}. 
However, methods for $r > 2$ tend to be expensive numerically,
with not so many benefits (if any) over methods of order $r=2$.}

\subsection{Conservative flows and symplectic integrators}
\label{sec:conservative flows}

As a stepping stone, we  first discuss the construction of suitable \emph{conservative flows}, namely flows along which some function $f: \X \to\mathbb{R}$ is constant, where $\X$ is the phase space manifold of the system, i.e., the space in which the dynamics evolves.
Such flows, which  are amongst the most well-studied  due to their importance in physics,  
will enable us to obtain our desired ``rate-matching'' optimisation methods
and  will also be central in our construction of geometric samplers. 

To construct vector fields 
along the derivative of $f$  we shall need \emph{brackets}. Geometrically, these are  
morphisms $\X^* \to \X$, also known as contravariant tensors of rank $2$ in physics,
where $\X^*$ is the dual space of $\X$. Note that
on Riemannian manifolds (e.g., $\X = \mathbb{R}^n$) both spaces are isomorphic.
In Euclidean space, $x \in \X = \mathbb{R}^n$, we define such $\bi$-vector fields
in terms of a state-dependent matrix $\bi = \bi(x)$ as\footnote{We denote
by $x^i$ the $i$th component of $x$ and 
$\partial_i \equiv  \partial / \partial_{x^i}$. We also use Einstein's summation convention, i.e., repeated upper and lower indices are summed over.}
\begin{equation}  \label{Bvecdef}
X^\bi_f(x) \defn \bi^{ij}(x) \partial_i f(x) \partial_j .
\end{equation}
Any vector field that depends linearly and locally on $f$ may be written in this manner. Notice that
a decomposition  
$f=\sum_a f_a$ induces a decomposition  $X_f^\bi = \sum_a X_{f_a}^\bi$ that is 
amenable to the splitting integrators previously mentioned. 
Importantly, vector fields that preserve  $f$ %can be fully characterised: they 
correspond to \emph{bracket vector fields}
in which $\bi$ is \emph{antisymmetric}~\cite{mclachlan1999geometric}.
%, denoted by $\bi \equiv \A$.
%In other words, these vector fields have the form 
%$ X^\A_f(z) \defn \A^{ij}(z) \partial_i H \partial_j$, with 
%$\A^{ij}(z)$ an antisymmetric matrix for all $z$.
Constructing conservative flows is thus straightforward.
Unfortunately, it is a rather more challenging task to construct 
efficient discretisations that retain this property;
 most well-known procedures, namely 
discrete-gradient  and projection methods,
only give rise to  integrators
that require solving  implicit equations at every step, and they may break
other important properties of the system.

For a particular class of conservative flows, it is possible 
to construct splitting integrators 
that---exactly---preserve another function $\tilde{f}$ that remains close to $f$.
Indeed, going back to the BCH formula \eqref{bch}, we see that if we were to approximate a conservative flow of
$X^\B_f = X^\B_{f_1}+ X^\bi_{f_2}$ by composing the flows of 
$X^\B_{f_1}$ and $X^\B_{f_2}$, and, crucially, if we had a bracket for which the commutators
can be written as
\begin{equation*}
\big[X^\B_{f_1},X^\B_{f_2}\big] = X^\B_{f_3},
\end{equation*}
for some function $f_{3}$, and so on for all commutators in \eqref{bch}, then the right-hand side of the BCH formula would itself be an expansion in terms of a vector field $X^\B_{\tilde{f}}$ 
for some \emph{shadow function} 
%\begin{equation}
$\tilde{f} = f + f_{3} \dt + f_4 \dt^2 + \dotsm$.
%\end{equation}
In particular, $\tilde{f}$ would inherit all the properties of $f$, i.e., properties common to $\bi$-vector fields.
This is precisely the case for \emph{Poisson brackets}, 
written $\bi \defn \Poi$,
%\AB{$\Pi$ (and $\pi$) are standard notion for Poisson brackets, while we do use $\mathcal{P}$ for the set or probabilty measures}, 
which are antisymmetric brackets for which the Jacobi identity holds:
\begin{equation} \label{jacobi}
\big[X^\Poi_{f},X^\Poi_{g}\big] = X^{\Poi}_{\{f,g\}} , \qquad
\{f,g\} \defn \partial_{i} f \Poi^{ij} \partial_j g ,
\end{equation}
where  $\{f,g\}$ is the Poisson bracket between functions
$f$ and $g$. 
The  BCH formula then implies 
\begin{equation}\label{shadow_poisson}
\tilde{f} = (f_1 + f_2) +\frac{1}{2} \{f_1,f_2\} \dt+
\frac{1}{12} \left( \{f_1, \{f_1,f_2\} \}
+  
 \{f_2, \{f_2,f_1\} \} \right) \dt^2+ \cdots.
\end{equation}
Such an integrator can thus be seen as a Poisson system itself,
generated by the above asymptotic shadow  $\tilde{f}$, %\eqref{shadow_poisson}, 
which is exactly preserved.
%(since it corresponds to the exact flow of the conservative vector field $X_{\tilde{f}}^\Poi$), 
%which is a slight perturbation around $f = f_1+f_2$.

Poisson brackets and their dynamics are the most important
class of conservative dynamical
systems, describing many physical systems, including all of fluid and classical mechanics.
The two main classes of Poisson brackets are constant antisymmetric matrices on Euclidean space, and \emph{symplectic brackets} for which $\Poi^{ij}(x)$ is invertible at every point  $x$. Its inverse is denoted 
by $\Omega_{ij} = \big( \Poi^{-1}\big)_{ij}$ and  called
a \emph{symplectic form}. In this case, the function $f$ is called a Hamiltonian, denoted
$f = H$. The invertibility of the Poisson tensor $\Poi^{ij}$ implies that such a bracket exists only on \emph{even-dimensional spaces}. Darboux theorem then
ensures the existence of local coordinates $x \equiv (q^1,\dotsc, q^d, p_1, \dotsc, p_d)$
in which the symplectic form can be represented as $\Omega = \left( \begin{smallmatrix} 0 & I \\ -I & 0 \end{smallmatrix}\right)$.
Dynamically, this corresponds to the fact that these are 2nd order differential equations, requiring not only a position $q \in \M$ but also a momentum $p \in T^*_q\M$.\footnote{More precisely, the dynamics evolve on the cotangent bundle $\X = T^*\M$, with coordinates $x = (q, p)$; momentum $p \in T^*_q\M$ and velocity $v = dq/dt \in T_q\M$ 
are equivalent on the Riemannian manifolds 
that are used in practice. $\M$ is called the
configuration manifold with coordinates $q$.}  
Note that  if $H= p_i$ then $X_{H} = \partial / \partial_{q^i}$, and conversely if
$H=q^i$ then $X_{H} = - \partial / \partial_{p_i}$. Thus,
a change in coordinate $q^i$ is generated by its conjugate 
momentum $p_i$, and vice-versa. Thus, the only way to generate dynamics on $\M$ in this case 
is by introducing a Hamiltonian depending on both position and momentum.  
From a numerical viewpoint, the extended phase space introduces extra degrees of freedom  that allow us to incorporate ``symmetries'' in the Hamiltonian, which facilitate integration. Indeed, in practice, the Hamiltonian usually decomposes into a potential energy, associated to position and independent of momentum, and a 
kinetic energy, associated to momentum and invariant under position changes, both generating tractable flows.
Thanks to this decomposition,  we are able to construct numerical methods through splitting the vector field.
%This also introduces some sort of ``symmetry'' 
%between position and momentum that allows a decomposition of the Hamiltonian vector
%field into a contribution from a potential energy (associated to position) and a 
%kinetic energy (associated to momentum). It is because of this ``decomposition/symmetry'' that in practice we are able to construct numerical methods through splitting the vector field. % as previously discussed. 
Note also that, for symplectic brackets, the existence of a shadow Hamiltonian can be guaranteed beyond the case of splitting methods, e.g., for variational integrators---which use a discrete version of Hamilton's principle of least action---and more generally for most symplectic integrators in which the symplectic bracket is preserved up to topological considerations described by the 1st de Rham cohomology of phase space.

\subsection{Rate-matching integrators for smooth optimisation}

Having obtained a vast family of smooth dynamics and integrators that closely preserve  $f$, we can now apply these ideas to optimisation. Vector fields for which a Hamiltonian function $f=H$ \emph{dissipates} can %also be fully characterised: they can
be written as a bracket vector field $X_H^B$ for some  negative semi-definite matrix $B$~\cite{mclachlan1999geometric}. 
Let us consider a concrete example in $\X = \mathbb{R}^{2 d}$ 
in the form of a (generalised) 
\emph{conformal Hamiltonian system}~\cite{McLachlan:2001,Franca:2020,Franca:2021}. Consider thus the Hamiltonian
\begin{equation} \label{ham0}
H(q, p)  = \dfrac{1}{2} p_i \gmet^{ij}  p_j + \obj(q) ,
\end{equation}
where $\gmet_{ij}$ is  a constant \emph{symmetric positive definite} matrix
with inverse $\gmet^{ij}$.
The associated vector field is
%\begin{equation}
$X_H^{\bi} = \gmet^{ij} p_j \partial_{q^i}  - \big[ \partial_{q^i} \obj  + \gamma(t)p_i \big] \partial_{p_i}$,
%\end{equation}
with $\gamma(t) > 0$ being a ``damping coefficient.'' This
is  associated to  the \emph{negative definite matrix} 
\begin{equation} \label{Bdamp}
\bi \equiv \underbrace{ \begin{pmatrix} 0 & -I \\ I & 0 \end{pmatrix} }_{\substack{\textnormal{conservative}}} -
\underbrace{\gamma(t)  \begin{pmatrix} 0 & 0 \\ 0 & I \end{pmatrix} }_{\substack{\textnormal{dissipative}}}.
\end{equation}
The  equations of motion are
\begin{equation} \label{conformal1}
\dfrac{d q^i}{dt} = \gmet^{ij} p_j, \qquad \dfrac{dp_i}{dt}  = -\dfrac{\partial \obj}{\partial q^i} - \gamma(t) p_i ,
\end{equation}
and obey 
\begin{equation} \label{hdot}
\dfrac{d H}{dt} = - \gamma(t) p_i g^{ij} p_j  \le 0 ,
\end{equation}
so  the system is \emph{dissipative}. Suppose $\obj(q)$ has a minimizer 
$q^\star \equiv \argmin_q \obj(q)$ in some region of interest 
and, without loss of generality, has value
$\obj^\star \equiv \obj(q^\star) \equiv 0$. 
Then $H > 0$ and $d H / dt  < 0$ outside such a critical point, implying that  $H$ is also
a (strict) \emph{Lyapunov function}; the existence of such a Lyapunov
function
implies that trajectories starting in the neighborhood of $q^\star$
 will converge to $q^\star$.  In other words, the above
system provably solves the optimisation problem 
\begin{equation}  \label{opt1}
\min_{q \in \mathbb{R}^d} \obj(q).
\end{equation}

Two common choices for the damping  are
the constant case, $\gamma(t)  = \gamma$, and the asymptotic vanishing
case, $\gamma(t) = r/t$ for some constant $r \ge 3$ (other choices
are also possible). 
When $\obj(q)$ is a convex function (resp. strongly convex function with parameter $\mu > 0$) 
it is possible to show the following convergence rates~\cite{Franca:2018b}:
%\begin{equation} \label{rate1}
%\obj(q(t)) - \obj^\star = \begin{cases}
%\order(1/t) & \mbox{for $\gamma$} , \\
%\order(\lambda_1^2(g)/t^2) & \mbox{for $\gamma(t) = r/t$},
%\end{cases}
%\end{equation}
%whereas if  $\obj(q)$ is \emph{strongly convex} function with parameter $\mu > 0$ then~\cite{Franca:2018b}
%\begin{equation} \label{rate2}
%\obj(q(t)) - \obj^\star = \begin{cases}
%\order\big(e^{-\sqrt{\mu / \lambda_1^2(g)} t }\big) & \mbox{for $\gamma= (3/2) \sqrt{\mu / \lambda_1^2(g)} $} , \\
%\order(1/t^{2r/3}) & \mbox{for $\gamma(t) = r/t$},
%\end{cases}
%\end{equation}
%
\begin{equation} \label{rates}
\def\arraystretch{1.5}
\begin{tabular}{l|l l l}
\toprule[1pt] %\hline\hline
& \emph{convex} & $\mu$-\emph{strongly convex} & \emph{damping} \\
\midrule[0.5pt] %\hline
\multirow{2}{*}{$\obj(q(t)) - \obj^\star$} & $\order\big(t^{-1}\big)$ & $\order\big( \exp\big\{-\sqrt{ \mu/\lambda_1^2(\gmet) } t \big\} \big)$ & $\gamma(t)=\mbox{const.}$ \\
& $\order\big(\lambda_1^2(\gmet) t^{-2} \big)$ & $\order\big(t^{-2r/3}\big)$ & $\gamma(t) = r/t$ \\
\bottomrule[1pt] %\hline \hline
\end{tabular}
\end{equation}
where $\lambda_1(\gmet)$ is the largest eigenvalue of the metric $\gmet$.
The \emph{convergence rates} of this system are therefore known under such convexity assumptions. Ideally, we want to design optimisation methods that preserve
these rates, i.e., are ``rate-matching'', 
and are also numerically stable. 
As we will see, such geometric integrators can be constructed by leveraging the shadow Hamiltonian property of symplectic methods on higher-dimensional conservative Hamiltonian systems~\cite{Franca:2021} (see also~\cite{marthinsen2014geometric,asorey1983generalized}).
%geometric integrators as discussed along the previous lines 
%are able to achieve this as  a consequence of having
%a \emph{shadow Hamiltonian}~\cite{Franca:2021}. 
This  holds not
only on $\mathbb{R}^{2d}$ but on general settings, namely on 
arbitrary smooth manifolds~\cite{Franca:2021,FrancaBarp:2021}.
%upon which a Hamiltonian
%dynamics can be defined.

In the conformal Hamiltonian case,
the dissipation appears explicitly in the equations
of motion. It is however theoretically convenient to consider an 
equivalent 
\emph{explicit time-dependent} Hamiltonian formulation. 
Consider the following 
coordinate transformation into system \eqref{conformal1}:
\begin{equation} \label{transf}
p \mapsto e^{-\eta(t)} p, \qquad H(q,p) \mapsto e^{\eta(t)} H\big(q, e^{-\eta(t)}p \big), \qquad \eta(t) \defn \int \gamma(t) dt.
\end{equation}
It is easy to see that \eqref{conformal1} is equivalent to standard
Hamilton's equations,
\begin{equation*}
\dfrac{d {q}^i}{dt} = \dfrac{  \partial H}{\partial p_i}, \qquad \dfrac{d{p}_i}{dt} = - \dfrac{\partial  H}{\partial q^i},
\end{equation*}
with the explicit time-dependent Hamiltonian 
\begin{equation} \label{timeham}
H(t, q, p) = \dfrac{1}{2}e^{-\eta(t)} p_i \gmet^{ij} p_j  + e^{\eta(t)} \obj(q).
\end{equation}
The rate of change of $H$ along the flow now satisfies 
\begin{equation} \label{hdot2}
\dfrac{d H}{dt} = \dfrac{ \partial H}{ \partial t} \ne 0 ,
\end{equation}
so  the system is \emph{nonconservative}; this equation is equivalent to \eqref{hdot}.

Going one step further, let us now promote $t$ to a new coordinate and
introduce its (conjugate) momentum $u$. 
Consider thus the  higher-dimensional Hamiltonian
\begin{equation} \label{ham3}
\Hc(t, q, u, p) \defn \dfrac{1}{2} e^{-\eta(t)} p_i g^{ij} p_j  + e^{\eta(t)} \obj(q) + u.
\end{equation}
Note that $t$ and $u$ are  two  arbitrary canonical coordinates. 
Denoting the time parameter of this system by $s$, 
Hamilton's equations  read 
\begin{equation}
\label{hameqhigh}
%\begin{aligned}
\dfrac{d t}{ds} = %\dfrac{\partial  \Hc}{\partial u} = 
1,   \qquad 
\dfrac{d u}{ds} = -\dfrac{\partial  \Hc}{\partial t}, \qquad
\dfrac{d q^i}{ds} = %\dfrac{\partial  \Hc}{\partial p_i} = 
e^{-\eta(t) } \gmet^{ij} p_j,  \qquad
\dfrac{d p_i}{ds} = %-\dfrac{\partial  \Hc}{\partial x^i} = 
- e^{\eta(t) }\dfrac{\partial \obj}{\partial q^i}.
%\end{aligned}
\end{equation}
%and are induced by the vector field 
%\begin{equation}
%    X_{ \Hc } = \partial_t -  \dfrac{\partial  \Hc}{\partial t} \partial_{u}+e^{-\eta(t) } g^{ij} p_j \partial_{q^i} - e^{-\eta(t) }  \dfrac{\partial  \obj}{\partial q^i} \partial_{p_i} .
%\end{equation}
This system is \emph{conservative} since
%\begin{equation}
$ d \Hc /   ds  = 0$.
%\end{equation}
Now, if we fix  coordinates  as
\begin{equation} \label{gauge1}
t = s, \qquad 
u(s) = - H(s, q(s), p(s)),
\end{equation}
the conservative system \eqref{hameqhigh}  reduces  precisely 
the original dissipative system \eqref{timeham};  
the 2nd equation in
\eqref{hameqhigh} reproduces \eqref{hdot2}, and the remaining
equations are equivalent to the equations of motion associated to 
\eqref{timeham},  
which in turn are equivalent to \eqref{conformal1} as previously noted.
Formally, what we have done is to embed the original dissipative system
with phase space $\mathbb{R}^{2d}$ into a higher-dimensional conservative
system with phase space $\mathbb{R}^{2d+2}$. The dissipative dynamics
thus lies on a hypersurface of constant energy,
$\Hc   = 0$, in  high dimensions; see ~\cite{Franca:2021} for details.  
The reason for doing this procedure, \emph{called symplectification}, is purely theoretical: since the theory of symplectic integrators only accounts
for conservative systems, we can now extend this theory to dissipative
settings by applying a symplectic integrator to \eqref{timeham} and then fixing
the relevant coordinates \eqref{gauge1} in the resulting  method. Geometrically, this corresponds to integrating the time flow exactly~\cite{Franca:2021,marthinsen2014geometric}. In~\cite{Franca:2021} such a procedure was defined under the name of \emph{presymplectic integrators}, and these connections hold not only for the specific example above but also for general non-conservative Hamiltonian systems.

We are now  ready to explain why this approach
is suitable to construct practical optimisation methods. 
Let $\Psi_{\delta s} : \mathbb{R}^{2d + 2} \to \mathbb{R}^{2d + 2}$ be
a symplectic integrator of order $r \ge 1$ applied to
system \eqref{ham3}. Denote by %\footnote{A numerical method is said to be
%of order $r \ge 1$ if the global error in approximating the
%continuum state is of $\order(\delta s^r)$; i.e., if $z(s)$ 
%is the state of the system then $|z_k - z(s_k)| \le C \dt^r$, where
%$z_k$ is obtained with $k$ iterations of the mapping $\Psi_{\delta s}$ starting from a given initial condition $z_0 \equiv z(0)$.}
$(t_k, q_k, u_k, p_k)$ the numerical state, obtained by $k = 0,1,\dotsc$ iterations of $\Psi_{\delta s}$. Time is simulated over
the grid $s_k = (\delta s) k$, with step size $\dt > 0 $.
Because a symplectic integrator has a shadow Hamiltonian we have
\begin{equation*}
\tilde{\Hc}(t_k, q_k, u_k, p_k) = \Hc(t(s_k), q(s_k), u(s_k), p(s_k)) + \order\big(\delta s^r\big).
\end{equation*}
Enforcing  \eqref{gauge1}, the coordinate 
$t_k$ becomes simply the time  discretization 
$s_k$, which is exact, and  so is $u_k = u(t_k)$ since it is a function
of time alone; importantly, $u$ does not couple to
any of the other degrees of freedom so it is irrelevant 
whether we have access to $u(s)$ or not.
Replacing \eqref{ham3} into the above equation we 
conclude:
\begin{equation}\label{shadowHtime}
\tilde{H}(t_k, q_k, p_k) =  H(t_k, q(t_k), p(t_k)) + \order(\dt^r)   ,
\end{equation}
where we now denote $t_k = (\dt ) k$, for $k=0,1,\dotsc$.
Hence, the time-dependent Hamiltonian also has a shadow, 
thanks to the cancellation of the variable $u$.
In particular, if we replace the explicit form of the Hamiltonian \eqref{timeham} we obtain\footnote{The kinetic  part only contributes to the small error since 
$g$ is positive definite and $|p_k - p(t_k)| = \order(\dt^r)$. 
There are several technical details we are omitting, such as Lipschitz conditions on the Hamiltonian and on the numerical 
method, which
we refer to~\cite{Franca:2021} for details.}
\begin{equation} \label{shadowrate}
\underbrace{\obj(q_k) - \obj^\star}_{\textnormal{numerical rate}} = \underbrace{\obj(q(t_k)) - \obj^\star}_{\textnormal{continuum rate}} + \underbrace{\order\big( e^{-\eta(t_k)} \dt^r \big)}_{\textnormal{small error}} .
\end{equation}
Therefore, the known rates \eqref{rates} for the continuum system are nearly preserved---and so would be any rates of more
general time-dependent (dissipative) Hamiltonian systems.
Moreover, as a consequence of \eqref{shadowHtime}, the original time-independent
Hamiltonian \eqref{ham0} of the conformal formulation 
is also closely preserved, i.e., within
the same \emph{bounded error} in $\delta t$---recall transformation \eqref{transf}. However, this is also a Lyapunov function, hence the numerical method respects the \emph{stability properties} of the original system as well.\footnote{Naturally, all these results  hold for suitable choices of step size, which can
be determined by a linear stability analysis of the particular
numerical method under consideration.}

In short, as a consequence of having a shadow Hamiltonian, such
geometric integrators are able to reproduce all the relevant properties
of the continuum  system. These arguments
are completely general; namely, they ultimately rely on the BCH formula, the existence of bracket vector fields and the symplectification procedure. 
%A discretisation method obtained in this way can thus be seen as a perturbed copy that shares the same properties with original  system we want to simulate\AB{this seem to repeat the sentence at the beginning of the paragraph}. 
Under these basic principles, no discrete-time analyses were necessary to obtain guarantees for the numerical method; 
which may not be particularly enlightening from a dynamical systems viewpoint
and are only applicable on a (painful) case-by-case basis.

\begin{comment}
\begin{itemize}
    \item Now want dissipation. Flows with a Liapunov are characterised: symmmetric positive definite bracket
    \item Mention  conformal vector fields, and provide rate of dissipation for these systems, explain one way to preserve dissipation is exact integration of dissipation through splitting
    \item mention algorithm 
    \item Discuss known rate of dissipation 
    \item We will instead use time-dependent Hamiltonian vector field, show transition between these and generalised conformal
    \item for such systems we can use the same methods as we did for conservative system, briefly explain this presymplectic integtator: simply do as above sec 5 \url{https://arxiv.org/pdf/1409.5058.pdf} and give formula 
    \item Algorithm like in our paper
\end{itemize}
\end{comment}

Let us now present an explict algorithm to solve
the optimisation problem \eqref{opt1}. Consider a
generic (conservative) Hamiltonian $H(q, p)$, evolving in time $s$. 
The well-known \emph{leapfrog} or St\" ormer-Verlet
method, the most used symplectic integrator
in the literature, is  based on the composition 
\eqref{leap_comp} and reads~\cite{HairerBook}
\begin{equation*}
\begin{split}
p_{k+1/2} &= p_k - (\delta s/2) \partial_{q} H(q_{k}, p_{k+1/2}), \\
q_{k+1} &= q_k - (\delta s / 2)  \left[ \partial_{p} H(q_k, p_{k+1/2}) + \partial_p H(q_{k+1}, p_{k+1/2}) \right], \\
p_{k+1} &= p_{k+1/2} - (\delta s / 2) \partial_q H(q_{k+1}, p_{k+1/2}).
\end{split}
\end{equation*}
According to our prescription,
replacing the higher-dimensional Hamiltonian \eqref{ham3}, imposing
the gauge fixing conditions \eqref{gauge1}, and recalling that $u$ cancels out, we obtain the following method:%
%%%%%%%%%%%%%
\footnote{\label{leap_practical}
In a practical implementation, it is convenient to make
the change of variables $p_{k} \mapsto e^{\eta(t_{k})} p_k$ into \eqref{dissip_leap}; recall the transformations \eqref{transf}. In this case the method reads
\begin{displaymath}
\begin{split}
p_{k+1/2} &= e^{-\Delta \eta_k }  \left[  p_k - (\delta t / 2) \partial_q \obj(q_k) \right],\\
q_{k+1} &= q_k - \delta t \cosh(\Delta \eta_k) \gmet^{-1} p_{k+1/2}, \\
p_{k+1} &= e^{-\Delta \eta_k} p_{k+1/2} - (\delta t  / 2) \partial_q \obj(q_{k+1}) ,
\end{split}
\end{displaymath}
where $\Delta \eta_k \defn \eta(t_{k + 1/2}) - \eta(t_k) = \int_{t_k}^{t_k + 1/2} \gamma(t) dt$. Note that only a half-step difference of $\eta(t)$ appears in these updates.
The algorithm is thus written in the same variables as the conformal
representation \eqref{conformal1}. The  advantage is that
we do not have large or small exponentials, which can be problematic numerically.
Furthermore, when solving optimisation problems, it is convenient to set 
the matrix $\gmet = (\delta t) I $;
this was noted in~\cite{Franca:2020} but can also be understood from the rates
\eqref{rates} since then the step size $\delta t$ disappears from 
some of these formulas.}
%%%%%%%%%%%%%%%
\begin{equation} \label{dissip_leap}
\begin{split}
p_{k+1/2} &= p_k - (\delta t/2) e^{\eta(t_k)}\partial_{q} \obj(q_k), \\
q_{k+1} &= q_k - (\delta t / 2) \big[ e^{-\eta(t_k)} + e^{-\eta(t_{k+1})} \big] \gmet^{-1}  p_{k+1/2} , \\
p_{k+1} &= p_{k+1/2} - (\delta t/2) e^{\eta(t_{k+1})}\partial_q \obj(q_{k+1}),
\end{split}
\end{equation}
where we recall that $\delta t > 0$ is the step size and $t_k = (\delta t) k $, for iterations
$k=0,1,\dotsc$.
This method, which is a dissipative generalisation of the leapfrog, 
was proposed in~\cite{Franca:2021} and has very
good performance when solving unconstrained 
problems \eqref{opt1}.  
In a similar fashion, one can extend any (known) symplectic
integrator to a dissipative setting; the above method is just one such example. 

\subsection{Manifold and constrained optimisation}
\label{sec:opt_manifold}

%Based on~\cite{FrancaBarp:2021}, we now briefly highlight that the previous approach can be extended in great generality: namely, to an optimisation problem in the form
Following~\cite{FrancaBarp:2021}, we briefly mention how the previous approach can be extended in great generality, i.e., to an optimisation problem 
\begin{equation} \label{opt_man}
\min_{q \in \M} \obj(q) ,
\end{equation}
where $\M$ is an arbitrary  Riemannian manifold.
There are essentially two ways to solve
this problem through a (dissipative) Hamiltonian approach. 
One is to is to simulate
a Hamiltonian dynamics on $T^*\M$ by incorporating the metric of $\M$ in the kinetic part of the Hamiltonian.  Another  is to consider 
a Hamiltonian dynamics on $\mathbb{R}^n$ and embed
$\M$ into $\mathbb{R}^{n}$ by imposing several constraints,\footnote{Theoretically, there is no loss of generality 
 since
Nash or Whitney embedding theorems tells us that any smooth manifold $\M$ can
be embedded into $\mathbb{R}^n$ for sufficiently large $n$.}
\begin{equation} \label{constraints}
\psi_a(q) = 0, \qquad a=1,\dotsc,m.
\end{equation}
This constrained case turns out to be particularly useful since we typically are unable to compute the geodesic flow on $\M$, but are able to construct robust constrained symplectic integrators for it.
%  (which is unfeasible in most cases)
%  or even know the metric of $\M$.
%~\AB{People like Mclachlan will tell you the Nash embedding is useless in application (it's just a theoretical result)! Also we do approximate the geodesic flow with RATTLE in the constraint case}.

As an example of the first approach, consider $\M =\GG$ being a \emph{Lie group},
with Lie algebra $\Gg$ and generators $\{T_i\}$. The analogous of
 Hamiltonian \eqref{timeham} is given by  
$H  = -\tfrac{1}{4g} e^{-\eta(t)}\tr(P^2) + e^{\eta(t)} \obj(Q)$,
where $g > 0$ is a constant, 
$Q \in \GG$ and $P \in \Gg$ (they can be seen as matrices).
The method  \eqref{dissip_leap} can be adapted to this setting, resulting
in the following algorithm~\cite{FrancaBarp:2021} (recall footnote~\ref{leap_practical}):
\begin{equation} \label{lie_algo}
\begin{split}
P_{k+1/2} &= e^{-\Delta \eta_k }  \left\{  P_k - (\delta t / 2) 
\tr\left[ \partial_{Q}\obj(Q_k) \cdot Q_k \cdot P_k \right] P_k  \right\},\\
Q_{k+1} &= Q_k \exp\left[\delta t \cosh(\Delta \eta_k) g^{-1} P_{k+1/2}\right], \\
P_{k+1} &= e^{-\Delta \eta_k} P_{k+1/2} - 
(\delta t / 2) 
\tr\left[ \partial_{Q}\obj(Q_{k+1}) \cdot Q_{k+1} \cdot P_{k+1/2} \right] P_k, 
\end{split}
\end{equation}
where $\big(\partial_Q \obj(Q)\big)_{ij} = \partial \obj / \partial Q_{ji}$ is a matrix.
%This algorithm solves optimisation problems \eqref{opt_man} over arbitrary Lie groups.

As an example of the second approach, 
one can constrain the integrator on $\R^n$ to define a symplectic integrator on $\M$ via the discrete constrained variational approach~\cite{marsden2001discrete} 
by 
%one can introduce  constraints
%through 
introducing Lagrange multipliers, i.e.,
by considering the Hamiltonian 
$ H + e^{\eta(t)} \sum_a \lambda^a \psi_a(q)$,
where $H$  is the Hamiltonian  \eqref{timeham}.
In particular, the method \eqref{dissip_leap}
can be constrained  to yield~\cite{FrancaBarp:2021}
\begin{equation} \label{dissip_rattle}
\begin{split}
p_{k+1/2} &= e^{-\Delta \eta_k } \Lambda_g(q_k) \left[  p_k - (\delta t / 2) \partial_q \obj(q_k) \right],\\
\bar{p}_{k+1/2} &= p_{k+1/2} - (\delta t / 2)e^{-\Delta \eta_k } \left[\partial_q  \psi(q_{k})\right]^{\top} \!\lambda, \\
q_{k+1} &= q_k - \delta t \cosh(\Delta \eta_k) \gmet^{-1} \bar{p}_{k+1/2}, \\
0 &= \psi_a(q_{k+1}) \qquad (a=1,\dotsc,m), \\
p_{k+1} &= \Lambda_g(q_{k+1})\left[ e^{-\Delta \eta_k} \bar{p}_{k+1/2} - (\delta t  / 2) \partial_q \obj(q_{k+1}) \right] ,
\end{split}
\end{equation}
where we have the projector
$\Lambda_g(q) \equiv I - R_g^{-1}(q) \partial_q \psi(q) g^{-1}$ with 
$R_g(q) \equiv \partial_q \psi(q) g^{-1} \partial_q\psi(q)^{\top}$, and
$(\partial_q \psi)_{ij} \equiv \partial \psi_i / \partial q^j $ is
the Jacobian matrix of the constraints;  $\lambda = (\lambda_1, \dotsc, \lambda_m)^{\top}$ is the vector of Lagrange multipliers and $\Delta \eta_k \equiv \int_{t_k}^{t_{k+1/2}} \gamma(t) dt$ accounts for the damping.
In practice, the Lagrange multipliers are determined by solving
the (nonlinear) algebraic equations for the constraints, i.e., the
2nd to 4th updates above are solved simultaneously.
The above method consists in a dissipative generalisation of the well-known
RATTLE integrator from molecular dynamics
\cite{Andersen:1983,Leimkuhler:1994,McLachlan:2014,leimkuhler2016efficient}.

It is possible to generalise any other (conservative)
symplectic method to this (dissipative) optimisation setting on manifolds.
In this %more
general setting, there still exists a shadow Hamiltonian
%for such methods,
so that convergence rates and stability are closely preserved numerically~\cite{FrancaBarp:2021} 
(similarly to \eqref{shadowHtime} and \eqref{shadowrate}).
%; see~\cite{FrancaBarp:2021} for details.
In particular, one can also consider different types of kinetic energy, beyond
the quadratic case discussed above, which may perform better in specific
problems~\cite{Franca:2020}.
This approach therefore allows one to adapt existing symplectic integrators to
solve  
optimisation problems on Lie 
groups and other %matrix 
manifolds commonly appearing in machine learning, such as Stiefel, Grassmanians, or to solve constrained optimisation problems on $\mathbb{R}^n$.

\subsection{Gradient flow as a high friction limit}
\label{sec:high_friction}

Let us provide some intuition  why simulating 2nd order systems is expected
to yield faster algorithms.
It has been shown that several other accelerated optimisation methods%
\footnote{Besides accelerated gradient based methods, accelerated extensions of important proximal-based methods such as
proximal point, proximal-gradient, alternating direction method of multipliers (ADMM),
Douglas-Rachford, Tseng splitting, etc., are implicit  discretizations of \eqref{conformal1}; see~\cite{Franca:2021b} for details.}
are also 
discretisations of system \eqref{conformal1}~\cite{Franca:2021b}.
Moreover, in the large
friction limit, $\gamma \to \infty$, this system reduces to the 1st order gradient flow,
$d q / dt = - \partial_{q} \obj(q)$ (assuming $g \propto I$), which is the continuum limit of standard, i.e., nonaccelerated 
methods ~\cite{Franca:2021b}. The same happens in more general settings; when the damping is too strong,
the second derivative becomes negligible and the dynamics is approximately 1st order. 

\begin{figure}
\centering
\includegraphics[scale=0.45,trim={0 -50 0 0}]{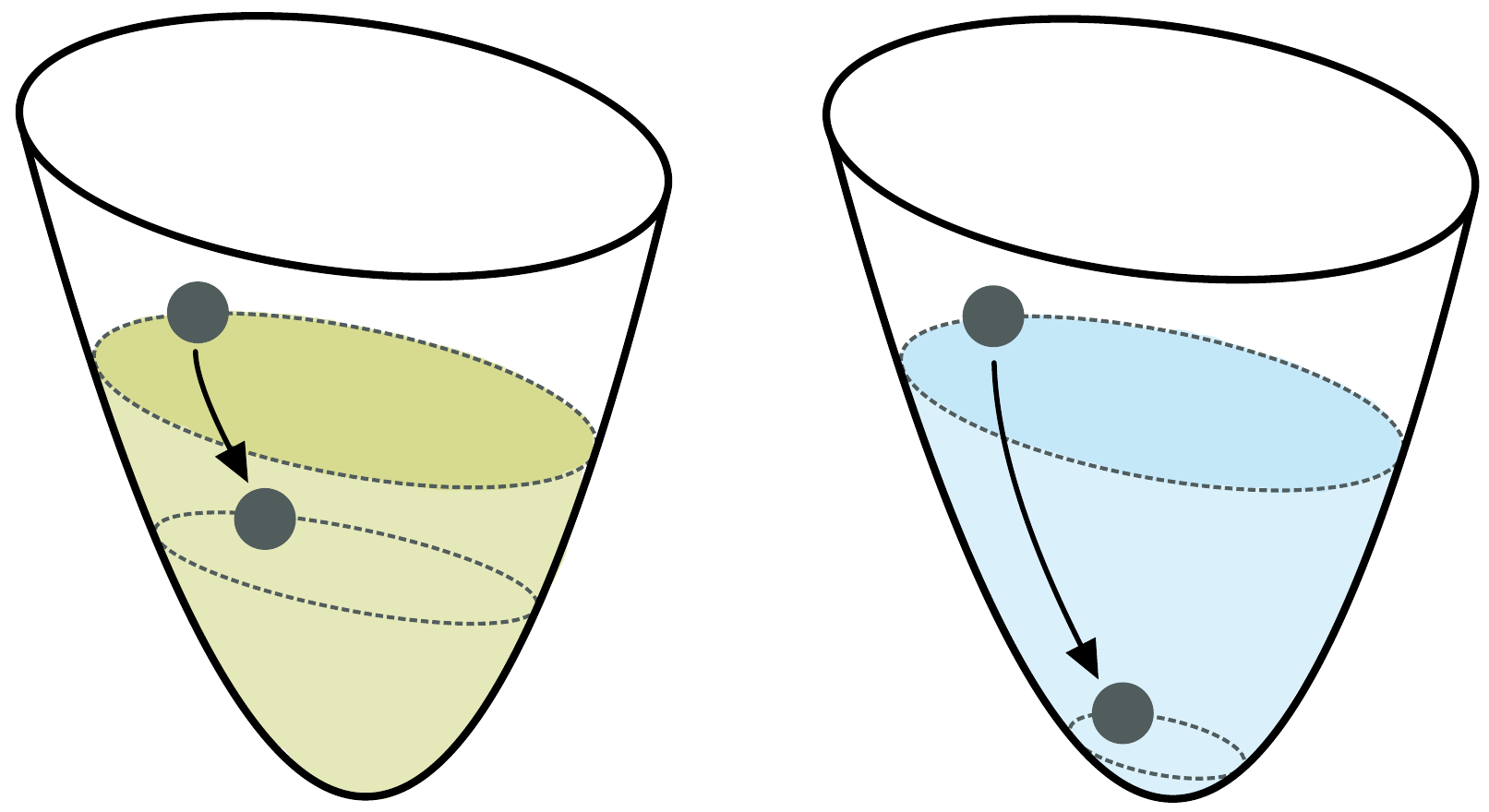}~~~~~~~%
\includegraphics[scale=0.6]{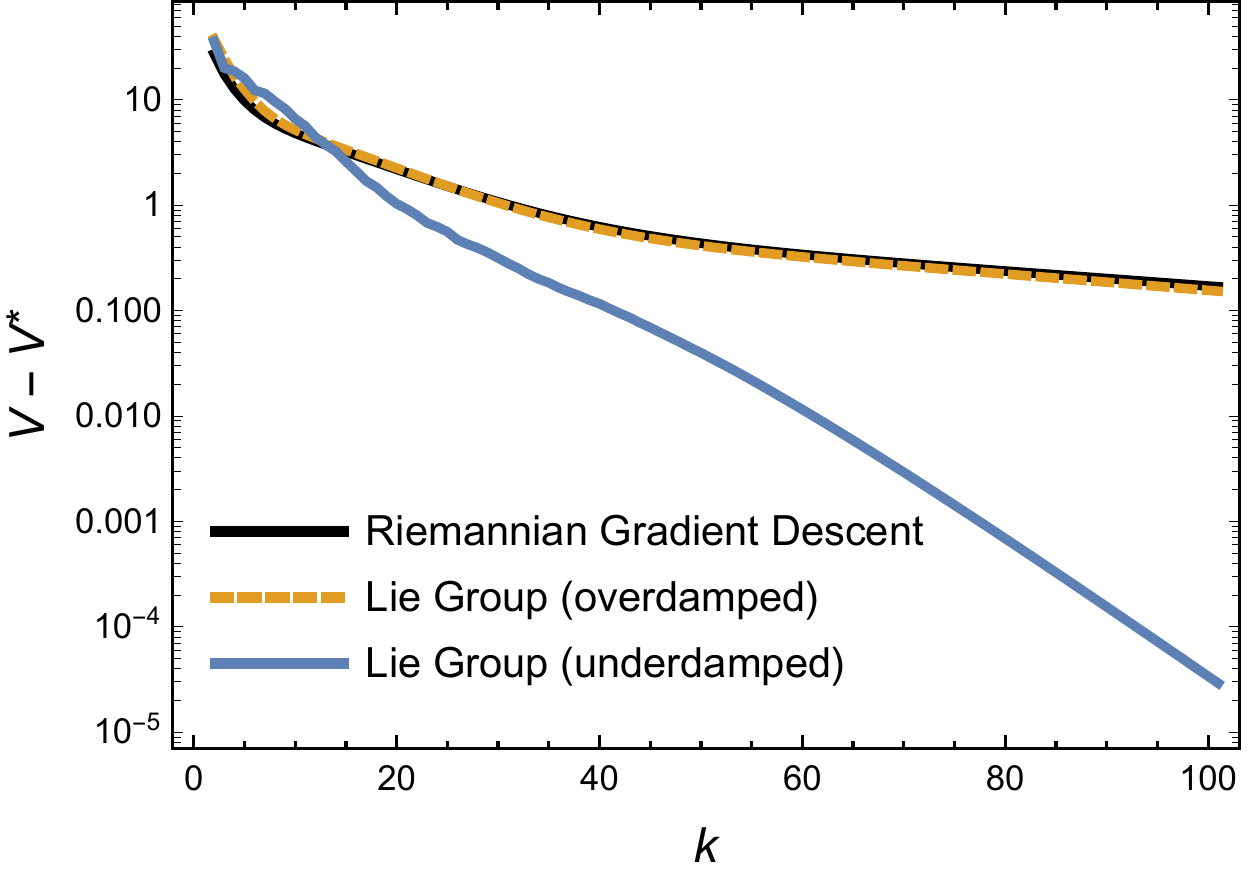}
\put(-420,140){$\gamma \to \infty$}
\put(-310,140){$\gamma = \order(1)$}
\put(-453,12){{overdamped (honey)}}
\put(-428, 0){{1st order}}
\put(-338,12){{underdamped (water)}}
\put(-312, 0){{2nd order}}
\put(-303,57){$\delta t$}
\put(-418,77){$\delta t$}
\put(-360,30){$-\partial_q \obj$}
\put(-350,50){$\Big\downarrow$}
\put(-355,95){$-\gamma p$}
\put(-350,75){$\Big\uparrow$}
\caption{\emph{Why simulating 2nd order systems yields accelerated methods.}
\emph{Left:} Constrained particle falling in fluids of different viscosity. When the drag force is strong the particle
cannot accelerate and has a 1st order dynamics (see text). \emph{Right:} Simulation of algorithm \eqref{lie_algo} where
$\obj(Q)$ is the energy of a spherical spin glass (Lie group $SO(n)$, with $n=500$)~\cite{FrancaBarp:2021}. 
In the overdamped regime the method is close to Riemannian
gradient descent~\cite{Sra:2016}, which is a 1st order dynamics;
\eqref{lie_algo} is much faster in the underdamped regime.
}
\label{fig:fig_damp}
\end{figure}

As an illustration, consider
Figure~\ref{fig:fig_damp} (left) where a particle immersed in a  fluid falls under the influence
of a potential force $-\partial_q V(q)$, that plays the role of ``gravity'', and is constrained to move on a
surface. In the \emph{underdamped} case, the particle is under water, which is not so viscous, so it 
has acceleration and  moves fast (even oscillate). In the \emph{overdamped} case, the particle is in a highly viscous
fluid, such as honey, and the drag force $-\gamma p$ is comparable
or stronger to  $-\partial_q V(q)$, thus the particle moves slowly since it cannot accelerate; 
during the same elapsed time $\dt$, an accelerated particle would travel a longer distance.
We can indeed verify this behaviour numerically. In Figure~\ref{fig:fig_damp} (right) we run algorithm \eqref{lie_algo}
in the underdamped and overdamped regimes when solving an optimisation problem
on the $n$-sphere, i.e., on the Lie group $SO(n)$.% 
\footnote{The details are not important here, but this problem minimises
the Hamiltonian of a spherical spin glass (see~\cite{FrancaBarp:2021} for details). The same behaviour is seen with the constrained method \eqref{dissip_rattle} as well.}
We can see that, in the overdamped regime, 
this method has essentially the same dynamics as the Riemannian gradient descent~\cite{Sra:2016}, which is nonaccelerated and corresponds to a 1st order dynamics; all methods use the same step size, only the damping coefficient is changed.

\subsection{Optimisation on the space of probability measures}

%\AB{Here I need to add one paragraph on information geometry and divergences, citing editors :p}
There is a tight connection between sampling and optimisation
on the space of probability measures which goes back to~\cite{Otto:2001,Jordan:1998}.
Let $\mathcal{P}_2(\mathbb{R}^n)$ be the space of probability measures
on $\mathbb{R}^n$ with finite second moments, endowed with
%and $\Pi(\mu, \nu)$ be the
%set of probability measures on $\mathbb{R}^n \times \mathbb{R}^n$ with marginals
% $\mu$ and $\nu$.  The Wasserstein-2 metric is 
%\begin{equation}\label{wass}
%W_2^2(\mu, \nu) \defn \inf_{\pi \in \Pi(\mu, \nu) } \left\{ \dfrac{1}{2} \iint | x-y |^2 %d\pi(x, y) \right\} ,
%\end{equation}
%so that  $(\mathcal{P}_2(\mathbb{R}^n), W_2 )$ is a metric space.
a Wasserstein-2 metric $W_2$.
%If $\mu : \mathbb{R} \to \mathcal{P}_2(\mathbb{R}^n)$ is a curve, parametrized by $t$,
The \emph{gradient flow} of a functional $F[\mu]$ on the space of probability measures is the solution to the partial differential equation
$\partial_t \mu(q,t) = - \nabla_{W_2} F[\mu(q,t)]$, which,
under sufficient regularity conditions, %the gradient flow $-\nabla_{W_2} F[\mu]$ on Wasserstein space can be computed in a variational way and this
is equivalent to
\cite{Otto:2001,Jordan:1998,ambrosioGradientFlowsMetric2005}
\begin{equation}\label{over_fpe}
\partial_t \mu = \partial \cdot  \left( \mu  \partial \dfrac{\delta F[\mu]}{\delta \mu} \right),
\end{equation}
where $\partial \equiv \partial_{q}$ and $\partial \,\cdot$ are the derivative and the divergence operators on $\mathbb{R}^n$, respectively.
The evolution of this system solves the optimisation problem
%minimises the functional
\begin{equation} \label{min_wass}
\rho \equiv \argmin_{\mu \in \mathcal{P}_2(\mathbb{R}^n) }  
F[\mu] ,
\end{equation}
i.e., $\mu(q,t) \to \rho(q)$ as $t \to \infty$ in the sense of distributions.
%corresponds to 
%the continuum limit ($\dt \to 0$) of
%a minimization problem on $\mathcal{P}_2(\mathbb{R}^n)$~\cite{Jordan:1998}:
%\begin{equation}
%\mu^{k+1} = \argmin_{\nu \in \mathcal{P}_2(\mathbb{R}^n)} \left\{ 
%\dt \mathcal{F}[\nu] +  \mathcal{W}_2^2(\mu^{k}, \nu)
%\right\}.
%\end{equation}
We can consider the analogous situation with a dissipative flow induced by the conformal Hamiltonian tensor \eqref{Bdamp} on the space of probability measures;  we set  $g = I$ and $\gamma(t) = \gamma = \mbox{const.}$ for simplicity. Thus, instead of
\eqref{over_fpe},
we have
a \emph{conformal Hamiltonian system in Wasserstein space} 
%given by
%\begin{equation}
%$
% \partial_t \mu(q,p,t) = - \B \nabla_{W_2}  F[\mu(q,p,t)]
%$,
%\end{equation}
given by the continuity equation
\begin{equation} \label{continuity}
\partial_t \mu = \partial \cdot \left( \mu \B \partial \dfrac{\delta F[\mu]}{\delta \mu} \right) ,
\end{equation}
where now $\partial \equiv (\partial_q, \partial_p)$ and $\mu$ is a measure
over  $\mathcal{P}_2(\mathbb{R}^{2d})$.
%and $\mu(q, p, t)$ denotes a curve on this space. 
Let $F$ be the \emph{free energy} defined as
\begin{equation}
\label{Helm_free}
F[\mu] \equiv  U[\mu] - \beta^{-1} \ent[\mu], \qquad
U[\mu] \equiv \E_\mu [ H] , \qquad 
\ent[\mu] \equiv \E_{\mu} [-\log \mu],
%F[\mu] \defn \E_{\mu} [H]  - \beta^{-1} \E_{\mu}[ -\log \mu ],
\end{equation}
where $U$ is the (internal) energy, $H$ is the Hamiltonian \eqref{ham0},
$\ent$ is the Shannon entropy, and $\beta$ is the inverse
temperature. %Explicitly,
%\begin{equation}
%U[\mu] = \E_\mu [ H] , \qquad 
%\ent[\mu] = \E_{\mu} [-\log \mu] ,
%- \int d^d q \int d^d p \, \mu  \log \mu  ,
%\end{equation}
The functional derivative of the free energy equals
\begin{equation}  \label{func_der}
\begin{split}
\dfrac{\delta F }{\delta \mu} = H + \beta^{-1} \log \mu 
=  \dfrac{1}{2} \|p\|^2  + \obj(q) + \beta^{-1} \log \mu  .
\end{split}
\end{equation}
In particular, the minimiser of ${F}$ is the stationary
density
\begin{equation} \label{gibbs}
\rho (q, p) = Z_\beta^{-1} e^{-\beta H(q,p)}, \qquad
%Z_\beta \defn \int d^d q \int d^d p \, e^{-\beta H(q,p)} ,
Z_\beta \equiv \E_\mu \big[ e^{-\beta H(q, p)} \big] .
\end{equation}
%where $\rho $ is the solution of the optimisation problem \eqref{min_wass}.
Note also that  the free energy  \eqref{Helm_free}  is nothing but the
$\KL$ divergence (up to a constant which is the partition function):
\begin{equation*}
\KL[\mu \mid \rho] \equiv \E_\mu\left[\log( \mu / \rho )  \right] = \beta F[\mu] - \log Z_\beta .
\end{equation*}
Therefore, the evolution of $\mu$ as given by the conformal Hamiltonian system 
\eqref{continuity} minimises the divergence from the  
the stationary density \eqref{gibbs}. % on $\mathcal{P}_2(\mathbb{R}^{2d})$.
%, i.e., $\mu^\star$ is the solution of problem
%\eqref{min_wass} and .
Replacing \eqref{func_der} into \eqref{continuity} we obtain
%\begin{equation} \label{fpe_general}
%\partial_t \mu = - \nabla_q \cdot \left[  \mu \nabla_p H  \right]
%+ \nabla_p \cdot \left[ \mu \nabla_q H + \gamma(t) \mu \nabla_p H  \right]
%+ \gamma(t) \beta^{-1} \nabla^2_p \mu .
%\end{equation}
%We recognize this as a generalised Fokker-Planck equation.
%For the specific case of \eqref{ham0}, with $g=I$ for simplicity, we have
\begin{equation*}
\partial_t \mu = - \partial_{q} \cdot \left[  \mu  p \right]  + \partial_{p} \cdot \left[ \mu \partial_{q} \obj(q) + \gamma \mu p  \right] + \gamma \beta^{-1} \partial^2_{p} \mu ,
\end{equation*}
which is nothing but the Fokker-Planck equation associated to the \textit{underdamped}
Langevin diffusion
\begin{equation} \label{under_lang}
dq_t = p_t dt, \qquad 
dp_t = - \partial_{q} \obj(q_t) dt - \gamma p_t dt + \sqrt{2 \gamma \beta^{-1}} d w_t ,
\end{equation}
where $w_t$ is a standard Wiener process.
%and we set $\gamma(t) \equiv \gamma$. 
Thus, the underdamped Langevin can be seen as performing \emph{accelerated} optimisation on the space of probability measures.
%as defined
%by conformal Hamiltonian systems. 
A  quantitative study of its speed of convergence is given by the theory of hypocoercivity~\cite{ottobreMarkovChainMonte2016,pavliotisStochasticProcessesApplications2014,villaniHypocoercivity2009}. 

The above results provide a tight connection between sampling and optimisation.
Interestingly, by the same argument as used in section~\ref{sec:high_friction}
(see Figure \ref{fig:fig_damp}), 
the high friction limit, $\gamma \to \infty$, of the underdamped Langevin diffusion \eqref{under_lang} yields the \textit{overdamped
Langevin diffusion}~\cite{Franca:2021b,pavliotisStochasticProcessesApplications2014}
\begin{equation}
\label{over_lang}
    dq_t = -\nabla \obj(q_t) dt + \sqrt{2 \beta^{-1}} dw_t,
\end{equation}
which corresponds precisely to the gradient flow \eqref{over_fpe} on the free energy functional $F[\mu]$~\cite{Jordan:1998,ambrosioGradientFlowsMetric2005}, where
now $\mu = \mu(q, t) \in \mathcal{P}_2(\mathbb{R}^{d})$. 
Thus, in the same manner that a 2nd order damped
Hamiltonian system may achieve accelerated optimisation compared to a 1st order gradient flow, the underdamped Langevin diffusion \eqref{under_lang}
may achieve accelerated sampling compared to the overdamped Langevin diffusion \eqref{over_lang}. Such an acceleration has indeed been  
demonstrated~\cite{Ma:2019} in continuous-time and for
a particular discretisation.

\section{Hamiltonian-based accelerated sampling}

The purpose of sampling methods is to efficiently draw samples from a given target distribution $\targ$ or, more commonly, to calculate expectations with respect to $\targ$: %, i.e., integrals of the form
% \begin{align}
% \label{eq: integral}
%     \int_{\chi} f\: \dd \targ.
% \end{align}
%  The Monte Carlo solution aims to approximate \eqref{eq: integral} with estimators of the form
% \begin{align}
% \label{eq: MCMC estimator}
%     \frac{1}{n} \sum_{k=0}^{n-1} f\left(x_{k}\right).
% \end{align}
\begin{align}
%\label{eq: integral}
\label{eq: MCMC estimator}
    \int_{\X} f\: \dd \targ \approx
   \frac{1}{n} \sum_{k=0}^{n-1} f\left(x_{k}\right).
\end{align}
However, generating   i.i.d. samples $\{x_k\}$ is usually practically infeasible, 
even for finite sample spaces, 
as in high dimensions probability mass tends to concentrate in small regions of the sample space, while regions of high probability mass tend to be separated by large regions of negligible probability.
Moreover,
$\rho$ is usually only known up to a normalisation constant~\cite{mackay2003information}.
To circumvent this issue, MCMC methods rely on constructing ergodic Markov chains $\left\{x_{n}\right\}_{n \in \mathbb{N}}$ that preserve the target distribution $\targ$.
If we run the chain long enough ($n \rightarrow \infty$), Birkhoff's ergodic theorem guarantees that the estimator on the right-hand side of \eqref{eq: MCMC estimator}
converges to our target integral on the left-hand side almost surely~\cite{hairerErgodicPropertiesMarkov2018}.
% \begin{align}
%   \lim _{n \rightarrow \infty} \frac{1}{n} \sum_{k=0}^{n-1} f\left(x_{k}\right)=\int_{\chi} f \: \dd \targ \text{ a.s.} 
% \end{align}
An efficient sampling scheme is one that minimises the variance of the MCMC estimator. %\eqref{eq: MCMC estimator}. 
In other words, fewer samples will be needed to obtain a good estimate. %to our target quantity. 
Intuitively, good samplers are Markov chains that converge as fast as possible to the target distribution. %---this leads to a faster convergence of the MCMC estimator.

\subsection{Optimising diffusion processes for sampling}

%Sampling in continuous time requires specifying an (ergodic) stochastic process that preserves the target distribution $\targ$. For Markov processes, % preserving the target measure $\targ$ 
%measure preservation is equivalent to a condition on the generator $\mathcal L$, $\int \mathcal L f \: \dd \targ =0$ for all $f$ in the domain of the generator.
%Given a target distribution, there is a wide class of Markov processes that preserve the measure; these may include diffusion processes~\cite{pavliotisStochasticProcessesApplications2014}, partially deterministic Markov processes (which feature jumps) and interacting particle systems 
%\cite{garbuno-inigoInteractingLangevinDiffusions2019}.
%\footnote{Interacting Langevin Diffusions- Gradient Structure And Ensemble Kalman Sampler- Garbuno-Inigo, Stuart et al (2019); Optimizing interacting Langevin dynamics using spectral gaps- Pavliotis}.

As many MCMC methods are based on discretising continuous-time stochastic processes, the analysis of continuous-time processes is informative of the properties of efficient samplers.

Diffusion processes possess a rich geometric theory, extending that of vector fields, and have been widely studied in the context of sampling. They are Markov processes featuring almost surely continuous sample paths (i.e., no jumps) and correspond to the solutions of stochastic differential equations (SDEs).
While a deterministic flow is given by a first order differential operator---namely, a vector field $X$ as used in \refsec{sec:opt}---diffusions require specifying a set of vector fields $X,Y_1,\ldots, Y_N$, where $X$ represents the (deterministic) drift and $Y_i$ the directions of the (random) Wiener processes $w^i_t$, and are characterised by a second order differential operator of the form 
$\mathcal{L} \defn X + Y_i \circ Y_i$, known as the generator of the process.
Equivalently, diffusions can be written as Stratonovich SDEs: $\dd x_t = X(x_t) \dd t+ Y_i(x_t) \circ \dd w_t^i$.

For a smooth positive target measure $\rho$, the \textit{complete} family of $\rho$-preserving diffusions is given by (up to a topological obstruction contribution)~\cite{barpUnifyingCanonicalDescription2021}
    \begin{equation}
    \label{eq: complete recipe manifold}
        \dd x_t = \curl_\rho(\A) \dd t+\dfrac{1}{2} \div_\rho(Y_i)Y_i \dd t + Y_i \circ \dd w^i_t,
    \end{equation}
for a choice of antisymmetric bracket $\A$. Here $\curl_\rho $ is a differential operator on multi-vector fields, generalising the divergence on vector fields $\div_\rho $ of $\rho $, and is induced via an isomorphism $\rho^{\sharp}$ defined by $\rho$ which allows to transfer the calculus of twisted differential forms to a \emph{measure-informed} calculus on multi-vector fields~\cite{barpUnifyingCanonicalDescription2021}. 
The ergodicity of \eqref{eq: complete recipe manifold} is essentially characterised by H\"ormander's \textit{hypoellipticity} condition; i.e., whether the Lie algebra of vector fields generated by $\left\{Y_{i},\left[X, Y_{i}\right] \right\}_{i=1}^N$ spans the tangent spaces at every point~\cite{hormanderHypoellipticSecondOrder1967,pavliotisStochasticProcessesApplications2014,bismut1981martingales}.
%Concretely, introducing the symmetric positive semi-definite bracket $\S$ defined as the map $f\mapsto \half Y_i(f)Y_i$ from functions to vector fields,
 On Euclidean space the above complete class of measure preserving diffusions can be given succinctly by It\^{o} SDEs~\cite{maCompleteRecipeStochastic2015}:
    \begin{equation}
    \label{eq: complete recipe euclidean}
    \begin{split}
        \dd x_{t} &=-(\A+\S)\left(x_{t}\right) \partial V\left(x_{t}\right) \dd t+\partial \cdot\left( \A+\S\right)\left(x_{t}\right) \dd t+\sqrt{2\S\left(x_{t}\right)} \dd w_{t},
    \end{split}
    \end{equation}
where $\S,\A$ reduce to symmetric and antisymmetric matrix fields and $V$ is the negative Lebesgue log-density of $\rho$.
%In order to sample, diffusions need to converge to the target measure from any arbitrary initial condition. For Markov processes, this %property 
%holds whenever the invariant measure is unique~\cite{hairerErgodicPropertiesMarkov2018},%\cite[Section 5]{hairerErgodicPropertiesMarkov2018}
%\cite[Section 10]{barpUnifyingCanonicalDescription2021}, 
%which is guaranteed under H\" ormander's hypoellipticity condition~\cite{barpUnifyingCanonicalDescription2021}; i.e., when the Lie algebra of vector fields generated by $\left\{Y_{i},\left[X, Y_{i}\right] \right\}_{i=1}^N$ spans the tangent spaces at every point~\cite{hormanderHypoellipticSecondOrder1967,pavliotisStochasticProcessesApplications2014}.

%Uniqueness of the invariant measure does not necessarily imply efficient 
%sampling so %we would like to characterise efficient samplers. 
%we need a concrete characterisation.
There are two well-studied criteria describing sampling efficiency in Markov processes: 1) the worst-case asymptotic variance of the MCMC estimator \eqref{eq: MCMC estimator} over functions in $\l^2(\targ)$, and 2) the spectral gap. The spectral gap is the lowest non-zero eigenvalue of the (negative) generator $-\mathcal L$ on $\l^2(\targ)$. When it exists, it 
is an exponential convergence rate of the density of the process to the target density~\cite{hwangAcceleratingDiffusions2005,pavliotisStochasticProcessesApplications2014,chafaiEntropiesConvexityFunctional2004}.
Together, these criteria yield confidence intervals on the non-asymptotic variance of the MCMC estimator, which determines sampling performance~\cite{joulinCurvatureConcentrationError2010}.

A fundamental criterion for efficient sampling is non-reversibility~\cite{hwangAcceleratingDiffusions2005,rey-belletIrreversibleLangevinSamplers2015,duncanVarianceReductionUsing2016}. A process is non-reversible if it is statistically distinguishable from its time-reversal when initialised at the target distribution~\cite{pavliotisStochasticProcessesApplications2014}. Measure-preserving diffusions are non-reversible precisely when $A \not\equiv 0$~\cite{haussmannTimeReversalDiffusions1986}. Intuitively, non-reversible processes backtrack less often and thus furnish more diverse samples~\cite{nealImprovingAsymptoticVariance2004}. Furthermore, non-reversibility leads to mixing, which accelerates convergence to the target measure. It is well known that removing non-reversibility worsens the spectral gap and the asymptotic variance of the MCMC estimator~\cite{rey-belletIrreversibleLangevinSamplers2015,duncanVarianceReductionUsing2016,hwangAcceleratingDiffusions2005}. In diffusions with linear coefficients, one can construct the optimal non-reversible matrix $A$ to optimise the spectral gap~\cite{lelievreOptimalNonreversibleLinear2013,wuAttainingOptimalGaussian2014}
or the asymptotic variance~\cite{duncanVarianceReductionUsing2016}. %(notice that the solutions differ).
However, there are no generic guidelines on how to optimise non-reversibility in arbitrary diffusions. This suggests a two-step strategy to construct efficient samplers: 1) optimise reversible diffusions, and 2) add a non-reversible perturbation $A \not\equiv 0$~\cite{zhangGeometryinformedIrreversiblePerturbations2021}.

%Let us briefly describe the class of reversible processes. A processes is reversible when it is statistically indistinguishable from its time-reversal when initialised at the target distribution. %Mathematically, $\rho = \rho^-$, where $\rho, \rho^-$ are the path space measures of the stationary process as time goes forward or backward, respectively, on an arbitrary time-interval. %Measure preserving Markov processes are reversible if and only if the generator is a self-adjoint\AB{should this by just symmetric? (I think self-adjoint is stronger)} operator in $\l^2(\targ)$.
%. This does not hold in the non-reversible case! (Simply compare~\cite{duncanVarianceReductionUsing2016} with~\cite{lelievreOptimalNonreversibleLinear2013}).
Diffusions on manifolds are reversible when  $\A\equiv 0$, and thus have the form $ \dd x_t = \half \div_\rho(Y_i)Y_i \dd t + Y_i \circ \dd w^i_t$,
which on Euclidean space reads
    \begin{equation}
    \label{eq: complete recipe reversible diffusion}
    \begin{split}
        \dd x_{t} &=  - \S\left(x_{t}\right) \partial V\left(x_{t}\right) \dd t+\partial \cdot  \S\left(x_{t}\right)\dd t +\sqrt{2\S\left(x_{t}\right)} \dd w_{t}.
   \end{split}
    \end{equation}
The spectral gap and the asymptotic variance of the MCMC estimator are the same optimality criteria in reversible Markov processes~\cite{miraOrderingImprovingPerformance2001}. When $\S$ is positive definite everywhere, it defines a Riemannian metric $\gmet$ on the state space. The generator is then the elliptic differential operator
    \begin{align}
    \label{eq: reversible elliptic diffusion}
        \mathcal L =  \nabla_{\gmet} + \Delta_{\gmet} ,
    \end{align}
where $\nabla_{\gmet}$ is the Riemannian gradient and $\Delta_{\gmet}$ is the Laplace-Beltrami operator, i.e., the Riemannian counterpart of the Laplace operator. Thus, reversible (elliptic) diffusions \eqref{eq: reversible elliptic diffusion} are the natural generalisation of the overdamped Langevin dynamics \eqref{over_lang} to Riemannian manifolds~\cite{girolami2011riemann}. Optimising $\S$ to improve sampling amounts to endowing the state space with a suitable Riemannian geometry that exploits the structure of the target density. For example, sampling is improved by directing noise along vector fields that preserve the target density~\cite{abdulleAcceleratedConvergenceEquilibrium2019}. When the potential $V$ is strongly convex, the optimal Riemannian geometry is given by $\gmet \equiv \partial^2 V$~\cite{helfferRemarksDecayCorrelations1998,saumardLogconcavityStrongLogconcavity2014}. Sampling can also be improved in hypoelliptic diffusions with degenerate noise (i.e., when $\S$ is not positive definite). Intuitively, the absence of noise in some directions of space leads the process to backtrack less often and thus yield more diverse samples. For instance, in the linear case, the optimal spectral gap is attained for an irreversible diffusion with degenerate noise~\cite{guillinOptimalLinearDrift2021}. However, degenerate diffusions can be very slow to start with, as the absence of noise in some directions of space make it more difficult for the process to explore the state space~\cite{guillinOptimalLinearDrift2021}.

Underdamped Langevin dynamics \eqref{under_lang} combines all the desirable properties of an efficient sampler: it is irreversible, has degenerate noise, and achieves accelerated convergence to the target density~\cite{maThereAnalogNesterov2019}.  We can optimise the reversible part of the dynamics (i.e., the friction $\gamma$) to improve the asymptotic variance of the MCMC estimator~\cite{chakOptimalFrictionMatrix2021}. Lastly, we can significantly improve underdamped Langevin dynamics by adding additional non-reversible perturbations to the drift~\cite{duncanUsingPerturbedUnderdamped2017}.

One way to obtain MCMC algorithms is to numerically integrate diffusion processes. As virtually all non-linear diffusion processes cannot be simulated exactly, we ultimately need to study the performance of discrete algorithms instead of their continuous counterparts. Alarmingly, many properties of diffusions can be lost in numerical integration. For example, numerical integration can affect ergodicity~\cite{mattinglyErgodicitySDEsApproximations2002}. An irreversible diffusion may sample more poorly than its reversible counterpart after integration~\cite{katsoulakisMeasuringIrreversibilityNumerical2014}. 
This may be because numerical discretisation can introduce, or otherwise change, the amount of non-reversibility~\cite{katsoulakisMeasuringIrreversibilityNumerical2014}. The invariant measure of the diffusion and its numerical integration may differ, a feature known as bias. We may observe very large bias even in the simplest schemes, such as  the Euler-Maruyama integration of overdamped Langevin~\cite{roberts1996exponential}. Luckily, there are schemes whose bias can be controlled by the integration step size ~\cite{mattinglyConvergenceNumericalTimeAveraging2010}; yet, this precludes using large step sizes. Alternatively, one can remove bias by supplementing the integration step with a Metropolis-Hastings corrective step; however, this makes the resulting algorithm reversible.
In conclusion, designing efficient sampling algorithms with strong theoretical 
guarantees is a non-trivial problem that needs to be addressed in its own right. 

\subsection{Hamiltonian Monte Carlo}

Constructing measure-preserving processes, in particular diffusions, is
%, as we have seen, 
relatively straightforward. 
A much more challenging task consists of constructing efficient \emph{sampling algorithms} with strong theoretical guarantees.
We now discuss an important family of well-studied methods, known as \emph{Hamiltonian Monte Carlo} (HMC), which can be implemented on any manifold, for any smooth fully supported target measure that is 
known up to a normalising constant. Some of these methods can be seen as an appropriate geometric integration of the underdamped Langevin diffusion, but it is in general simpler to view them as combining a geometrically integrated  deterministic dynamics with a simple stochastic process that ensures ergodicity.

The \emph{conservative} Hamiltonian systems previously discussed 
%in the context of accelerated optimisation 
provide a natural candidate for the deterministic dynamics. 
Indeed, given a target measure  
 $\targ \propto e^{-V}\mu_\M$,
 with $\mu_\M$ a Riemannian measure (such as the Lebesgue measure $\dd q$ on $\M = \R^d$),
 if we interpret the negative log-density $V(q)$ as a potential energy, i.e., a function depending on \emph{position} $q$,
 one can then plug in the potential within Newton's equation to obtain a deterministic proposal that is
 well-defined on any manifold, as soon as the acceleration and derivative operators have been replaced by their curved analogues
\begin{equation}\label{eq:Riemannian Newton}
    \underbrace{ m \ddot q= - \partial V(q)}_\text{flat Newton}  \quad \longrightarrow  \quad  \underbrace{ 
    \overbrace{ \frac{\nabla \dot q}{\dd t}}^{\text{acceleration}} = \overbrace{ - \nabla V(q) }^{
       \substack{\text{direction of} \\ \text{greatest decrease}}
       }}_{\text{Riemannian Newton}},
\end{equation}  
with given initial conditions for the position $q$ and \emph{velocity} $v=dq/dt$. This is a 2nd order  system which evolves in the tangent bundle,  $(q,v) \in T\M$, which is $T\M = \R^d \times \R^d$ when $\M = \R^d$.
The resulting flow is conservative since it   corresponds to a Hamiltonian system as discussed in section~\refsec{sec:conservative flows}, with Hamiltonian $H(q,v) \defn \tfrac{1}{2} \| v \|_q^2 + V(q)$, where $\| v \|_q^2$ is the Riemannian squared-norm,
which is   $v^T \gmet(q) v$ when $\M=\R^d$ and $\gmet(q)$ is the Riemannian metric; this is the manifold version of the Hamiltonian \eqref{ham0}. 
This system  preserves
the symplectic measure $\mu_\Omega(q,v) =  \det \gmet(q) \dd q \dd v$, 
and thus also
the  \emph{canonical distribution}
$\mu \propto e^{-H(q,v)}  \mu_\Omega$,
which is the product of the target distribution over position with the Gaussian measures on velocity (with covariance $\gmet$).
For instance, on $\M = \R^d$,   
$$  \mu(q,v) \propto  \rho(q) \times \mathcal N\big(0,\gmet^{-1}(q)\big)(v) \propto e^{-V(q)} \sqrt{\det \gmet(q)}\dd q \times \sqrt{\det \gmet(q)} e^{-\half v^T \gmet(q) v } \dd v.  $$ 
Moreover, the pushforward under the projection $\text{Proj}:(q,v)  \mapsto q$  is precisely the target measure:
$ \text{Proj}_{*} \mu = \targ$.
Concretely, the samples generated by moving according to Newton's law, after  ignoring their velocity component, have $\targ$ as their law.
 The main critical features and advantages in using  Hamiltonian systems arises
%, as for optimisation methods, 
from their numerical implementation.
Indeed, the flow of \eqref{eq:Riemannian Newton}
is only tractable for the simplest target measures, namely those possessing a high degree of symmetry. In order to proceed, we must devise suitable numerical approximations which,  unfortunately, not only break such symmetries but may lose key properties of the dynamics such as stationarity (typically not retained by discretisations).
However, as we saw in  section \refsec{sec:conservative flows}, 
most \emph{symplectic integrators} have a \emph{shadow Hamiltonian} and thus generate discrete trajectories that are close to the associated bona fide (shadow) Hamiltonian dynamics, that in particular preserve the shadow canonical distribution. 

Most integrators used in sampling, such as the  leapfrog, are \emph{geodesic integrators}.
These are \emph{splitting methods} (see  section \refsec{sec:principle}) obtained by
splitting the Hamiltonian $H(q,v) = H_1(q,v) + H_2(q)$,  where
$H_1(q,v) = \half \| v \|_q^2 $ and $H_2(q) = V(q)$ are 
to be treated as independent Hamiltonians in their own right.
Both of these Hamiltonians generate dynamics that might be tractable: 
the Riemannian component $H_1$, associated to the Riemannian reference measure, 
induces the geodesic flow, while
the target density component $H_2$
gives rise to a vertical gradient flow, 
wherein the velocity is shifted by the direction of maximal density change, 
i.e., $(q,v) \mapsto (q,v-\dt \nabla_q V(q))$.
The Jacobi identity and 
the BCH formula imply these integrators do possess a shadow Hamiltonian $\tilde{H}$, and reproduce its dynamics. Such a
shadow can in fact be explicitly obtained from  \eqref{shadow_poisson} by computing  iterated Poisson brackets;
e.g.,  on $\M=\R^d$ and for $H(q,v) = (1/2) v^T \gmet v+V(q)$, the 
three-stage geodesic integrator $\Phi^{H_1}_{b \dt} \circ\Phi^{H_2}_{a \dt} \circ \Phi^{H_1}_{(1-\half b ) \dt} \circ \Phi^{H_2}_{(1-2a) \dt} \circ \Phi^{H_1}_{(1-\half b) \dt} \circ \Phi^{H_2}_{a \dt} \circ \Phi^{H_1}_{b \dt} $, with parameters $a,b \in \R$,
yields~\cite{radivojevic2018multi}
$$ \tilde H(q,v) = H(q,v) + \dt^2\left[ c_{1} \partial V(q)^T \gmet^{-1} \partial V(q) +c_2 v^T \partial^2 V(q) v \right] + O(\dt ^4),$$
for some constants $c_1$ and $c_2$.
As an immediate consequence, these symplectic integrators preserve the reference symplectic measure $\mu_{\Omega}$
and can be used as a (deterministic) Markov proposal,
which when combined with the Metropolis-Hastings acceptance step that depends only on the target density, gives rise to a measure-preserving process.
Moreover, the existence of the shadow Hamiltonian ensures that the acceptance rate will remain high for distant proposals, allowing small correlations.
However, since Hamiltonian flows are conservative, they remain
stuck within energy level sets, which prevents  ergodicity. It is thus necessary to introduce  another measure-preserving process, known as the heat bath or thermostat, that explores different energy levels;
the simplest such process corresponds to sampling a velocity from a Gaussian distribution. Bringing these ingredients together, we  thus have  the following  HMC algorithm: given  $z^n=(q,v)$, compute $z^{n+1}$ according to
\begin{enumerate}
\item \emph{Heat bath:} sample a velocity according to a Gaussian, $v^\dagger \sim \mathcal N(0,\gmet^{-1}(q))$.
\item \emph{Shadow Hamiltonian dynamics:} move along the Hamiltonian flow generated by the geodesic integrator,  $z^* = \Psi_{\dt}(z^\dagger)$, where $z^\dagger = (q,v^\dagger)$.
\item \emph{Metropolis correction:} accept $z^*$ with probability $\min\big\{ 1, e^{- \Delta H } \big\}$, where $\Delta H = H(z^\star) - H(z^\dagger)$. If accepted then set $z^{n+1} = z^*$, otherwise set  $z^{n+1} = (q,-v^\dagger)$.
\end{enumerate}
The above rudimentary HMC method (originally known as Hybrid Monte Carlo) was proposed for simulations in lattice quantum chromodynamics  with $\M$ being the special unitary group, $SU(n)$, and used a Hamiltonian dynamics ingeniously constructed from the Maurer-Cartan frame   to compute the partition function of discretised gauge theories~\cite{Duane:1987}. This method has later been applied in molecular dynamics and  statistics~\cite{rousset2010free,betancourt2017conceptual,neal1992bayesian,cances2007theoretical,Neal:2011}.

While the above discussion provides a justification for the use of Hamiltonian mechanics,  a more constructive argument  from first principles can also be given.
From the family of measure-preserving dynamics, which as we have 
seen  
can be written as $\curl_\mu(\A)$
(recall \eqref{eq: complete recipe manifold}), we want to identify those suited to practical implementations (here $\mu$ could be any distribution on some space $\mathcal{F}$ having the target $\rho$ has a marginal).
Only for the simplest distributions $\mu$ we can hope to find brackets $\A$ for which the flow of $\curl_\mu(\A)$  is tractable. 
Instead, the standard approach to geometrically integrate this flow relies as before on splitting methods, which effectively decompose $\mu \propto e^{-\sum H_\ell}\mu_\mathcal{F}$ into simpler components by decomposing the reference measure from the density and taking advantage of any product structure of the density, so that $\curl_\mu(\A) = \curl_{\mu_\mathcal{F}}(\A)+\sum_\ell X^\A_{H_\ell}$.

There are  three critical properties underpinning  the success of  HMC in practice. 
The first two  are the preservation of the reference measure and the existence of a conserved shadow Hamiltonian for the numerical method. These imply that we remain close to preserving $\mu$ along the flow, and in particular leads to Jacobian-free Metropolis corrections with good acceptance rates for distant proposals (see~\cite{fang2014compressible} for examples of schemes with Jacobian corrections).
Achieving these  properties yield strong constraints on the choice of $\A$~\cite{barp2020bracket}; the shadow property is essentially exclusive to Poisson systems, for which the conservation of a common reference measure 
is equivalent to the triviality of the modular class in the first Poisson cohomology group~\cite{weinstein1997modular}.
In particular, Poisson brackets that admit such an \emph{invariant measure} have been carefully analysed and are known as \emph{unimodular}; the only unimodular Poisson bracket that can be constructed on general manifolds seems to be precisely the symplectic one.

The third critical property is the 
existence of splittings methods for which all the composing flows are either tractable or have adequate approximations, namely the geodesic integrators.
Indeed, as we have seen, the flow $\Phi^{H_2}$---induced by the potential $H_2(q) = V(q)$---is always tractable, independently of the complexity of the target density; this is possible mainly due to the extra ``symmetries'' resulting from implementing the flow on a higher-dimensional space $T\M$ rather than $\M$. 
On the other hand, one key consideration for the tractability of the the geodesic flow $\Phi^{H_1}_{\dt}$---induced by the kinetic energy $H_1(q, v)$---% 
is the choice of Riemannian metric; for most cases, it is numerically
hard to implement $\Phi^{H_1}_{\dt}$ since several implicit equations need
to be solved.  
In general,  it is desirable to use a Riemannian metric that reflects the intrinsic symmetries of the sample space,  mathematically described by a Lie group action.
Indeed, by using an \emph{invariant} Riemannian metric, one greatly simplifies the equations of motion of the 
geodesic flow, reducing the 
usual $2$nd order Euler-Langrange equations to the 1st order Euler-Arnold equations~\cite{holm1998euler,modin2010geodesics,barp2019hamiltonianB}, with tractable solutions in many cases of interest, e.g., for naturally reductive homogeneous spaces;  including $\R^d$, the space of positive definite matrices, Stiefel manifolds, Grassmannian manifolds, and many Lie groups.
In such cases, it is possible to find a Riemannian metric whose geodesic flow is known and given by the Lie group exponential~\cite{barp2019hamiltonian,holbrook2018geodesic,holbrook2016bayesian}.
For the other main class of  spaces, namely those given by constraints, if one chooses the restriction of the Euclidean metric,
then the RATTLE scheme discussed in optimisation (see section \refsec{sec:opt_manifold}) 
is a suitable symplectic integrator
\cite{lelievre2019hybrid,leimkuhler2016efficient,lelievre2020multiple,graham2019manifold,au2020manifold} (perhaps up to a reversiblity check).
Occasionally, it may be suitable to use a Riemannian metric associated to the target distribution rather than the sample space; e.g.,
when it belongs to a statistical manifold.
In that case, any choice of (information) divergence gives rise to an information tensor that may be
used  in the HMC algorithm.
Notably, this is the case in Bayesian statistics, wherein attempting to find a Riemannian metric that locally matches the Hessian of the posterior motivates the use of
the  Fisher information tensor summed with the Hessian of the prior,  giving rise to the  Riemannian  HMC~\cite{girolami2011riemann,livingstone2014information}.
When a Riemannian metric whose geodesic flow is unknown is chosen, one can use  the trick of  increasing the dimension of the phase space to add symmetries to derive explicit symplectic integrators~\cite{tao2016explicit,cobb2019introducing}.  

Once we have an integrator for the geodesic flow, another important consideration is the construction and tuning of the overall integrator, i.e., the specific composition of  $\Phi^{H_1}_{\dt}$ and $\Phi^{H_2}_{\dt}$.
Traditional numerical integrators  are tuned to provide highly accurate approximations for the trajectories in the limit $\dt \to 0$; for instance, a forth order Runge-Kutta method. However, samplers aim to have the largest possible step size $\dt$ in order to reduce correlations.
One approach consists in tuning the integrator to obtain good density preservation in the Gaussian case.  Another approach consists in tuning the integrator to ensure the shadow Hamiltonian $\tilde{H}$ agrees with $H$ up to the desired order; see~\cite{predescu2012computationally,blanes2014numerical,fernandez2016adaptive,campos2017palindromic,bou2018geometric,clark2011improving}.
We note that when the target density contains two components, one computationally expensive and the other computationally cheap, it may be desirable to %build integrators that 
further split
the potential $H_2(q) = V(q)$ to obtain higher acceptance rates~\cite{tuckerman1992reversible,sexton1992hamiltonian,shahbaba2014split}.
In order to achieve ergodicity in HMC methods, it is usually sufficient to randomise the trajectory length of the integrator~\cite{mackenze1989improved,betancourt2016identifying,wang2013adaptive}. 
However, deriving guarantees on the rate of convergence of HMC is difficult, though recent work have established sufficient conditions for  geometric ergodicity~\cite{durmus2017convergence,livingstone2019geometric}.

Let us also briefly mention some useful upgrades that have been proposed in recent years.
First, whenever the Metropolis step rejects the proposed sample, the (expensive) computation of the numerical trajectory is wasted, and several modifications have been proposed to address this issue,
for example by granting the method extra integration steps when the proposal is rejected~\cite{campos2015extra,dick2014},  or using a dynamic integration with a termination criterion that aims to ensure  the motion is long enough to avoid random walks, but short enough that we do not waste computational effort, such as the  No-U-Turn sampler~\cite{hoffman2014no}.

Second, the Metropolis algorithm gives rise to a \emph{reversible} method which, as discussed above, usually has slower convergence properties.
Modern HMC methods bypass this issue by replacing the heat bath by an Ornstein-Uhlenbeck process, which ensures the overall algorithm is \emph{irreversible}.
In this case, the overall HMC method can be viewed as a geometric integration of the underdamped Langevin diffusion~\cite{ottobre2016function,ottobreMarkovChainMonte2016,heber2020posterior}.
The connection between HMC and Langevin diffusion originates from the desire to replace the Gaussian heat bath with a partial momentum refreshment, yielding a more accurate simulation of dynamical properties and higher acceptance rates~\cite{horowitz1991generalized}.

Third, many modifications of the rudimentary HMC algorithm only provide improvements  when the acceptance rate is sufficiently high. 
A third class of upgrades improves the acceptance rate by using the fact that the shadow Hamiltonian is exactly preserved by the integrator.
%to improve the acceptance rate. 
These \emph{shadow HMC} methods sample from a biased target distribution, defined by the (truncated) shadow Hamiltonian, and correct the bias in the samples via an importance sampler~\cite{izaguirre2004shadow,radivojevic2020modified,radivojevic2018multi}.

Finally, the Metropolis step can be replaced with a multinomial correction that uses the entire numerical trajectory, accepting a given point along it according to the degree by which it distorts the target measure~\cite{betancourt2017conceptual}. 
Some methods entirely skip the accept/reject step, in particular those relying 
on appproximate gradients and surrogates~\cite{strathmann2015gradient,zhang2017hamiltonian}, such as  the \emph{stochastic} HMC methods; such methods approximate  the potential $V(q)$ and its derivative when they are given by a sum over data points, $V(q) = \sum_i V_i(q)$,  by a cheaper sum over a uniformly sampled minibatches~\cite{chen2014stochastic} (these are commonly called
stochastic gradients in machine learning). 
However, this may break the shadow property and reduce the scalable and robust properties of HMC methods~\cite{betancourt2015fundamental}.

\section{Statistical inference with kernel-based discrepancies}

The problem of \emph{parameter inference} consists of estimating an element  $\theta^* \in \Theta$ using a sequence of random functions (or estimators) $ \hat \theta_n : \Omega \to \Theta $, with $\hat \theta_n$ determined by a set of measurements $\{q_1,\ldots, q_n \}$ representing the available experimental data.
In the statistical context,  we  search for the optimal approximation $\mu_{\theta^*}$ of the target measure $\rho$  within a statistical model $\{ \mu_\theta: \theta \in \Theta\}$, with respect to a \emph{discrepancy}
$D: \mathcal{P} \times \mathcal{P} \to [0,\infty]$ over the set of probability measures $\mathcal{P}$.
A common choice of discrepancy is the $\KL$-divergence, and the resulting inference problem can be implemented via the asymptotically optimal maximum likelihood estimators~\cite{van2000asymptotic}. 
As in many applications we are interested in computing 
expectations, 
a particularly suitable notion of discrepancies are the \emph{integral probability pseudometrics} (IPM)~\cite{muller1997integral},
which quantify the worse-case integration error 
with respect to a family of functions $\F$
$$ d_\F(\rho,\mu) \defn \sup_{f \in \F} \left| \int f \dd \rho - \int f \dd \mu \right|.$$
An apparent difficulty  arises with IPMs in that we need to compute a supremum, which will be intractable for most choices of $\F$.
Observe, however, that if $\F$ were the unit ball of a normed vector space $\H$, and integration with respect to $\rho$ and $\mu$ was a continuous linear functional on $\H$, then 
$d_\F(\rho,\mu)$ would correspond to the distance between $\rho$ and $\mu$ in the dual norm over the dual $\H^*$, i.e., $d_\F(\rho,\mu) = \| \rho  - \mu  \|_*$.
Conveniently, reproducing kernel Hilbert spaces (RKHS)  are precisely Hilbert spaces over which the Dirac distributions $\delta_x : f \mapsto f(x)$ act continuously~\cite{aronszajn1950theory,berlinet2011reproducing,steinwart2008support}, and, more generally, the probability distributions that act continuously by integration on a RKHS $\H$ are exactly those for which all elements of $\H$  are integrable~\cite{barp2022simonlester}.
Denoting by $\mathcal{P}_\H$ the set of such probability measures, so that by definition $\delta_x \in \mathcal{P}_\H$,
we can define the \emph{Maximum Mean Discrepancy} (MMD) as
$$ \mmd: \mathcal{P}_\H \times \mathcal{P}_\H \to [0,\infty), \qquad  \mmd [\rho \mid \mu] = \| \rho - \mu \|,$$
where we further used the Riesz representation isomorphism to view $\rho,\mu \in \mathcal{P}_\H \subset \H^* \cong \H$ as elements of $\H$.
The map $\mathcal{P}_\H \to \H$ is usually referred to as the \emph{mean embedding}~\cite{sriperumbudur2010hilbert,muandet2016kernel}.
The angles between the mean embedding of Dirac distributions play a central role in the study of RKHS, and indeed characterise them. 
They define the \emph{reproducing kernel} 
 $$k: \M \times \M \to \R, \qquad k(x,y) \defn \metric{\delta_x}{\delta_y},$$
with $\metric{\cdot}{\cdot}$ denoting the  inner product on $\H$, from which we can obtain a practical expression for the squared MMD:
$$ \mmd^2[\rho \mid \mu] = \iint  k(x,y)(\rho-\mu) (dy) (\rho-\mu) (\dd x).  $$

\subsection{Topological methods for MMDs}

A key feature of RKHS, as identified by Laurent Schwartz, is the fact they are  \emph{Hilbertian subspaces}, i.e., 
Hilbert spaces continuously embedded within a topological vector space $\T$, denoted 
$\H \hookrightarrow \T$~\cite{schwartz1964sous}.
In this context, by composing the transpose of the inclusion $\H \hookrightarrow \T$ with the Riesz isomorphism, we can define  a (generalised) mean embedding as the weakly-continuous positive map 
$$ \phi: \T^* \hookrightarrow \H^* \to \H.$$
This mapping allows us to transfer structures between $\H$ and $\T^*$, an example of which is the  MMD, which is nothing else than the pullback of the Hilbert space metric from $\H$ to $\T^*$. 
Some important examples of $\T$ are 
$C_0$, $C_c^\infty$ and  $\R^\M$ (with their canonical topologies), whose duals are the spaces of finite Radon measures, 
Schwartz distributions, and measures with finite support, respectively~\cite{simon2018kernel}.
In particular a RKHS, as defined above, is any Hilbert space satisfying $\H \hookrightarrow \R^\M$. More generally, when $\T$ is continuously embedded in the space of $\R^n$-valued functions on $\M$---as in the examples above, which have $n=1$---then $\H$ inherits (and can be characterised in terms of) a reproducing kernel $K:\M \times \M \to \R^{n\times n}$, 
defined $\forall v,u \in \R^n$ by
$$ v^\top K(x,y)u = \delta_x^v \left[ \phi (\delta_y^u)\right],$$
where $\delta_x^u:h \mapsto u \cdot h(x)$;
but this need not be the case in general, and we will employ Hilbertian subspaces with no reproducing kernel to construct the score-matching discrepancy. 

This geometric description of RKHS and MMD allows us to swiftly apply topological methods in their analysis.
For example, in order for $\mmd^2$ to be a valid notion of statistical divergence, it should accurately discriminate distinct distributions, in the sense that $\mmd[\rho \mid \mu]=0$ iff $\rho = \mu$.
By construction,  MMD will be \emph{characteristic to a subset of $\T^*$}, that is be able to distinguish its elements,  iff $\phi$ is injective.
The Hahn--Banach theorem further shows that this is equivalent to the denseness of $\H$ in $\T$, 
reducing the matter to a topological question~\cite{sriperumbudur2011universality,sriperumbudur2010hilbert,simon2018kernel}. 
In many applications, we typically would like $\T^*$ to be the set of probability measures, but the latter is not even a vector space. Instead, just as is commonly done to define (statistical) manifolds, it is desirable to embed $\mathcal{P}$ within a more structured space, such as the space of finite Radon measures $C_0^*$.
Characteristicness to $ C_0^*$ is also known as \emph{universality} in learning theory, since  such RKHS are dense in $\l^2(\mu)$ for any $\mu \in \mathcal{P}$, which enables the method to learn the target function independently of the data-generating distribution~\cite{carmeli2010vector}.
However, in many important cases, we are interested in analysing the denseness of $\H$ in a space other than $C_0$.
For instance, in the case of unbounded reproducing kernels, we cannot aim to separate all finite distributions, since the RKHS  will contain unbounded functions and the MMD will only be defined on a subset of $\mathcal{P}$. 
In the particular case of the KSDs discussed below, which are given by transforming a base RKHS into a Stein RKHS via a differential operator, the characteristiness of the Stein RKHS to a set of probability measures is equivalent to the characteristiness of the base RKHS to  more general spaces $\T^*$ of Schwartz distributions~\cite{barp2022simonlester}.

Moreover, the ability of MMD to discriminate distributions  is also useful to ensure it further metrises, or at least controls, weak convergence, and thus provide a suitable quantification of the discrepancy between unequal distributions.
Indeed, on non-compact locally compact Hausdorff spaces such as $\R^d$, when $\H \hookrightarrow C_0$, then $\mmd$ will metrise weak convergence (of probability measures) iff the kernel
$k$ is continuous and $\H$ is characteristic to the space of finite Radon measures~\cite{simon2020metrizing}.
The fact that the RKHS must separate all finite measures in order to metrise weak convergence results from the fact that otherwise MMD cannot in general prevent positive
measures from degenerating into the null measure on non-compact spaces, beyond the  family of translation-invariant kernels, for which characteristicness to the sets of probability measures or that of finite measures are in fact equivalent~\cite{simon2018kernel}.
It is also possible to 
prevent probability mass from escaping to infinity---when the topology of the sequence of distributions is relatively compact with respect to the weak topology on the space of distributions---since, in that case, standard topological arguments relate MMD and weak convergence via characteristicness to $\mathcal{P}$~\cite{ethier2009markov}.
For example, by Prokhorov's theorem we may use the  tightness of a sequence of distributions to ensure characteristic MMDs detect any loss of mass, and thus control weak convergence~\cite{gorham2017measuring}.

\subsection{Smooth measures and KSDs}

 MMD have a computationally tractable expression whenever $\rho,\mu$ are discrete measures, or at least tractable $U$-statistics when their samples are readily available.
 Many applications involve distributions that are
 smooth and fully supported,  but hard to sample from.
Recalling the definition of $d_\F(\rho,\mu)$, 
it would be useful to construct a MMD for which the set $\F$ consists of functions whose integral under $\mu$  is tractable, for example equal to zero; the MMD would then reduce to a double integration with respect to $\rho$.
%In order
To achieve this, we will 
leverage ideas from \emph{Stein's method}~\cite{stein1972bound,anastasiou2021stein}, and apply Stein operators to a given RKHS 
%in order 
so as to construct a \emph{Stein RKHS} whose elements have vanishing expectation under a distribution of interest.

\subsubsection{The canonical Stein operator and Poincaré duality}

%In order 
To gain intuition on Stein operators, we begin by considering the integral with respect to $\mu$ as a linear operator on test functions,
$ \mu : C_c^{\infty}(\M) \to \R$, with $ \mu f \defn \int f \dd \mu, $
and we are interested in generating test functions in the kernel of this operator (i.e., with vanishing expectations).
There are two fundamental theorems that help us understand the integral-differential geometry of the manifold: de Rham's theorem and Poincar\' e duality.
The former relates the topology of the manifold to information on the solutions of differential equations defined over the manifold~\cite{lee2013smooth}.
The latter (which contains the fundamental theorem of calculus) describes the properties of the integral pairing $(\alpha, \beta) \mapsto \int \alpha \wedge \beta $ of differential forms,
which include the pairing of test functions with smooth measures $(f,\mu) \mapsto \int f  \dd\mu$.
While these results are canonical statements about the manifold, 
we can turn them into  measure-theoretic statements by means of the isomorphism $\mu^\sharp$.  
In particular, when $\M$ is connected,
there is an isomorphism between the top  compactly supported twisted de Rham cohomology group $H^n_c(\M)$ (which depends on the topology of $\M$) and $\R$ given by integration,  $\omega \mapsto \int_\M \omega $. 
Applying  the transformation $\mu^\sharp$ to this isomorphism  yields the isomorphism of vector spaces
$$ \mu: C_c^{\infty}(\M)/\mathrm{Im}(\div_\mu|_c) \to \R,$$
where $\div_\mu |_c:\mathfrak X_c(\M) \to C_c^\infty(\M)$ is the divergence operator restricted to the set of compactly supported vector fields $\mathfrak X_c(\M)$.
Hence, if $h,f \in C_c^{\infty}(\M)$, 
then 
$$ \int f \dd \mu = \int h \dd \mu \quad \iff  \quad f=h + \div_\mu(X) \quad \text{ for some } X \in \mathfrak X_c(\M) .$$
Consequently, 
$$\mu^{-1}(\{0\})= \{\div_\mu(X) : X \in \mathfrak X_c(\M) \} . $$ 
Thus, the test functions that integrate to zero are precisely those that can be written as the divergence of compactly supported vector fields.
In particular, on compact manifolds, there is a canonical Stein operator, $\div_\mu$,  which turns vector fields into functions with vanishing expectations.
For other types of manifolds, 
one can obtain similar \emph{dualities} by using other classes of differential forms, such as the square-integrable ones, or by allowing boundaries. 
For our purposes, the above is sufficient to motivate calling
 $$ \steinop_\mu \defn \div_\mu |_{\mathfrak X_\mu}: \mathfrak X_\mu \to C^{\infty}(\M)$$
the \emph{canonical Stein operator},
whose domain $\mathfrak X_\mu$, called the \emph{Stein class}, is any set of vector fields satisfying the desired property that
$ \mathbb E_\mu [\steinop_\mu(X)]\defn \int \steinop_\mu(X) \dd \mu =0$, for all $ X \in \mathfrak X_\mu$.

If we have a bracket $\bi$ on $\M$, we can turn the canonical Stein operator on vector fields into a 2nd order differential operator  acting on functions, the $\bi$-Stein operator
on the Stein class 
 $$C^{\infty}_\mu  \defn\{f \in C^{\infty}(\M): X_f^\bi \in \mathfrak X_\mu \} ,$$
 by
$$  \steinop_\mu^\bi: C^{\infty}_\mu \to C^{\infty}(\M), \qquad  \steinop_\mu^\bi f \defn \div_\mu(X_f^\bi). $$
If $\mu = e^{-H}\mu_\M$ then we  have the following useful decomposition:
$$ \div_{e^{-H} \mu_\M}(X_f^\bi)= \div_{\mu_\M}(X_f^\bi) - X_f^\bi(H).$$
Let us give some important examples of bracket Stein operators.
 When $\bi \defn \A$  is antisymmetric, the $\A$-Stein operator is simply a 1st order differential operator, namely the $\mu$-preserving \emph{curl vector field} 
  $ \steinop_\mu^{ A} = \curl_\mu(\A)$.
  When $\bi$ is Riemannian,  and $\mu_\M$ is the Riemannian measure, then 
 $$\steinop_\mu^{\gmet} (f) = \nabla \cdot X_f^\gmet- \metric{\nabla f}{\nabla H} = \Delta f - \metric{\nabla f}{\nabla H}=\Delta f + \metric{\nabla f}{\nabla \log \dd \mu/\dd \mu_\M},$$
  where $\nabla, \Delta, \nabla \cdot, \metric{\cdot}{\cdot}$ are the Riemannian gradient, Laplacian, divergence, and metric, respectively;  the $\gmet^{-1}$-Stein operator becomes the Riemannian Stein operator~\cite{barpRiemannSteinKernelMethod2020,liu2018riemannian,hodgkinson2020reproducing}.
  Hence, the Riemannian Stein operator is the restriction of the canonical Stein operator to gradient vector fields.
  In general, decomposing the bracket into its symmetric and antisymmetric parts, $\bi \defn  \S +  \A$, we obtain the following useful decomposition of the $\bi$-Stein operator:
  \begin{equation}\label{eq:bi-Stein-S+A}
  \mathrm{S}_\mu^{\S +\A}(f) = \div_\mu(X_f^{\S})+\curl_\mu(\A)(f).
\end{equation}    
In particular, if we restrict ourselves to a symmetric positive semi-definite $\B$,
associated to a set of vector fields $\{Y_i\}$, $X_f^\bi \defn Y_i(f) Y_i$ for any function $f$,
then \eqref{eq:bi-Stein-S+A} corresponds to the generator of a $\mu$-preserving diffusion.
A suitable Stein class  is then the
domain of the generator, since for any function in that domain
$\mathbb E_{\mu}\left[ \mathrm{S}_\mu^{\S +\A}f \right]$ vanishes  by the Fokker-Planck equation.
The construction of Stein operators via measure-preserving diffusions is known as the Barbour approach~\cite{barbour1988stein}.
In fact, the brackets allow us to define a more general notion of Stein operator acting on  1-forms
  $ \{ \alpha  : X^\bi_{\alpha} \in \mathfrak X_\mu \}$, and, on flat Euclidean space,
$
  \mathrm{S}_\mu^{\bi}(\alpha) \defn \div_\mu ( X^\bi_{\alpha})$ recovers the ``diffusion'' Stein operator~\cite{gorham2019measuring}.

\subsubsection{Kernel Stein discrepancies and score matching}

Once we have a Stein operator, we need to construct a Stein class for it, i.e., a set $\mathscr V$ of vector fields (or more general tensor fields) whose image $\F$ under the operator  has mean zero under $\mu$.
The resulting IPM is then known as a \emph{Stein Discrepancy}:
$$
d_{\steinop_\mu(\mathscr{V})}(\mu,\rho) = \sup_{X \in \mathscr{V}} \left |\int \steinop_\mu(X) \dd \rho \right|. $$
The expression $\int \steinop_\mu(X) \dd \rho$ is precisely the rate of change of the $\KL$ divergence
along measures satisfying the continuity equation; an observation that leads to \emph{Stein variational gradient descent} (SVGD) methods to approximate distributions~\cite{liu2016stein,liu2018riemannian}.
Specifically, in SVGD %algorithms 
the target measure is approximated using a finite distribution $\sum_\ell \delta_{x_\ell}$,
where the location of the \emph{particles} $\{x_\ell\}_\ell$ is updated by moving along the direction that maximises the rate of change of $\KL$ within a space of vector fields isomorphic to a RKHS (e.g., the space of gradients  of functions in a RKHS).

When $\steinop_\mu$ is the  canonical Stein operator,
there is a canonical Stein class,  provided by Stokes' theorem,
which essentially only depends on the manifold: for  a connected manifold $\M$, viewing integration as an operator on smooth $\mu$-integrable functions, then 
$ \int f \dd \mu =0 \iff \int \dd \alpha =0$, where $f = \div_\mu(\mu^{\sharp}(\alpha))$.
Unfortunately,  Stokes' theorem usually does not provide a practical description of the differential forms that satisfy $\int \dd \alpha=0$, aside from the compactly supported case.
There are, however, several choices of Stein class constructed from Hilbertian subspaces that lead to computationally tractable Stein discrepancies.
One route consists in constructing a RKHS of mean-zero functions as the image of another RKHS under a Stein operator.
In this case,  we can use $\steinop_\mu$ to map a given RKHS of $\R^d$-valued functions $\H$, with (matrix-valued) reproducing kernel $K$,  into a \emph{Stein RKHS} of $\R$-valued functions $\steinop_\mu(\H)$ associated to a \emph{Stein reproducing kernel}
$k_{\mu}$, given by (here $q$ is the Lebesgue density of $\mu$)
$$ k_{\mu}(x,y) = \frac{1}{q(x) q(y)} \partial_y \cdot \partial_x \cdot \left(q(x) K(x,y)q(y) \right).  $$
The resulting Stein discrepancy can be thought of as an MMD that depends only on $\rho$ and is known as \emph{kernel Stein discrepancy}~\cite{oates2017control}:
$$ \ksd[\rho]^2 \defn  \mmd[\rho \mid \mu]^2 = \int \int  \frac{1}{q(x) q(y)} \partial_y \cdot \partial_x \cdot \left(q(x) K(x,y)q(y) \right) \dd \rho( y) \dd \rho( x).$$
Another class of discrepancies relies on a choice of bracket $\bi$ together with a corollary from Stokes' theorem: 
% $$ \int \bi( \alpha , \dd f ) \Q = - \int f \div_\Q(i_\alpha \bi) \Q,$$
% which, in the Riemannian and symplectic cases respectively, recovers the famous identities $ \int \metric{\nabla f}{X} \vol = - \int f \nabla \cdot X \vol$ and $\int \omega(X_f,X_g) \omega^n =0$. 
% In particular, for $\Q = e^{-H} \mu_\M$ then 
$ \int \steinop_\mu^\bi(\alpha) \dd \rho = \int \alpha (X^{\bi^*}_{H-K}) \dd \rho$, where $e^{-H}$ and $e^{-K}$ are the densities of $\rho$ and $\mu$ with respect to a common smooth measure (below the Riemannian one), while $\bi^*$ is the dual bracket (the transpose of $\bi$).
% \textcolor{red}{(GF: Just double checking if there is a typo since $d\rho$ appears in both integrals? When reading these integrals I was looking for something like  $\int_{\partial R} \alpha = \int_{R} d\alpha$, i.e., it wasn't clear at first a ``volume'' vs. ``area'' integral there.)}
%A: Yes I double checked
We can thus re-write the Stein discrepancy as
\begin{equation}
\sup_{\alpha \in \mathscr{A}} \left | \int \alpha(X^{\bi^*}_{H-K}) \dd \rho \right |
\end{equation}  
over some family of 1-forms $\mathscr{A}$.
As we did previously,  we can ``remove'' the supremum by re-writing the above as a supremum
over some unit ball of a continuous linear functional.
This can be achieved once we have a Riemannian metric $\metric{\cdot}{\cdot}$, which induces a natural inner product that is central to the  theory of Harmonic forms, namely
$
(\alpha,\beta)_\mu \defn
\int \metric{\alpha}{\beta} \dd \mu$.
 In particular, taking as $\mathscr{A}$ the smooth compactly supported 1-forms in
 the unit ball of  
$\l^2(T^*\M,\mu)$---the Hilbert space of square $\mu$-integrable 1-forms---the Stein discrepancy recovers a generalisation of the  score matching~\cite{hyvarinen2005estimation}:
\begin{equation}\label{eq:B-score-matching}
  \operatorname{SM}_\bi[\rho \mid \mu ]  = \int  \|   X_{H-K}^{\bi^{ *}}  \|^2 \dd \rho = \mathbb E_\rho \left[ \|   X_{H-K}^{\bi^{ *}}  \|^2\right].
\end{equation}
It is worth noting that, while $\l^2(T^*\M,\mu)$ is not a RKHS, and does not have a reproducing kernel, it remains a Hilbertian subspace of the space of de Rham currents.
When $\bi$ is Riemannian we recover the Riemannian score matching~\cite{barpRiemannSteinKernelMethod2020}
$$ \operatorname{SM}_{G}[\rho \mid \mu ] = \int \| \nabla H -  \nabla K\|^2 \dd \rho,$$
while in Euclidean space \eqref{eq:B-score-matching} yields the diffusion score matching~\cite{NEURIPS2019_ba7609ee}.

\subsection{Information geometry of MMDs  and natural gradient descent}

MMDs and Stein discrepancies have proved to be important tools in a wide range of contexts, from hypothesis testing and  training generative neural networks to measuring sample quality~\cite{chen2019stein,chwialkowski2016kernel,liu2016kernelized,gretton2012kernel,abdulleAcceleratedConvergenceEquilibrium2019,dziugaite2015training}.
In the context of statistical inference, once we have chosen a suitable discrepancy, $D$,  and a statistical model, $\{ \mu_\Theta \}$,
our aim is to find the best approximation of the target distribution within the model; this corresponds to solving the optimisation problem 
$\theta^* \in \argmin_{\theta \in \Theta} D[\targ \mid \mu_\theta ]$.
As mentioned previously,  computing the value of the discrepancy $D[\targ \mid \mu_\theta ]$ is computationally challenging.
Fortunately,  we can often obtain robust Stein discrepancy estimators for smooth
 statistical models, whose distributions have a smooth positive Lebesgue density, as well as  MMD estimators for generative model  that are easy to sample from but have intractable model densities.

In either case, once we have an estimator $\hat D_m$ based on $m$ samples from the target, we must solve the approximate optimisation problem 
$\theta^*_m \in \argmin_{\theta \in \Theta} \hat D_m[\targ \mid \mu_\theta ]$.
When the function $\hat D_m[\targ \mid \mu_\theta ]$ is smooth, this may be done via the accelerated Hamiltonian-based optimisation methods previously discussed (section \refsec{sec:opt}). 
If $D$ is a divergence function,   one can also usually improve the speed of convergence by following the \emph{natural gradient descent}, associated with the information Riemannian metric $\gmet_{\theta}$ induced by $D$~\cite{park2000adaptive,chen2018natural,karakida2016adaptive,kakade2001natural}.
In practice, this leads to implementing the update
$$ \hat \theta_{t+1} = \hat \theta_t -\gamma_t \hat \gmet_{\theta_t}^{-1} \partial_{\theta_t} \hat D_m[\targ \mid \mu_\theta ], $$
where $\{\gamma_t\}$ is an appropriate sequence of step sizes,  and
$\hat{\gmet}_{\theta}^{-1} $ is the inverse of a regularised estimate of the information tensor~\cite{bonnabel2013stochastic}.
Finally, note that there is a deep connection between divergences and the geometric mechanics discussed in sampling and optimisation, as any divergence may be interpreted as a \emph{discrete Lagrangian}, and hence generates a symplectic structure and integrator~\cite{leok2017connecting}.

\subsubsection{Minimum Stein discrepancy estimators}\label{sec:Minimum SD estimators}

When the model $\{ \mu_\theta \}$ consists of smooth measures with positive  densities $\{q_\theta\}$, and we have access to samples $\{ x_\ell \}$ from the target, the Stein discrepancies  offer a flexible family of inference methods.
For SM we can use the estimator 
$$ \hat \sm_m[\targ \mid \mu_\theta] = \frac1 m \sum_{\ell=1}^m \left( \| \bi^T \partial_x \log q_\theta \|^2_2 +2 \partial_x \cdot (\bi \bi^T \partial_x \log q_\theta) \right)(x_\ell)$$
combined with the following expression for the information tensor:
$$ (\gmet_{\theta})_{ij} = \int \bi^T \partial_x \partial_{\theta^i} \log q_\theta \cdot \bi^T \partial_x \partial_{\theta^j} \log q_\theta \, \dd  \mu_\theta.$$
For KSD
it is convenient to choose a family of matrix kernels $K_\theta(x,y) = \bi_\theta(x) k(x,y) \bi_\theta(y)^T $,
for some scalar kernel $k$, and 
parameter-dependent matrix function $\bi_\theta$.
Denoting the associated Stein reproducing kernel by $k_{\mu_\theta,\theta}$,
we have the  unbiased estimator
$$ \hat \ksd_m[\targ] = \frac1{m(m-1)}\sum_{i \neq j}^m k_{\mu_\theta,\theta}(x_i,x_j) , $$ 
and information tensor 
$$ (\gmet_{\theta})_{ij} = \iint \left(\partial_x \partial_{\theta^j} \log q_\theta \right)^T \bi_\theta(x) k(x,y) \bi_\theta^T(y)
\partial_x \partial_{\theta^i} \log q_\theta \dd \mu_\theta(x) \dd \mu_\theta(y).$$
The parameters $\bi$ and $k$, and the choice of statistical model, can often be adjusted to achieve characteristiness, consistency, bias-robustness, and obtain central limit theorems; see~\cite{NEURIPS2019_ba7609ee} for details, and for numerical experiments showing an acceleration induced by the information Riemannian metric.

\subsubsection{Likelihood-free inference with generative models}\label{sec:likelihood-free MMD estimators}

For many applications of interests, the densities of the model, $\{ \mu_\theta \}$, cannot be evaluated or differentiated. We thus  need density-free inference methods.
This is the case, for instance, in the context of generative models wherein  $\mu_\theta$
is the pushforward of a distribution $\mu$, from which we can sample efficiently, with respect to a generator function $T_\theta$.
Then, the minimum Stein discrepancy estimators based on $\ksd$ and $\sm$, or other discrepancies that rely on the scores, are intractable.
The MMDs are suited to this case
since they depend on the target and model only through integration, which can be straightforwardly estimated using the samples.
The associated information tensor is 
$$ (\gmet_{\theta})_{ij} = \iint
\partial_\theta T_\theta(u)^T \partial_x \partial_y k(x,y) |_{(T_\theta(u),T_\theta(v))} \partial_\theta T_\theta(v) \dd \mu( u) \dd \mu(v).$$
Under appropriate choices of kernels and models one can derive theoretical guarantees, such as concentration/generalisation bounds, consistency, asymptotic normality, and robustness; see, e.g.,
\cite{briol2019statistical,gretton2009fast,dziugaite2015training}.
Moreover, many  approaches to kernel selection in a wide range of contexts have been studied, which include the median heuristic or maximising the power of hypothesis tests, and in practice mixtures of Gaussian kernels are often employed~\cite{briol2019statistical,dziugaite2015training,garreau2017large,sutherland2016generative,ramdas2015adaptivity,li2017mmd}.

\section{Adaptive agents through active inference}

The previous sections have established some of the mathematical fundaments of optimisation, sampling and inference. In this final section, we close with a generic use case called \emph{active inference}. Active inference is a general framework for describing and designing adaptive agents that unifies many aspects of behaviour---including perception, planning and learning---as processes of inference.
Active inference emerged in the late 2000s as a unifying theory of human brain function ~\cite{fristonFreeEnergyPrinciple2006,fristonFreeenergyPrincipleUnified2010,fristonActionBehaviorFreeenergy2010,fristonActiveInferenceEpistemic2015}, and has since been applied to simulate a wide range of behaviours in neuroscience~\cite{dacostaActiveInferenceDiscrete2020,parrComputationalNeurologyActive2019}, machine learning~\cite{millidgeDeepActiveInference2020,fountasDeepActiveInference2020,mazzagliaContrastiveActiveInference2021}, and robotics~\cite{lanillosActiveInferenceRobotics2021,dacostaHowActiveInference2022a}. In what follows, we derive the objective functional overarching decision-making in active inference and describe its information geometric structure, revealing several special cases that are established notions in statistics, cognitive science and engineering. Finally, we exploit this geometric structure in a generic framework for designing adaptive agents.

\subsection{Modelling adaptive decision-making}

\subsubsection{Behaviour, agents and environments}

We define behaviour as the interaction between an agent and its environment. Together the agent and its environment form a \textit{system} that evolves over time according to a stochastic process $x$. This definition entails a notion of time $\mathcal T$, which may be discrete or continuous, and a state space $\mathcal X$, which should be a measure space (e.g., discrete space, manifold, etc.). A stochastic process $x$ is a time-indexed collection of random variables $x_t$ on the state space. More concisely, it is a random variable over trajectories on the state space $\mathcal T \to \mathcal X$:
    \begin{align*}
       % x : \Omega &\to (\overbrace{\mathcal T \to \mathcal X}^{Paths})\,, \: \omega \mapsto x(\omega) \\ \iff x_{t} : \Omega &\to \underbrace{\mathcal X}_{Space}\,,\:\omega \mapsto x(\omega)(t)\,, \:\forall t\in \mathcal T.
       x : \Omega \to (\mathcal T \to \mathcal X), \  \omega \mapsto x(\omega)  \quad \iff  \quad  x_{t} : \Omega \to \mathcal X, \ \omega \mapsto x(\omega)(t)  \quad \forall t\in \mathcal T.
    \end{align*}
We denote by $P$ the probability density of $x$ on the space of paths $\mathcal T \to \mathcal X$ with respect to a pre-specified base measure.

%Typically, systems comprising an agent and its environment have four sets of states: \textit{internal} and \textit{external} states that are separated by \textit{observable} and \textit{active} states. This separation is in terms of sparse coupling such that external states can only influence themselves and outcome states, while internal states can only influence themselves and active states. This enables a statistical separation of external and internal states, while allowing for a reciprocal exchange with the environment, via observable and active states. For simplicity, we will combine active and internal states and call them \textit{autonomous} states (because they are not directly influenced by external states).

\begin{figure}[t!]
\centering
    \includegraphics[width=0.7\textwidth]{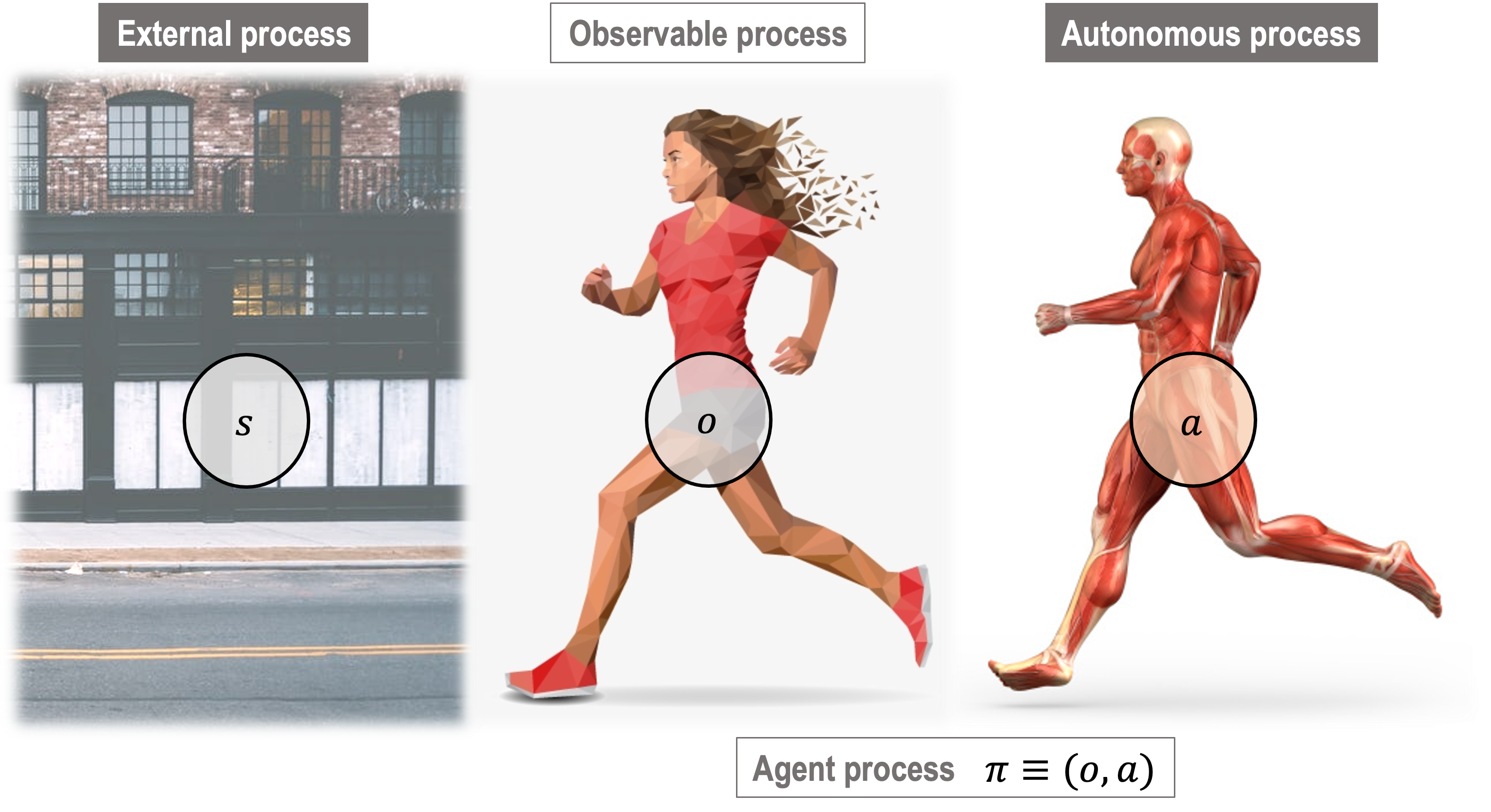}
    \caption{\emph{Partitions and agents.} This figure illustrates a human (agent $\pi$) interacting with its environment (external process $s$), and the resulting partition into external $s$, observable $o$, and autonomous $a$ processes. The external states are the environment, which the agent does not have direct access to, but which is sampled through the observable states. These could include states of the sensory epithelia (e.g., eyes and skin).
    The autonomous states constitute the muscles and nervous system that factor available information into decisions. In the example of human behaviour, the environment causes observations (i.e., sensations), which informs a nervous and muscular response, which in turn influences the environment. In general, autonomous responses may be informed by all past agent states $\pi_{\leq t}=(o_{\leq t},a_{\leq t})$ (the information available to the agent at time $t$), which means that the systems we are describing are typically non-Markovian.}
    \label{fig: partition}
\end{figure}

Typically, systems comprising an agent and its environment have three sets of states: \textit{external} states are unknown to the agent and constitute the environment; the \textit{observable} states are those agent’s states that the agent sees but cannot directly control; finally, the \textit{autonomous} states are those agent’s states that the agent sees and can directly control. This produces a partition of the state space $\mathcal X$ into states external to the agent $\mathcal S$ and states belonging to the agent $\Pi$, which themselves comprise observable $\mathcal O$ and autonomous states $\mathcal A$. As a consequence, the system $x$ can be decomposed into external $s$, observable $o$, and autonomous $a$ processes:
    \begin{align*}
       % \mathcal X \defn \underbrace{\mathcal S}_{\text{External}} \times \underbrace{\Pi}_{\text{Agent}} &\defn \underbrace{\mathcal S}_{\text{External}} \times \underbrace{\mathcal O}_{\text{Observable}} \times \underbrace{\mathcal A}_{\text{Autonomous}},\\
        \mathcal X \defn \mathcal S \times \Pi \defn \mathcal S \times \mathcal O \times \mathcal A
       \quad \implies \quad  x \defn (s, \pi) \defn (s, o, a),
    \end{align*}
here written as random interacting trajectories on their respective spaces (see Figure~\ref{fig: partition} for an illustration).

\subsubsection{Decision-making in precise agents}

The description of behaviour adopted so far could, in principle, describe particles interacting with a heat bath~\cite{pavliotisStochasticProcessesApplications2014} as well as humans interacting with their environment (see Figure~\ref{fig: partition}). We would like a description that accounts for purposeful behaviour~\cite{fristonStochasticChaosMarkov2021,fristonFreeEnergyPrinciple2022,dacostaBayesianMechanicsStationary2021a,dacostaActiveInferenceDiscrete2020,fristonSophisticatedInference2021,parrMemoryMarkovBlankets2021}. So what distinguishes people from small particles? An obvious distinction is that human behaviour is subject to classical as opposed to statistical mechanics. In other words, people are precise agents, with conservative dynamics.

\begin{definition}[Precise agent]
\label{def: precise agent}
    An agent is precise when it evolves deterministically in a (possibly) stochastic environment, i.e., when $P(\pi \mid s)$ is a Dirac measure for any $s$.
   % \begin{equation}
%    \label{eq: precise agent}
 %   \begin{split}
 %       &P(\pi \mid s, \pi_{\leq t}) \text{ is a Dirac measure for any } s,\pi_{\leq t}\\
 %       \iff &\begin{cases}
  %        P(a \mid s, o, \pi_{\leq t})\:\:\:%\stackrel{\eqref{eq: MB}}{=}P(a \mid o, \pi_{\leq t})
  %       \text{is a Dirac measure for any } s,o,\pi_{\leq t}.\\
  %       P(o \mid s,a, \pi_{\leq t}) \text{ "------------------------------------"} \,a,\pi_{\leq t}.
  %      \end{cases} 
  %  \end{split}
%    \end{equation}
    For example,
    \begin{align*}
    d s_t = f(s_t,\pi_t)dt + d w_t,\qquad
    d \pi_t = g(s_t,\pi_t)dt.
\end{align*} %if we had stochasticity on o, we would get APD. See comput 10.1.22
\end{definition}

%\subsubsection{Decision-making in precise agents}

At any moment in time $t$, the agent has access, at most, to its past trajectory $\pi_{\leq t}$, and has agency over its future autonomous trajectory $a_{>t}$. We define a \textit{decision} to be a choice of autonomous trajectory in the future  given available knowledge $a_{>t}\mid \pi_{\leq t}$. We interpret $P(s,o \mid \pi_{\leq t})$ as expressing the agent's \textit{preferences} over external and observable trajectories given available data, and $P(s,o \mid a_{>t},\pi_{\leq t})$ as expressing the agent's \textit{predictions} over external and observable paths given a decision. 
Crucially, decision-making in (precise) agents is a functional of the agent's predictions and preferences\footnote{Under the precise agent assumption (Definition ~\ref{def: precise agent}) it is straightforward to show that $\E_{P(s,o\mid a_{>t} ,\pi_{\leq t})}[\log P(o \mid s,a_{>t}, \pi_{\leq t})-\log P(a_{>t} \mid s,o,\pi_{\leq t})]=0$ when the path space $\mathcal T \to \mathcal X$ is countable \cite{tenka}. Presumably, this equality can be extended to more general path spaces by a limiting argument.}
%In the following, we describe how (precise) agents' decisions $P(a_{>t} \mid \pi_{\leq t})$ relate to predictions and preferences.
%Since both observable and autonomous processes evolve deterministically in precise agents, the Shannon entropy of observable and autonomous paths are equal:\footnote{To obtain \eqref{eq: classical particle} note that when the path space is finite, we have the equality $\ent[P(s,o \mid \pi_{\leq t})]-\ent[P(s,a \mid \pi_{\leq t})]= \E_{P(x \mid \pi_{\leq t})}[\log P(a \mid s,o,\pi_{\leq t})-\log P(o \mid s,a ,\pi_{\leq t})]=0$ due to Definition~\ref{def: precise agent}. This equality can be extended to more general path spaces via a limiting argument, by expressing entropies as a limiting density of discrete points~\cite{jaynesInformationTheoryStatistical1957}.}
%\begin{equation}
%\label{eq: classical particle}
%\ent[P(s,o \mid \pi_{\leq t})]
  %\stackrel{\eqref{eq: precise agent}}{=} \E_{P(s \mid  \pi_{\leq t})}[\ent[P(o , a \mid s , \pi_{\leq t})]]\stackrel{\eqref{eq: characteristic agent}}{=} 
%  =\ent[P(s,a \mid \pi_{\leq t})] \quad \text{ for any } \pi_{\leq t}.
%\end{equation}
%Crucially, this allows us to express agents' decisions as a functional of their predictions and preferences\cite{fristonFreeEnergyPrinciple2022}:\footnote{The 2nd equality follows from \eqref{eq: classical particle}, and the implication follows since the $\KL$ divergence vanishes only when its arguments are equal.}
\begin{align}
    &-\log P(a_{>t} \mid \pi_{\leq t}) = \E_{P(s,o\mid a_{>t} ,\pi_{\leq t})}[-\log P(a_{>t} \mid \pi_{\leq t})]\nonumber= \E[\log P(s,o \mid a_{>t},\pi_{\leq t})-\log P(s,o, a_{>t} \mid\pi_{\leq t})]\nonumber\\
    &= \E[ \log P(s \mid a_{>t}, \pi_{\leq t})- \log P(s,o  \mid \pi_{\leq t})+\underbrace{\log P(o \mid s,a_{>t}, \pi_{\leq t})-\log P(a_{>t} \mid s,o,\pi_{\leq t})}_{=0}]\nonumber\\
  &\Rightarrow -\log P(a_{>t} \mid \pi_{\leq t})  = \E_{P(s,o\mid a_{>t} ,\pi_{\leq t})}[ \log P(s \mid a_{>t}, \pi_{\leq t})- \log P(s,o  \mid \pi_{\leq t})].\label{eq: EFE} \tag{EFE}
\end{align}
This functional is known as an \textit{expected free energy} \eqref{eq: EFE}~\cite{schwartenbeckComputationalMechanismsCuriosity2019,fristonActiveInferenceEpistemic2015,dacostaActiveInferenceDiscrete2020} because it resembles the expectation of the free energy functional (a.k.a. evidence lower bound~\cite{bleiVariationalInferenceReview2017}) used in approximate Bayesian inference~\cite{fristonActiveInferenceEpistemic2015,fristonFreeEnergyPrinciple2022}.
We define \textit{active inference} as Hamilton's principle of least action on expected free energy\footnote{As a negative log density over paths, the expected free energy is an action in the physical sense of the word.} that expresses the most likely decision $\boldsymbol a_{>t}$, where
\begin{equation}
\label{eq: active inference} \tag{AIF}
    \boldsymbol a_{>t} \defn \argmin_{a_{>t}}  -\log P(a_{>t} \mid \pi_{\leq t}).
\end{equation}

\subsubsection{The information geometry of decision-making}
\label{sec: inf geom decision theory}

\begin{figure}[t!]
\centering
    \includegraphics[width=\textwidth]{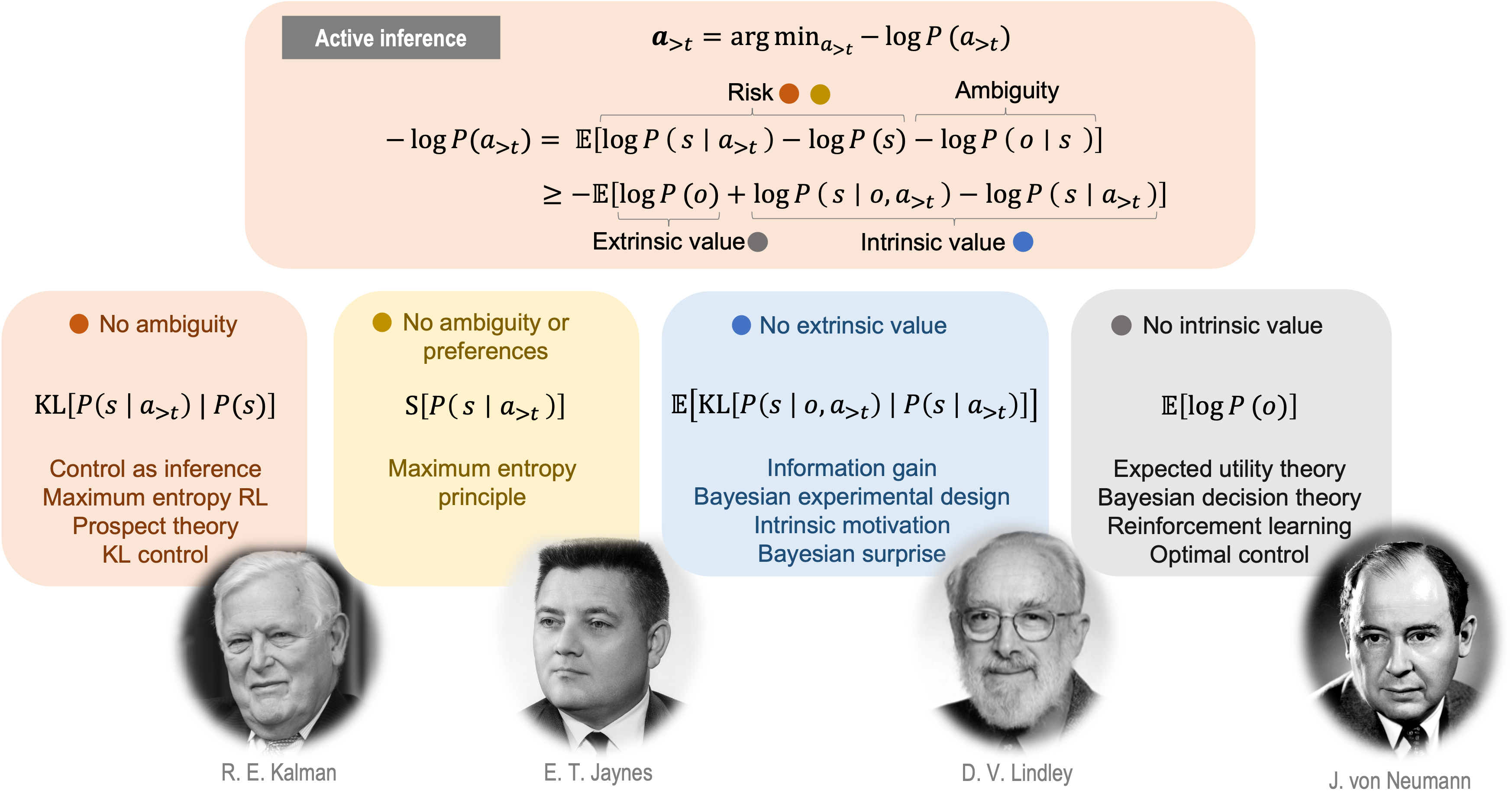}
    \caption{{\emph{Decision-making under active inference}. This figure illustrates various imperatives that underwrite decision-making under active inference in terms of several constructs that predominate in statistics, cognitive science and engineering. These formulations are disclosed when one removes certain sources of uncertainty.
    For example, if we remove ambiguity, decision-making minimises risk, which corresponds to aligning predictions with preferences about the external course of events. This aligns with prospect theory of human choice behaviour in economics~\cite{kahnemanProspectTheoryAnalysis1979} and underwrites modern approaches to control as inference~\cite{levineReinforcementLearningControl2018,rawlikStochasticOptimalControl2013,toussaintRobotTrajectoryOptimization2009}, variously known as Kalman duality~\cite{kalmanNewApproachLinear1960,todorovGeneralDualityOptimal2008a}, KL control~\cite{kappenOptimalControlGraphical2012} and maximum entropy reinforcement learning~\cite{ziebartModelingPurposefulAdaptive2010}.
    If we further remove preferences, decision-making maximises the entropy of external trajectories. This maximum entropy principle~\cite{jaynesInformationTheoryStatistical1957,lasotaChaosFractalsNoise1994} allows one to least commit to a pre-specified external trajectory and therefore keep options open.
    If we reintroduce ambiguity, but ignore preferences, decision-making maximises intrinsic value or expected information gain~\cite{mackayInformationTheoryInference2003}. This underwrites Bayesian experimental design~\cite{lindleyMeasureInformationProvided1956} and active learning in statistics~\cite{mackayInformationBasedObjectiveFunctions1992}, intrinsic motivation and artificial curiosity in machine learning and robotics~\cite{oudeyerWhatIntrinsicMotivation2007,schmidhuberFormalTheoryCreativity2010,bartoNoveltySurprise2013,sunPlanningBeSurprised2011,deciIntrinsicMotivationSelfDetermination1985}. This is mathematically equivalent to optimising expected Bayesian surprise and mutual information, which underwrites visual search~\cite{ittiBayesianSurpriseAttracts2009,parrGenerativeModelsActive2021} and the organisation of our visual apparatus~\cite{barlowPossiblePrinciplesUnderlying1961,linskerPerceptualNeuralOrganization1990,opticanTemporalEncodingTwodimensional1987a}.
    Lastly, if we remove intrinsic value, we are left with maximising extrinsic value or expected utility. This underwrites expected utility theory~\cite{vonneumannTheoryGamesEconomic1944}, game theory, optimal control~\cite{bellmanDynamicProgramming1957, astromOptimalControlMarkov1965a} and reinforcement learning~\cite{bartoReinforcementLearningIntroduction1992}. Bayesian formulations of maximising expected utility under uncertainty are also known as Bayesian decision theory~\cite{bergerStatisticalDecisionTheory1985}. To ease notation, we omitted to condition every distribution in the figure by $\pi_{\leq t}$.}
    }%maybe a ref about infomax principle
    \label{fig: active inference summary}
\end{figure}

Interestingly, active inference \eqref{eq: active inference} looks like it describes agents that engage in purposeful behaviour. Indeed, we can rearrange the expected free energy \eqref{eq: EFE} in several ways, each of which reveals a fundamental trade-off that underwrites decision-making. This allows us to relate active inference to information theoretic formulations of decision-making that predominate in statistics, cognitive science and engineering (see Figure~\ref{fig: active inference summary}).
%When the environment is not fully observed, there is a trade-off between exploration and exploitation. Exploitation refers to fulfilling one's goals (e.g., maximising utility), while exploration refers to gaining information about environmental states and environmental contingencies. The more one knows about the environment, the more one is able to fulfil one's goals. As we will see, autonomous states are selected to weigh exploration and exploitation, which suggests a principled solution to the exploration exploitation dilemma.
\begin{comment}
\begin{align*}
     -\log P(a \mid \pi_{\leq t}) &= \mathbb{E}_{P(s, o \mid a, \pi_{\leq t})}[\overbrace{\log P(s \mid  a, \pi_{\leq t})- \log P(s \mid \pi_{\leq t})}^{\text{Risk}}\overbrace{-\log P(o \mid s, \pi_{\leq t})}^{\text{Ambiguity}}]\\
     &=-\mathbb{E}_{P(s, o \mid a, \pi_{\leq t})}[\underbrace{\log P\left(o \mid \pi_{\leq t}\right)}_{\text{Extrinsic value}}+\underbrace{\log P\left(s \mid o, a, \pi_{\leq t}\right)-\log P\left(s \mid a, \pi_{\leq t}\right)}_{\text{Intrinsic value}}]
\end{align*}
\begin{align*}
     -\log P(a \mid \pi_{\leq t}) &= \mathbb{E}[\log P(s \mid  a, \pi_{\leq t})- \log P(s \mid \pi_{\leq t})-\log P(o \mid s, \pi_{\leq t})]\\
     &=-\mathbb{E}[\log P\left(o \mid \pi_{\leq t}\right)+\log P\left(s \mid o, a, \pi_{\leq t}\right)-\log P\left(s \mid a, \pi_{\leq t}\right)]
\end{align*}
\end{comment}
%For example, we can rewrite decision-making as a trade-off between minimising risk and ambiguity
For example, decision-making minimises both risk and ambiguity:
\vspace*{-10pt}
\begin{equation}
\label{eq: risk ambiguity}
%\begin{split}
    -\log P(a_{>t} \mid \pi_{\leq t}) = \underbrace{\kl\big[\,\overbrace{P(s \mid  a_{>t}, \pi_{\leq t})}^{\text{predicted paths}} \mid \overbrace{P( s\mid \pi_{\leq t})}^{\text{preferred paths}}\,\big]}_{\text{risk}} +
    \underbrace{\E_{P(s,o \mid a_{>t}, \pi_{\leq t})}\big[-\log P(o \mid s , \pi_{\leq t})\big]}_{\text{ambiguity}}.
%\end{split}
\end{equation}
\textit{Risk} refers to the KL divergence between the predicted and preferred external course of events. Minimising risk entails making predicted (external) trajectories fulfil preferred external trajectories. 
In a nutshell, \textit{ambiguity} refers to the expected entropy of future observations, given future external trajectories. An external trajectory that can lead to various distinct observation trajectories is highly ambiguous---and vice-versa. Thus, minimising ambiguity leads to sampling observations that enable to recognise the external course of events. This leads to a type of observational bias commonly known as the streetlight effect~\cite{kaplanConductInquiry1973}: when a person loses their keys at night, they initially search for them under the streetlight because the resulting observations (``I see my keys under the streetlight'' or ``I do not see my keys under the streetlight'') accurately disambiguate external states of affairs.

Similarly, decision-making maximises extrinsic and intrinsic value~\cite{sajidActiveInferenceBayesian2021}:%\footnote{In more elaborate treatments, the latest term can be shown to vanish and the (negative) expected free energy equals the sum of extrinsic and intrinsic value [cite FEP made simpler].}
%\vspace*{-2pt}
\begin{align*}%Comput 7.12.21
   -\log P(a_{>t} \mid \pi_{\leq t}) %&\stackrel{\eqref{eq: MB}}{=}
  &= \mathbb{E}_{P\left(o \mid a_{>t}, \pi_{\leq t}\right)}\big[\underbrace{\kl\left[P\left(s \mid o,a_{>t}, \pi_{\leq t}\right) \mid P\left(s \mid o, \pi_{\leq t}\right)\right]}_{\geq 0}\big]\\[-11pt] -\underbrace{\mathbb{E}_{P(o \mid a_{>t} , \pi_{\leq t})}\big[\log \overbrace{P\left(o \mid \pi_{\leq t}\right)}^{\text {preferred paths}}\big]}_{\text {extrinsic value}}
   &-\underbrace{\mathbb{E}_{P\left(o \mid a_{>t}, \pi_{\leq t}\right)}\big[\kl\left[P\left(s \mid o, a_{>t}, \pi_{\leq t}\right) \mid P\left(s \mid a_{>t}, \pi_{\leq t}\right)\right]\big]}_{\text{intrinsic value}}.
 % \text{ by } \eqref{eq: MB}} ]
\end{align*}
\textit{Extrinsic value} refers to the (log) likelihood of observations under the model of preferences. This corresponds to an expected utility or expected reward in behavioural economics, control theory and reinforcement learning~\cite{vonneumannTheoryGamesEconomic1944,bartoReinforcementLearningIntroduction1992}. In short, maximising extrinsic value leads to sampling observations that are likely under the model of preferences.
\textit{Intrinsic value} refers to the amount of information gained about external courses of events. This measures the expected degree of belief updating about external trajectories under a decision, with versus without future observations. Making decisions to maximise information gain leads to a goal-directed form of exploration~\cite{schwartenbeckComputationalMechanismsCuriosity2019}, driven to answer ``What would happen if I did that?''~\cite{schmidhuberFormalTheoryCreativity2010}. Interestingly, this decision-making procedure underwrites Bayesian experimental design in statistics~\cite{lindleyMeasureInformationProvided1956}, which describes optimal experiments as those that maximise expected information gain. In summary, decision-making weighs the imperatives of maximising utility and information gain, which suggests a principled solution to the exploration-exploitation dilemma~\cite{berger-talExplorationExploitationDilemmaMultidisciplinary2014}.

\subsection{Realising adaptive agents}
\label{sec: realising adaptive agents}

%We derived a generic description of decision-making called active inference (see Figure~\ref{fig: active inference summary}).
We now show how active inference affords a generic recipe to generate adaptive agents.  %The key is to design agents that make decisions consistently with the information geometry of decision-making described in Section~\ref{sec: inf geom decision theory}.
%
%
%
%One can then compute the distribution over agent's decisions, which leads to Executing the agent's decisions then fulfils the agent's preferences.
%\begin{equation*}
%    \E_{\underbrace{P(a \mid \pi_{\leq t})}_{\text{Decisions}}}[\underbrace{P(s,o \mid a , \pi_{\leq t})}_{\text{Predictions}}]= \underbrace{P(s,o \mid \pi_{\leq t})}_{\text{Preferences}}.
%\end{equation*}
%We derive the basic active inference algorithm in discrete time.
%For now, we operate in discrete time and identify autonomous states with actions.
%Loosely speaking we identify autonomous states with actions to emphasise their influence on the environment.
%

\subsubsection{The basic active inference algorithm}

Active inference specifies an agent by a \textit{prediction model} $P(s,o \mid a)$, expressing the distribution of external and observable paths given autonomous paths, and a \textit{preference model} $P(s,o)$, expressing the preferred external and observable trajectories. To aid intuition, we will refer to autonomous states as actions. At any time $t$, the agent knows past observations and actions $\pi_{\leq t}=(o_{\leq t}, a_{\leq t})$, and must make a decision $a_{>t}$. In discrete time, active inference proceeds by assessing the expected free energy of each possible decision and then executing the best one: 
\begin{enumerate}
\item \emph{Preferential inference:} infer preferences about external and observable trajectories, i.e.,
    \begin{equation}
    \label{eq: preference inference}
        \text{ Approximate } P(s,o \mid \pi_{\leq t}) \text{ by } Q(s,o).
    \end{equation}
\item For each possible sequence of future actions $a_{>t}$:
\begin{enumerate}
    \item \emph{Perceptual inference:} infer external and observable paths under the action sequence, i.e., %under the action sequence and available data
    \begin{equation}
    \label{eq: perceptual inference}
        \text{ Approximate } P(s,o \mid a_{>t} , \pi_{\leq t}) \text{ by } Q(s,o \mid a_{>t}).
    \end{equation}
    \item \emph{Planning as inference:} assess the action sequence by evaluating its expected free energy~\eqref{eq: EFE}, i.e., %counterfactual consequences of action
    \begin{equation}
    \label{eq: planning as inference}
        -\log Q(a_{>t} ) \defn \E_{Q(s, o \mid a_{>t})}\big[\log Q(s \mid  a_{>t})- \log Q(s,o) \big].
    \end{equation}
\end{enumerate}
    \item \emph{Decision-making:} execute the most likely decision $\boldsymbol a_{t+1}$ according to
    \begin{equation}
    \begin{split}
        \boldsymbol a_{t+1} = \argmax Q(a_{t+1} ), \qquad
        Q(a_{t+1} )= \sum_{a_{>t}} Q(a_{t+1} \mid a_{>t}) Q(a_{>t}). \label{eq: decision-making}
        \end{split}
    \end{equation}
\end{enumerate}

\subsubsection{Sequential decision-making under uncertainty}

\begin{figure}[t!]
\centering
    \includegraphics[width=\textwidth]{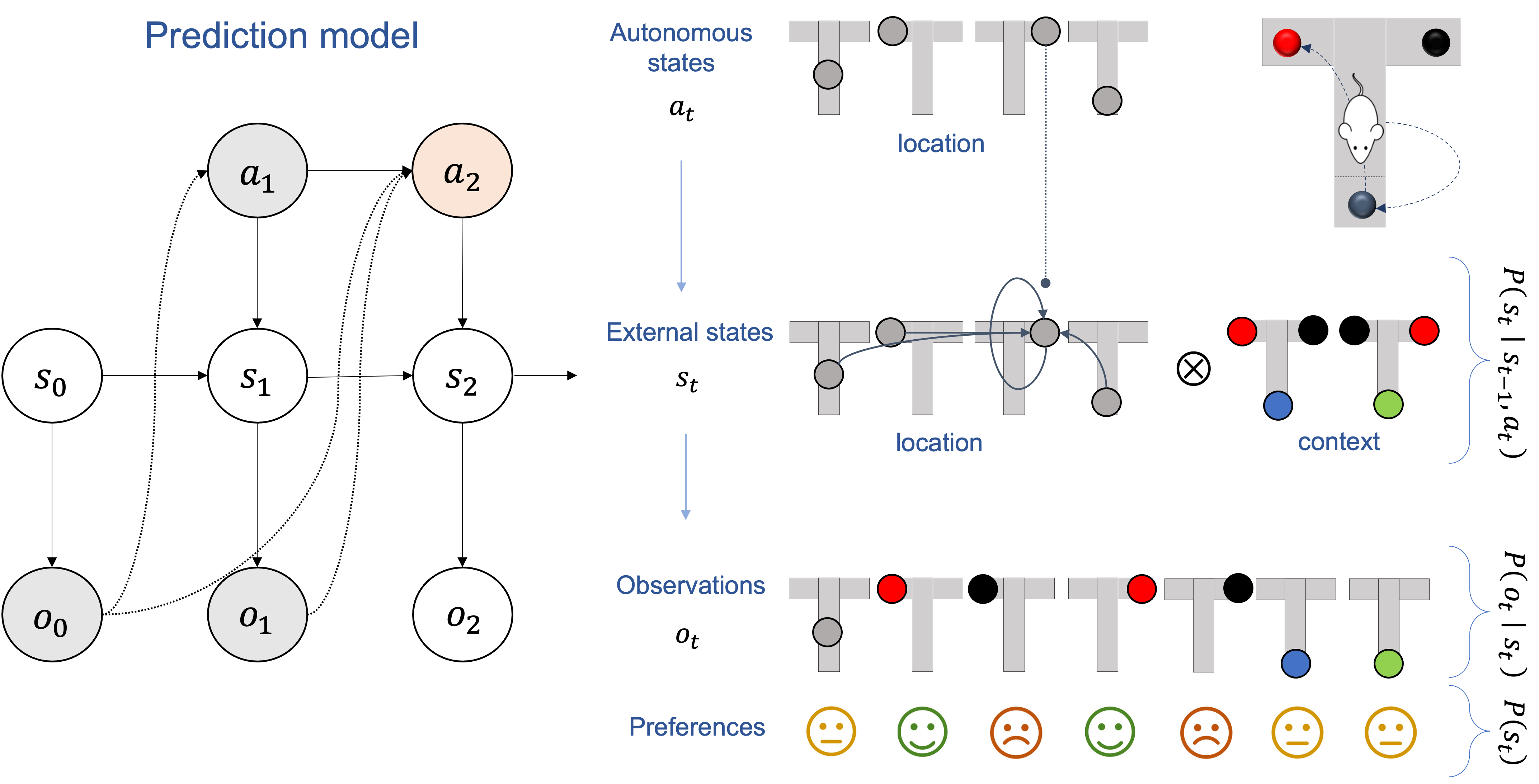}
    \caption{\emph{Sequential decision-making in a T-Maze environment.} \emph{Left:} The agent's prediction model is a partially observed Markov decision process (see text) represented here as a Bayesian network~\cite{bishopPatternRecognitionMachine2006}. %: the circles are random variables while the arrows represent causality \footnote{Causality Pearl}.
    The colour scheme illustrates the problem at $t=2$: the agent must make a decision (in red) based on previous actions and observations
    %its knowledge of present and past actions and observations
    (in grey), which are informative about external states and future observations (in white). \emph{Right:} $s_t$: The T-Maze has four possible spatial \emph{locations}: middle, top-left, top-right, bottom. One of the top locations contains a reward (in red), while the other contains a punishment (in black). The reward's location determines the \emph{context}. The bottom arm contains a cue whose colour (blue or green) discloses the context. Together, location and context determine the external state. $o_t$: The agent observes its spatial location. In addition, when it is at the top of the Maze, it observes the reward or the punishment; when it is at the bottom, it observes the colour of the cue. $a_t$: Each action corresponds to visiting one of the four spatial locations. $P(s_t)$: The agent prefers being at the reward's location ($-\log P(s_t) =+3$) and avoid the punishment's location ($-\log P(s_t) =-3$). All other states have a neutral preference ($-\log P(s_t)=0$). %Under this specification, the agent seeks to collect the reward and avoid the punishment. %By executing actions that minimise both risk and ambiguity \eqref{eq: risk ambiguity}, active inference agents first 1) \textit{explore}: observe the cue in order to infer the reward's location, and then 2) \textit{exploit}: collect the reward.
    $o_0$: The agent is in the middle of the Maze and is unaware of the context. $a_1$: Visiting the bottom or top arms have a lower ambiguity than staying, as they yield observations that disclose the context. However, staying or visiting the bottom arm are safer options, as visiting a top arm risks receiving the punishment. By acting to minimise both risk and ambiguity \eqref{eq: risk ambiguity} the agent goes to the bottom. $o_1$: The agent observes the colour of the cue and hence determines the context. $a_2$: All actions have equal ambiguity as the context is known. Collecting the reward has a lower risk than staying or visiting the middle, which themselves have a lower risk than collecting the punishment. Thus, the agent visits the arm with the reward. See~\cite{fristonActiveInferenceProcess2017} for more details.}
    \label{fig: T-Maze}
\end{figure}

%We explain how the basic active inference algorithm can be used to make sequential decisions under partial information.
A common model of sequential decision-making under uncertainty is a \textit{partially observable Markov decision process} (POMDP). A POMDP is a discrete time model of how actions influence external and observable states. In a POMDP, 1) each external state depends only on the current action and previous external state $P(s_t \mid  s_{t-1}, a_t)$, and 2) each observation depends only on the current external state $P(o_t \mid s_t)$. One can additionally specify 3) a distribution of preferences over external trajectories $P(s)$. Together, 1) \& 2) forms the agent's (POMDP) prediction model, and 2) \& 3) forms the agent's (hidden Markov) preference model, which defines an active inference agent. A simple simulation of active inference on a POMDP is provided in Figure~\ref{fig: T-Maze};
%sequential decisions under uncertainty is provided in Figure~\ref{fig: T-Maze}.
implementation details on generic POMDPs are available in~\cite{dacostaActiveInferenceDiscrete2020,heinsPymdpPythonLibrary2022,sajidActiveInferenceDemystified2021,smithStepbyStepTutorialActive2022}. 
%Implementation details for active inference on POMDPs are available in [67,75,229,230].
For more complex simulations of sequential decision-making (e.g., involving hierarchical POMDPs), please see~\cite{sajidActiveInferenceDemystified2021,fristonActiveInferenceCuriosity2017,parrComputationalNeurologyActive2019,fountasDeepActiveInference2020,millidgeDeepActiveInference2020,catalRobotNavigationHierarchical2021,fristonDeepTemporalModels2018}.

\subsubsection{World model learning as inference}

Due to a lack of domain knowledge, it may be challenging to specify an agent's prediction and preference model. For example, how do external states map to observations? Should external states be represented in a discrete or continuous state space?

In active inference, generative models are learned by inferring their parameters~\cite{sajidActiveInferenceDemystified2021,fristonActiveInferenceLearning2016,dacostaActiveInferenceDiscrete2020} and structure~\cite{smithActiveInferenceApproach2020,dacostaActiveInferenceDiscrete2020,fristonActiveInferenceCuriosity2017,fristonWorldModelLearning2021,wauthierSleepModelReduction2020}. Suppose there is an unknown parameter (or structure variable) $m$ in the prediction model, the preference model or both. By definition, each alternative parameterisation $m$ entails different predictions $P(o,s \mid a , m)$ and preferences $P(o,s \mid m)$. Since unknowns are simply external states, we treat the parameter as an additional external state. We equip the space of parameters with a prior distribution $P(m)$, and define the agent with an augmented prediction (resp. preference) model that combines the different alternatives $P(o,s ,m \mid a) \defn P(o,s \mid a,m)P(m)$ (resp. $P(o,s ,m) \defn P(o,s \mid m)P(m)$). The parameter can then be inferred along with other external states during preferential \eqref{eq: preference inference} or perceptual \eqref{eq: perceptual inference} inference~\cite{fristonActiveInferenceLearning2016,dacostaActiveInferenceDiscrete2020,sajidActiveInferenceDemystified2021}. Better yet, having specified priors over parameters that are independent of actions, we can infer them separately, for example, after fixed-length sequences of decisions to reduce computational cost~\cite{fristonActiveInferenceLearning2016,dacostaActiveInferenceDiscrete2020}. %To further save computational resources, one may infer the single best model from data through maximum a posteriori inference, by specifying $Q(m \mid \pi_{\leq t})$ to be a Dirac distribution.

All this says that a prior $P(m)$ and some data $\pi_{\leq t}$ leads to approximate posterior beliefs $Q(m) \approx P(m \mid \pi_{\leq t})$ about model parameters. But what are the right priors? One way to answer this question lies in optimising a free energy functional $F$ (a.k.a. an evidence lower bound~\cite{bleiVariationalInferenceReview2017}):
\begin{align*}
    F &\equiv \underbrace{\E_{Q(m )}\big[- \log P(m,\pi_{\leq t})\big]}_{\text{energy}}- \underbrace{\ent\big[Q(m )\big]}_{\text{entropy}}\\
    %&=\kl[ \mid P(m \mid \pi_{\leq t})]- \log P(\pi_{\leq t}) \\
    &= \underbrace{\kl\big[Q(m ) \mid P(m)\big]}_{\text{complexity}}- \underbrace{\E_{Q(m )}\big[\log P(\pi_{\leq t} \mid m)\big]}_{\text{accuracy}} . %\label{eq: complexity accuracy}
\end{align*}

Choosing priors that minimise free energy leads to parsimonious models that explain the data at hand~\cite{tschantzLearningActionorientedModels2020}. This follows since maximising accuracy increases the likelihood of the data under the posterior model, while minimising complexity decreases the movement from prior to posterior, which can be seen as a proxy for computational cost. Maximising accuracy usually results in generative models involving universal function approximators~\cite{millidgeDeepActiveInference2020,fountasDeepActiveInference2020,catalRobotNavigationHierarchical2021}, while minimising complexity results in organising representations in sparse, compartmentalised and hierarchical generative models~\cite{fristonActiveInferenceCuriosity2017,fristonWorldModelLearning2021,wauthierSleepModelReduction2020}, where higher levels of the hierarchy encode more abstract representations and vice-versa~\cite{fristonDeepTemporalModels2018}. A computationally efficient method to compare priors by their free energy is Bayesian model reduction~\cite{dacostaActiveInferenceDiscrete2020,fristonBayesianModelReduction2019}. In conclusion, free energy unifies inference and model selection under a single objective function.

\subsubsection{Scaling active inference}

%The main challenge for active inference is to scale to operate in a wide variety of application domains. Here
We conclude by identifying promising scaling methods for active inference that enable computationally tractable implementations in a variety of applications.
%We identify promising scaling methods for active inference that enable computationally tractable implementations in a wide variety of application domains.

Planning for all possible courses of action is computationally expensive as the number of action sequences is exponential in the length of the sequence. One way to finesse this is by planning only for intelligently chosen subsets of action sequences, using sampling algorithms such as Monte-Carlo tree search~\cite{silverMasteringGameGo2016,fountasDeepActiveInference2020,championBranchingTimeActive2021,championBranchingTimeActive2021a,maistoActiveTreeSearch2021}. Monte-Carlo sampling can be also be used to finesse the expectations inherent in assessing action sequences \eqref{eq: planning as inference}~\cite{fountasDeepActiveInference2020}. A complementary approach is to assess actions, instead of action sequences, by conditioning future actions to be optimal in the sense that they minimise the expected free energy~\cite{fristonSophisticatedInference2021,dacostaRelationshipDynamicProgramming2020a}. This idea leads to a backward form of planning, where the agent plans for the best action at the last time-step, followed by the best action at the penultimate time-step, and so on, until the present. Crucially, it leads to smarter agents~\cite{dacostaRelationshipDynamicProgramming2020a,fristonSophisticatedInference2021} whose computational complexity scales linearly (as opposed to exponentially) in the length of action sequences~\cite{paulActiveInferenceStochastic}. %Planning can be amortised by borrowing techniques from value iteration [Contrastive AIF, Millidge variational policy gradients]

Scalable inference methods~\cite{zhangAdvancesVariationalInference2017} can be used to make active inference more efficient~\cite{vandelaarSimulatingActiveInference2019}. For example, we can train neural networks to predict the various posterior distributions, including the posterior over actions~\cite{fountasDeepActiveInference2020,sajidMixedGenerativeModel2022,millidgeDeepActiveInference2020}. While training, the output of the neural network can be used as an initial conditions for variational inference~\cite{tschantzControlHybridInference2020}, resulting in accurate inferences whose computational cost decrease as the network learns.
Additionally, optimising free energy reduces to efficient \textit{message passing} schemes, when one imposes certain simplifying restrictions to the family of candidate distributions~\cite{winnVariationalMessagePassing2005,wainwrightGraphicalModelsExponential2007,parrNeuronalMessagePassing2019,schwobelActiveInferenceBelief2018,championRealizingActiveInference2021}.

%The discrete-time implementation of active inference discussed here may prove computationally expensive in continuous state spaces. Thankfully, 
A much cheaper implementation of active inference exists for continuous states evolving in continuous time. The method frames perception and decision-making as variational inference, by simulating a gradient flow on free energy in an extended state space~\cite{fristonActionBehaviorFreeenergy2010,fristonFreeEnergyPrinciple2022}. %This can be derived from a similar set of principles~\cite{fristonFreeEnergyPrinciple2022}.
Furthermore, it can be combined with discrete active inference to operate efficiently in generative models combining discrete and continuous states~\cite{fristonGraphicalBrainBelief2017}. As an example, high-dimensional observations in the continuous domain (e.g., speech) processed through continuous active inference are converted into discrete, abstract representations (e.g., semantics)~\cite{sajidMixedGenerativeModel2022}. Based on these representations, the agent makes high-level, categorical decisions (e.g., ``I want to move over there''), which contextualise low-level, continuous actions (e.g., the continuous motion of a limb towards the goal location)~\cite{parrComputationalNeurologyMovement2021}.

\subsection*{Acknowledgements}
The authors thank Noor Sajid, Samuel Tenka, Zafeiros Fountas and Panagiotis Tigas for helpful discussions on adaptive agents.
LD is supported by the Fonds National de la Recherche, Luxembourg (Project code: 13568875). KF is supported by funding for the Wellcome Centre for Human Neuroimaging (Ref: 205103/Z/16/Z) and a Canada-UK Artificial Intelligence Initiative (Ref: ES/T01279X/1). GAP was partially supported by JPMorgan Chase \& Co under J.P. Morgan A.I. Research Awards in 2019 and 2021 and by the EPSRC, grant number EP/P031587/1. This publication is based on work partially supported by the EPSRC Centre for Doctoral Training in Mathematics of Random Systems: Analysis, Modelling and Simulation (EP/S023925/1).
AB, GF, MG and MIJ thank the
support of the Army Research Office (ARO) under contract W911NF-17-1-0304
as part of the collaboration between US DOD, UK MOD and UK Engineering and Physical Research Council (EPSRC) under the Multidisciplinary University Research Initiative (MURI).

\bibliographystyle{unsrt}

{\footnotesize
\bibliography{main.bib}}

\begin{thebibliography}{100}

\bibitem{mclachlan2002splitting}
R.~I. Mc{L}achlan and G.~R.~W. Quispel.
\newblock Splitting methods.
\newblock {\em Acta Numer.}, 11:341, 2002.

\bibitem{HairerBook}
E.~Hairer, C.~Lubich, and G.~Wanner.
\newblock {\em Geometric Numerical Integration: Structure-Preserving Algorithms
  for Ordinary Differential Equations}.
\newblock Springer, 2010.

\bibitem{leimkuhler2004simulating}
B.~Leimkuhler and S.~Reich.
\newblock {\em Simulating {H}amiltonian Dynamics}.
\newblock Cambridge University Press, 2004.

\bibitem{Celledoni:2014}
E.~Celledoni, H.~Marthinsen, and B.~Owren.
\newblock An introduction to {L}ie group integrators: basics, new developments
  and applications.
\newblock {\em J. Comput. Phys.}, 257:1040--1061, 2014.

\bibitem{marsden2001discrete}
J.~E. Marsden and M.~West.
\newblock Discrete mechanics and variational integrators.
\newblock {\em Acta Numer.}, 10:357--514, 2001.

\bibitem{betancourt2018symplectic}
M.~Betancourt, M.~I. Jordan, and A.~Wilson.
\newblock On symplectic optimization.
\newblock 2018.
\newblock arXiv:1802.03653 [stat.CO].

\bibitem{bravetti2019optimization}
A.~Bravetti, M.~L. Daza-Torres, H.~Flores-Arguedas, and M.~Betancourt.
\newblock Optimization algorithms inspired by the geometry of dissipative
  systems.
\newblock 2019.
\newblock arXiv:1912.02928 [math.OC].

\bibitem{Franca:2020}
G.~Fran{\c{c}}a, J.~Sulam, D.~P. Robinson, and R.~Vidal.
\newblock Conformal symplectic and relativistic optimization.
\newblock {\em J. Stat. Mech.}, 2020(12):124008, 2020.

\bibitem{Franca:2021}
G.~Fran{\c{c}}a, M.~I. Jordan, and R.~Vidal.
\newblock On dissipative symplectic integration with applications to
  gradient-based optimization.
\newblock {\em J. Stat. Mech.}, 2021(4):043402, 2021.

\bibitem{FrancaBarp:2021}
G.~Fran{\c{c}}a, A.~Barp, M.~Girolami, and M.~I. Jordan.
\newblock Optimization on manifolds: A symplectic approach.
\newblock 2021.
\newblock arXiv:2107.11231 [cond-mat.stat-mech].

\bibitem{Alimisis21}
F.~Alimisis, A.~Orvieto, G.~Becigneul, and A.~Lucchi.
\newblock Momentum improves optimization on {R}iemannian manifolds.
\newblock {\em Int. Conf. Artificial Intelligence and Stats.}, 130:1351--1359,
  2021.

\bibitem{rousset2010free}
M.~Rousset, G.~Stoltz, and T.~Lelievre.
\newblock {\em Free Energy Computations: A Mathematical Perspective}.
\newblock World Scientific, 2010.

\bibitem{Duane:1987}
S.~Duane, A.D. Kennedy, B.~J. Pendleton, and D.~Roweth.
\newblock Hybrid {M}onte {C}arlo.
\newblock {\em Phys. Lett. B}, 195(2):216--222, 1987.

\bibitem{betancourt2017geometric}
M.~Betancourt, S.~Byrne, S.~Livingstone, and M.~Girolami.
\newblock The geometric foundations of {H}amiltonian {M}onte {C}arlo.
\newblock {\em Bernoulli}, 23(4A):2257--2298, 2017.

\bibitem{livingstone2019geometric}
S.~Livingstone, M.~Betancourt, S.~Byrne, and M.~Girolami.
\newblock On the geometric ergodicity of {H}amiltonian {M}onte {C}arlo.
\newblock {\em Bernoulli}, 25(4A):3109--3138, 2019.

\bibitem{barp2019hamiltonian}
A.~Barp, A.~Kennedy, and M.~Girolami.
\newblock Hamiltonian {M}onte {C}arlo on symmetric and homogeneous spaces via
  symplectic reduction.
\newblock 2019.

\bibitem{Celledoni:2021}
E.~Celledoni, M.~J. Ehrhardt, C.~Etmann, R.~I. McLachlan, B.~Owren, C.~B.
  Schonlieb, and F.~Sherry.
\newblock Structure-preserving deep learning.
\newblock {\em Eur. J. Applied Math.}, 32(5):888--936, 2021.

\bibitem{bronstein2021geometric}
M.~M. Bronstein, J.~Bruna, T.~Cohen, and P.~Velickovic.
\newblock Geometric deep learning: Grids, groups, graphs, geodesics, and
  gauges, 2021.
\newblock arXiv:2104.13478 [cs.LG].

\bibitem{vaillant2005surface}
M.~Vaillant and J.~Glaunes.
\newblock Surface matching via currents.
\newblock In {\em Biennial Int. Conf. Information Processing in Medical
  Imaging}, pages 381--392. Springer, 2005.

\bibitem{durrleman2009statistical}
S.~Durrleman, X.~Pennec, A.~Trouv{\'e}, and N.~Ayache.
\newblock Statistical models of sets of curves and surfaces based on currents.
\newblock {\em Medical Image Analysis}, 13(5):793--808, 2009.

\bibitem{barpRiemannSteinKernelMethod2020}
A.~Barp, Chris~J. Oates, E.~Porcu, and M.~Girolami.
\newblock A {{Riemann-Stein Kernel Method}}.
\newblock 2020.
\newblock arXiv:1810.04946 [math, stat].

\bibitem{harms2020geometry}
P.~Harms, P.~W. Michor, X.~Pennec, and S.~Sommer.
\newblock Geometry of sample spaces.
\newblock 2020.
\newblock arXiv:2010.08039.

\bibitem{rao1992information}
C.~R. Rao.
\newblock Information and the accuracy attainable in the estimation of
  statistical parameters.
\newblock In {\em Breakthroughs in Statistics}, pages 235--247. Springer, 1992.

\bibitem{jeffreys1946invariant}
H.~Jeffreys.
\newblock An invariant form for the prior probability in estimation problems.
\newblock {\em Proc. Royal Soc. London. Series A. Math. and Phys. Sci.},
  186(1007):453--461, 1946.

\bibitem{amari2016information}
S.~Amari.
\newblock {\em Information geometry and its applications}, volume 194.
\newblock Springer, 2016.

\bibitem{ay2017information}
N.~Ay, J.~Jost, H.~V{\^a}n~L{\^e}, and L.~Schwachh{\"o}fer.
\newblock {\em Information geometry}, volume~64.
\newblock Springer, 2017.

\bibitem{nielsen2020elementary}
F.~Nielsen.
\newblock An elementary introduction to information geometry.
\newblock {\em Entropy}, 22(10):1100, 2020.

\bibitem{amari2012differential}
S.~Amari.
\newblock {\em Differential-geometrical methods in statistics}, volume~28.
\newblock Springer Science \& Business Media, 2012.

\bibitem{chentsov1965categories}
N.~N. Chentsov.
\newblock Categories of mathematical statistics.
\newblock {\em Uspekhi Matematicheskikh Nauk}, 20(4):194--195, 1965.

\bibitem{jost2021probabilistic}
J.~Jost, H.~V. L{\^e}, and T.~D. Tran.
\newblock Probabilistic morphisms and bayesian nonparametrics.
\newblock {\em Eur. Phys. J. Plus}, 136(4):1--29, 2021.

\bibitem{Polyak:1964}
B.~T. Polyak.
\newblock Some methods of speeding up the convergence of iteration methods.
\newblock {\em {USSR Comp. Math. and Math. Phys.}}, 4(5):1--17, 1964.

\bibitem{Nesterov:1983}
Y.~Nesterov.
\newblock A method of solving a convex programming problem with convergence
  rate {$O(1/k^2)$}.
\newblock {\em Soviet Math. Doklady}, 27(2):372--376, 1983.

\bibitem{Candes:2016}
W.~Su, S.~Boyd, and E.~J. Cand{{\`e}}s.
\newblock A differential equation for modeling {N}esterov's accelerated
  gradient method: Theory and insights.
\newblock {\em J. Mach. Learn. Res.}, 17(153):1--43, 2016.

\bibitem{Wibisono:2016}
A.~Wibisono, A.~C. Wilson, and M.~I. Jordan.
\newblock A variational perspective on accelerated methods in optimization.
\newblock {\em {Proc. Nat. Acad. Sci.}}, 113(47):E7351--E7358, 2016.

\bibitem{Wilson:2021}
A.~Wilson, B.~Recht, and M.~I. Jordan.
\newblock A {L}yapunov analysis of accelerated methods in optimization.
\newblock {\em {J. Mach. Learn. Res.}}, 22:1--34, 2021.

\bibitem{Franca:2018a}
G.~Franca, D.~Robinson, and R.~Vidal.
\newblock {ADMM} and accelerated {ADMM} as continuous dynamical systems.
\newblock 80:1559--1567, 2018.

\bibitem{Franca:2018b}
G.~Fran{\c c}a, D.~P. Robinson, and R.~Vidal.
\newblock A nonsmooth dynamical systems perspective on accelerated extensions
  of {ADMM}.
\newblock 2018.
\newblock arXiv:1808.04048 [math.OC].

\bibitem{Franca:2021b}
G.~Fran{\c{c}}a, D.~P. Robinson, and R.~Vidal.
\newblock Gradient flows and proximal splitting methods: A unified view on
  accelerated and stochastic optimization.
\newblock {\em Phys. Rev. E}, 103:053304, 2021.

\bibitem{muehlebach2021optimization}
M.~Muehlebach and M.~I. Jordan.
\newblock Optimization with momentum: Dynamical, control-theoretic, and
  symplectic perspectives.
\newblock {\em J. Mach. Learn. Res.}, 22(73):1--50, 2021.

\bibitem{muehlebach2021constrained}
M.~Muehlebach and M.~I. Jordan.
\newblock On constraints in first-order optimization: A view from non-smooth
  dynamical systems.
\newblock 2021.
\newblock arXiv:2107.08225, [math.OC].

\bibitem{Takahashi:1984}
M.~Takahashi and M.~Imada.
\newblock {Monte {C}arlo calculation of quantum systems. II. Higher order
  correction}.
\newblock {\em J. Phys. Soc. Jpn.}, 53:3765–--3769, 1984.

\bibitem{Suzuki:1990}
M.~Suzuki.
\newblock Fractal decomposition of exponential operators with applications to
  many-body theories and {M}onte {C}arlo simulations.
\newblock {\em {Phys. Lett. A}}, 146:319--323, 1990.

\bibitem{Yoshida:1990}
H.~Yoshida.
\newblock {Construction of higher order symplectic integrators}.
\newblock {\em {Phys. Lett. A}}, 150(5):262--268, 1990.

\bibitem{SanzSerna:1992}
J.~M. Sanz-Serna.
\newblock Symplectic integrators for {H}amiltonian problems: An overview.
\newblock {\em {Acta Numerica}}, 1:243--286, 1992.

\bibitem{Benettin:1994}
G.~Benettin and A.~Giorgilli.
\newblock On the {H}amiltonian interpolation of near-to-the-identity symplectic
  mappings with application to symplectic integration algorithms.
\newblock {\em {J. Stat. Phys.}}, 74:1117--1143, 1994.

\bibitem{McLachlan:2006}
R.~I. McLachlan and G.~R.~W. Quispel.
\newblock Geometric integrators for {ODE}s.
\newblock {\em {J. Phys. A: Math. Gen.}}, 39:5251--5285, 2006.

\bibitem{Forest:2006}
E.~Forest.
\newblock {Geometric integration for particle accelerators}.
\newblock {\em {J. Phys. A: Math. Gen.}}, 39:5321--–5377, 2006.

\bibitem{Kennedy:2013}
A.~D. Kennedy, P.~J. Silva, and M.~A. Clark.
\newblock Shadow {H}amiltonians, {P}oisson brackets, and gauge theories.
\newblock {\em Phys. Rev. D}, 87:034511, 2013.

\bibitem{Neal:2011}
R.~M. Neal.
\newblock {MCMC} using {H}amiltonian dynamics.
\newblock In {\em Handbook of Markov Chain Monte Carlo}. Chapman and
  Hall/{CRC}, 2011.

\bibitem{Otto:2001}
F.~Otto.
\newblock The geometry of dissipative evolution equations: the porous medium
  equation.
\newblock {\em Comm. Partial Differential Equations}, 26:101--174, 2001.

\bibitem{Jordan:1998}
R.~Jordan, D.~Kinderlehrer, and F.~Otto.
\newblock The variational formulation of the {F}okker-{P}lanck equation.
\newblock {\em SIAM J. Math. Anal.}, 29(1):1--17, 1998.

\bibitem{neal2003slice}
R.~M. Neal.
\newblock Slice sampling.
\newblock {\em The annals of statistics}, 31(3):705--767, 2003.

\bibitem{murray2010elliptical}
I.~Murray, R.~Adams, and D.~MacKay.
\newblock Elliptical slice sampling.
\newblock In {\em Int. Conf. Artificial Intelligence and Stats.}, pages
  541--548. JMLR Workshop and Conference Proceedings, 2010.

\bibitem{davis1984piecewise}
Mark~HA Davis.
\newblock Piecewise-deterministic markov processes: A general class of
  non-diffusion stochastic models.
\newblock {\em Journal of the Royal Statistical Society: Series B
  (Methodological)}, 46(3):353--376, 1984.

\bibitem{bouchard2018bouncy}
A.~Bouchard-C{\^o}t{\'e}, S.~J. Vollmer, and A.~Doucet.
\newblock The bouncy particle sampler: A nonreversible rejection-free markov
  chain monte carlo method.
\newblock {\em J. Amer. Stats. Assoc.}, 113(522):855--867, 2018.

\bibitem{vanetti2017piecewise}
P.~Vanetti, A.~Bouchard-C{\^o}t{\'e}, G.~Deligiannidis, and A.~Doucet.
\newblock Piecewise-deterministic {M}arkov {C}hain {M}onte {C}arlo.
\newblock 2017.
\newblock arXiv:1707.05296.

\bibitem{bierkens2019zig}
J.~Bierkens, P.~Fearnhead, and G.~Roberts.
\newblock The zig-zag process and super-efficient sampling for {B}ayesian
  analysis of big data.
\newblock {\em Ann. Stats.}, 47(3):1288--1320, 2019.

\bibitem{bierkens2017piecewise}
J.~Bierkens and G.~Roberts.
\newblock A piecewise deterministic scaling limit of lifted
  metropolis--hastings in the curie--weiss model.
\newblock {\em Ann. App. Prob.}, 27(2):846--882, 2017.

\bibitem{peters2012rejection}
E.~A. J.~F. Peters and G.~de~With.
\newblock Rejection-free monte carlo sampling for general potentials.
\newblock {\em Phys. Rev. E}, 85(2):026703, 2012.

\bibitem{roberts1996exponential}
G.~O. Roberts and R.~L. Tweedie.
\newblock Exponential convergence of {L}angevin distributions and their
  discrete approximations.
\newblock {\em Bernoulli}, pages 341--363, 1996.

\bibitem{durmus2017nonasymptotic}
A.~Durmus and E.~Moulines.
\newblock Nonasymptotic convergence analysis for the unadjusted {L}angevin
  algorithm.
\newblock {\em Ann. App. Prob.}, 27(3):1551--1587, 2017.

\bibitem{durmus2018efficient}
A.~Durmus, E.~Moulines, and M.~Pereyra.
\newblock Efficient bayesian computation by proximal markov chain monte carlo:
  when langevin meets moreau.
\newblock {\em SIAM J. on Imaging Sci.}, 11(1):473--506, 2018.

\bibitem{garbuno-inigoInteractingLangevinDiffusions2019}
A.~{Garbuno-Inigo}, F.~Hoffmann, W.~Li, and A.~M. Stuart.
\newblock Interacting {{Langevin diffusions}}: {{gradient structure And
  ensemble Kalman sampler}}.
\newblock 2019.
\newblock arXiv:1903.08866 [math].

\bibitem{betancourt2017conceptual}
M.~Betancourt.
\newblock A conceptual introduction to hamiltonian monte carlo.
\newblock 2017.
\newblock arXiv:1701.02434.

\bibitem{barp2018geometry}
A.~Barp, F-X. Briol, A.~D. Kennedy, and M.~Girolami.
\newblock Geometry and dynamics for {M}arkov {C}hain {M}onte {C}arlo.
\newblock {\em Annual Rev. Stats. and its App.}, 5:451--471, 2018.

\bibitem{metropolis1953equation}
N.~R. Metropolis, A.~W. Rosenbluth, M.~N. Rosenbluth, A.~H. Teller, and
  E.~Teller.
\newblock Equation of state calculations by fast computing machines.
\newblock {\em J. Chem. Phys.}, 21(6):1087--1092, 1953.

\bibitem{hastings1970monte}
W.~K. Hastings.
\newblock {\em Monte {C}arlo Sampling Methods Using {M}arkov Chains and their
  Applications}.
\newblock Oxford University Press, 1970.

\bibitem{alder1959studies}
B.~J. Alder and T.~E. Wainwright.
\newblock Studies in molecular dynamics. {I}. general method.
\newblock {\em J. Chem. Phys.}, 31(2):459--466, 1959.

\bibitem{van2000asymptotic}
A.~W. Van~der Vaart.
\newblock {\em Asymptotic Statistics}, volume~3.
\newblock Cambridge university press, 2000.

\bibitem{vapnik1999nature}
V.~Vapnik.
\newblock {\em The Nature of Statistical Learning Theory}.
\newblock Springer Science \& Business Media, 1999.

\bibitem{hyvarinen2005estimation}
A.~Hyv{\"a}rinen and P.~Dayan.
\newblock Estimation of non-normalized statistical models by score matching.
\newblock {\em J. Mach. Learn. Res.}, 6(4), 2005.

\bibitem{parry2012proper}
M.~Parry, A.~P. Dawid, and S.~Lauritzen.
\newblock Proper local scoring rules.
\newblock {\em Ann. Stats.}, 40(1):561--592, 2012.

\bibitem{villani2009optimal}
C.~Villani.
\newblock {\em Optimal Transport: Old and New}.
\newblock Springer, 2009.

\bibitem{bassetti2006minimum}
F.~Bassetti, A.~Bodini, and E.~Regazzini.
\newblock On minimum {K}antorovich distance estimators.
\newblock {\em Stats. \& Prob. Lett.}, 76(12):1298--1302, 2006.

\bibitem{peyre2019computational}
G.~Peyr{\'e}, M.~Cuturi, et~al.
\newblock Computational optimal transport: With applications to data science.
\newblock {\em Found. Trends Mach. Learn.}, 11(5-6):355--607, 2019.

\bibitem{cuturi2013sinkhorn}
M.~Cuturi.
\newblock Sinkhorn distances: Lightspeed computation of optimal transport.
\newblock {\em NeurIPS}, 26:2292--2300, 2013.

\bibitem{bergerStatisticalDecisionTheory1985}
J.~O. Berger.
\newblock {\em Statistical {{Decision Theory}} and {{Bayesian Analysis}}}.
\newblock Springer {{Series}} in {{Statistics}}. {Springer-Verlag}, {New York},
  second edition, 1985.

\bibitem{vonneumannTheoryGamesEconomic1944}
J.~Von~Neumann and O.~Morgenstern.
\newblock {\em Theory of Games and Economic Behavior}.
\newblock {Princeton University Press}, 1944.

\bibitem{bellmanAppliedDynamicProgramming2015}
R.~E. Bellman and S.~E. Dreyfus.
\newblock {\em Applied {{Dynamic Programming}}}.
\newblock {Princeton University Press}, 2015.

\bibitem{bartoReinforcementLearningIntroduction1992}
A.~Barto and R.~Sutton.
\newblock {\em Reinforcement {{Learning}}: {{An Introduction}}}.
\newblock A Bradford Book, 1992.

\bibitem{dacostaActiveInferenceDiscrete2020}
L.~Da~Costa, T.~Parr, N.~Sajid, S.~Veselic, V.~Neacsu, and K.~Friston.
\newblock Active inference on discrete state-spaces: {{A}} synthesis.
\newblock {\em J. Math. Psychology}, 99:102447, 2020.

\bibitem{fristonActionBehaviorFreeenergy2010}
K.~J. Friston, J.~Daunizeau, J.~Kilner, and S.~J. Kiebel.
\newblock Action and behavior: A free-energy formulation.
\newblock {\em Biological Cybernetics}, 102(3):227--260, 2010.

\bibitem{fristonFreeenergyPrincipleUnified2010}
K.~Friston.
\newblock The free-energy principle: A unified brain theory?
\newblock {\em Nature Reviews Neuroscience}, 11(2):127--138, 2010.

\bibitem{fristonActiveInferenceEpistemic2015}
Karl Friston, Francesco Rigoli, Dimitri Ognibene, Christoph Mathys, Thomas
  Fitzgerald, and Giovanni Pezzulo.
\newblock Active inference and epistemic value.
\newblock {\em Cognitive Neuroscience}, 6(4):187--214, October 2015.

\bibitem{lanillosActiveInferenceRobotics2021}
Pablo Lanillos, Cristian Meo, Corrado Pezzato, Ajith~Anil Meera, Mohamed
  Baioumy, Wataru Ohata, Alexander Tschantz, Beren Millidge, Martijn Wisse,
  Christopher~L. Buckley, and Jun Tani.
\newblock Active {{Inference}} in {{Robotics}} and {{Artificial Agents}}:
  {{Survey}} and {{Challenges}}.
\newblock {\em arXiv:2112.01871 [cs]}, December 2021.

\bibitem{dacostaHowActiveInference2022a}
Lancelot Da~Costa, Pablo Lanillos, Noor Sajid, Karl Friston, and Shujhat Khan.
\newblock How {{Active Inference Could Help Revolutionise Robotics}}.
\newblock {\em Entropy}, 24(3):361, March 2022.

\bibitem{silverMasteringGameGo2016}
David Silver, Aja Huang, Chris~J. Maddison, Arthur Guez, Laurent Sifre, George
  {van den Driessche}, Julian Schrittwieser, Ioannis Antonoglou, Veda
  Panneershelvam, Marc Lanctot, Sander Dieleman, Dominik Grewe, John Nham, Nal
  Kalchbrenner, Ilya Sutskever, Timothy Lillicrap, Madeleine Leach, Koray
  Kavukcuoglu, Thore Graepel, and Demis Hassabis.
\newblock Mastering the game of {{Go}} with deep neural networks and tree
  search.
\newblock {\em Nature}, 529(7587):484--489, 2016.

\bibitem{millidgeDeepActiveInference2020}
B.~Millidge.
\newblock Deep active inference as variational policy gradients.
\newblock {\em J. Math. Psychology}, 96:102348, 2020.

\bibitem{vanderhimstDeepActiveInference2020}
O.~{van der Himst} and P.~Lanillos.
\newblock Deep {{Active Inference}} for {{Partially Observable MDPs}}.
\newblock 2020.
\newblock arXiv:2009.03622 [cs, stat].

\bibitem{cullenActiveInferenceOpenAI2018}
Maell Cullen, Ben Davey, Karl~J. Friston, and Rosalyn~J. Moran.
\newblock Active {{Inference}} in {{OpenAI Gym}}: {{A Paradigm}} for
  {{Computational Investigations Into Psychiatric Illness}}.
\newblock {\em Biological Psychiatry: Cognitive Neuroscience and Neuroimaging},
  3(9):809--818, September 2018.

\bibitem{sajidActiveInferenceDemystified2021}
Noor Sajid, Philip~J. Ball, Thomas Parr, and Karl~J. Friston.
\newblock Active {{Inference}}: {{Demystified}} and {{Compared}}.
\newblock {\em Neural Computation}, 33(3):674--712, January 2021.

\bibitem{markovicEmpiricalEvaluationActive2021a}
Dimitrije Markovi{\'c}, Hrvoje Stoji{\'c}, Sarah Schw{\"o}bel, and Stefan~J.
  Kiebel.
\newblock An empirical evaluation of active inference in multi-armed bandits.
\newblock {\em Neural Networks}, 144:229--246, December 2021.

\bibitem{paulActiveInferenceStochastic2021}
Aswin Paul, Noor Sajid, Manoj Gopalkrishnan, and Adeel Razi.
\newblock Active {{Inference}} for {{Stochastic Control}}.
\newblock {\em arXiv:2108.12245 [cs]}, August 2021.

\bibitem{mazzagliaContrastiveActiveInference2021}
Pietro Mazzaglia, Tim Verbelen, and Bart Dhoedt.
\newblock Contrastive {{Active Inference}}.
\newblock In {\em Advances in {{Neural Information Processing Systems}}}, May
  2021.

\bibitem{Hansen:2011}
A.~C. Hansen.
\newblock A theoretical framework for backward error analysis on manifolds.
\newblock {\em J. Geom. Mech.}, 3(1):81--111, 2011.

\bibitem{mclachlan1999geometric}
R.~I. McLachlan, G~R.~W. Quispel, and N.~Robidoux.
\newblock Geometric integration using discrete gradients.
\newblock {\em Philosophical Transactions of the Royal Society of London.
  Series A: Mathematical, Physical and Engineering Sciences},
  357(1754):1021--1045, 1999.

\bibitem{McLachlan:2001}
R.~McLachlan and M.~Perlmutter.
\newblock Conformal {H}amiltonian systems.
\newblock {\em J. Geom. and Phys.}, 39:276--300, 2001.

\bibitem{marthinsen2014geometric}
H.~Marthinsen and B.~Owren.
\newblock Geometric integration of non-autonomous {H}amiltonian problems.
\newblock {\em Adv. Comput. Math.}, 42:313--332, 2016.

\bibitem{asorey1983generalized}
M.~Asorey, J.~F. Carinena, and L.~A. Ibort.
\newblock Generalized canonical transformations for time-dependent systems.
\newblock {\em J. Math. Phys.}, 24(12):2745--2750, 1983.

\bibitem{Andersen:1983}
H.~C. Andersen.
\newblock Rattle: A ``velocity'' version of the shake algorithm for molecular
  dynamics calculations.
\newblock {\em J. Comput. Phys.}, 52(1):24--34, 1983.

\bibitem{Leimkuhler:1994}
B.~J. Leimkuhler and R.~D. Skeel.
\newblock Symplectic numerical integrators in constrained {H}amiltonian
  systems.
\newblock {\em J. Comput. Phys.}, 112:117--125, 1994.

\bibitem{McLachlan:2014}
R.~I. Mc{L}achlan, K.~Modin, O.~Verdier, and M.~Wilkins.
\newblock Geometric generalizations of {SHAKE} and {RATTLE}.
\newblock {\em Found. Comput. Math.}, (14):339--370, 2014.

\bibitem{leimkuhler2016efficient}
B.~Leimkuhler and C.~Matthews.
\newblock Efficient molecular dynamics using geodesic integration and
  solvent-solute splitting.
\newblock {\em Proc. Royal Soc. A: Math., Phys. and Eng. Sci.},
  472(2189):20160138, 2016.

\bibitem{Sra:2016}
H.~Zhang and S.~Sra.
\newblock First-order methods for geodesically convex optimization.
\newblock {\em Conf. Learning Theory}, pages 1617--1638, 2016.

\bibitem{ambrosioGradientFlowsMetric2005}
Luigi Ambrosio, Nicola Gigli, and Giuseppe Savare.
\newblock {\em Gradient {{Flows}}: {{In Metric Spaces}} and in the {{Space}} of
  {{Probability Measures}}}.
\newblock {Springer Science \& Business Media}, January 2005.

\bibitem{ottobreMarkovChainMonte2016}
Michela Ottobre.
\newblock Markov {{Chain Monte Carlo}} and {{Irreversibility}}.
\newblock {\em Reports on Mathematical Physics}, 77:267--292, June 2016.

\bibitem{pavliotisStochasticProcessesApplications2014}
Grigorios~A. Pavliotis.
\newblock {\em Stochastic Processes and Applications: Diffusion Processes, the
  {{Fokker-Planck}} and {{Langevin}} Equations}.
\newblock Number volume 60 in Texts in Applied Mathematics. {Springer}, {New
  York}, 2014.

\bibitem{villaniHypocoercivity2009}
C{\'e}dric Villani.
\newblock {\em Hypocoercivity}, volume 202 of {\em Memoirs of the {{American
  Mathematical Society}}}.
\newblock {American Mathematical Society}, November 2009.

\bibitem{Ma:2019}
Yi-An Ma, Niladri Chatterji, Xiang Cheng, Nicolas Flammarion, Peter Bartlett,
  and Michael~I. Jordan.
\newblock Is there an analog of nesterov acceleration for mcmc?, 2019.

\bibitem{mackay2003information}
D.~J.~C. MacKay.
\newblock {\em Information theory, inference and learning algorithms}.
\newblock Cambridge university press, 2003.

\bibitem{hairerErgodicPropertiesMarkov2018}
Martin Hairer.
\newblock Ergodic {{Properties}} of {{Markov Processes}}.
\newblock 2018.

\bibitem{barpUnifyingCanonicalDescription2021}
Alessandro Barp, So~Takao, Michael Betancourt, Alexis Arnaudon, and Mark
  Girolami.
\newblock A {{Unifying}} and {{Canonical Description}} of {{Measure-Preserving
  Diffusions}}.
\newblock {\em arXiv:2105.02845 [math, stat]}, May 2021.

\bibitem{hormanderHypoellipticSecondOrder1967}
Lars H{\"o}rmander.
\newblock Hypoelliptic second order differential equations.
\newblock {\em Acta Mathematica}, 119:147--171, 1967.

\bibitem{bismut1981martingales}
J.M. Bismut.
\newblock Martingales, the malliavin calculus and hypoellipticity under general
  h{\"o}rmander's conditions.
\newblock {\em Zeitschrift f{\"u}r Wahrscheinlichkeitstheorie und verwandte
  Gebiete}, 56(4):469--505, 1981.

\bibitem{maCompleteRecipeStochastic2015}
Yi-An Ma, Tianqi Chen, and Emily~B. Fox.
\newblock A {{Complete Recipe}} for {{Stochastic Gradient MCMC}}.
\newblock {\em arXiv:1506.04696 [math, stat]}, October 2015.

\bibitem{hwangAcceleratingDiffusions2005}
Chii-Ruey Hwang, Shu-Yin {Hwang-Ma}, and Shuenn-Jyi Sheu.
\newblock Accelerating diffusions.
\newblock {\em The Annals of Applied Probability}, 15(2):1433--1444, May 2005.

\bibitem{chafaiEntropiesConvexityFunctional2004}
Djalil Chafa{\"i}.
\newblock Entropies, convexity, and functional inequalities, {{On}}
  \$\textbackslash{{Phi}} \$-entropies and \$\textbackslash{{Phi}}
  \$-{{Sobolev}} inequalities.
\newblock {\em Journal of Mathematics of Kyoto University}, 44(2):325--363,
  January 2004.

\bibitem{joulinCurvatureConcentrationError2010}
Ald{\'e}ric Joulin and Yann Ollivier.
\newblock Curvature, concentration and error estimates for {{Markov}} chain
  {{Monte Carlo}}.
\newblock {\em The Annals of Probability}, 38(6):2418--2442, November 2010.

\bibitem{rey-belletIrreversibleLangevinSamplers2015}
Luc {Rey-Bellet} and Kostantinos Spiliopoulos.
\newblock Irreversible {{Langevin}} samplers and variance reduction: A large
  deviation approach.
\newblock {\em Nonlinearity}, 28(7):2081--2103, July 2015.

\bibitem{duncanVarianceReductionUsing2016}
A.~B. Duncan, T.~Leli{\`e}vre, and G.~A. Pavliotis.
\newblock Variance {{Reduction Using Nonreversible Langevin Samplers}}.
\newblock {\em Journal of Statistical Physics}, 163(3):457--491, May 2016.

\bibitem{haussmannTimeReversalDiffusions1986}
U.~G. Haussmann and E.~Pardoux.
\newblock Time {{Reversal}} of {{Diffusions}}.
\newblock {\em Annals of Probability}, 14(4):1188--1205, October 1986.

\bibitem{nealImprovingAsymptoticVariance2004}
Radford~M. Neal.
\newblock Improving {{Asymptotic Variance}} of {{MCMC Estimators}}:
  {{Non-reversible Chains}} are {{Better}}.
\newblock {\em arXiv:math/0407281}, July 2004.

\bibitem{lelievreOptimalNonreversibleLinear2013}
Tony Leli{\`e}vre, Francis Nier, and Grigorios~A. Pavliotis.
\newblock Optimal non-reversible linear drift for the convergence to
  equilibrium of a diffusion.
\newblock {\em Journal of Statistical Physics}, 152(2):237--274, July 2013.

\bibitem{wuAttainingOptimalGaussian2014}
Sheng-Jhih Wu, Chii-Ruey Hwang, and Moody~T. Chu.
\newblock Attaining the {{Optimal Gaussian Diffusion Acceleration}}.
\newblock {\em Journal of Statistical Physics}, 155:571--590, May 2014.

\bibitem{zhangGeometryinformedIrreversiblePerturbations2021}
Benjamin~J. Zhang, Youssef~M. Marzouk, and Konstantinos Spiliopoulos.
\newblock Geometry-informed irreversible perturbations for accelerated
  convergence of {{Langevin}} dynamics.
\newblock {\em arXiv:2108.08247 [math, stat]}, August 2021.

\bibitem{miraOrderingImprovingPerformance2001}
Antonietta Mira.
\newblock Ordering and {{Improving}} the {{Performance}} of {{Monte Carlo
  Markov Chains}}.
\newblock {\em Statistical Science}, 16(4):340--350, November 2001.

\bibitem{girolami2011riemann}
M.~Girolami and B.~Calderhead.
\newblock Riemann manifold langevin and hamiltonian monte carlo methods.
\newblock {\em Journal of the Royal Statistical Society: Series B (Statistical
  Methodology)}, 73(2):123--214, 2011.

\bibitem{abdulleAcceleratedConvergenceEquilibrium2019}
Assyr Abdulle, Grigorios~A. Pavliotis, and Gilles Vilmart.
\newblock Accelerated convergence to equilibrium and reduced asymptotic
  variance for {{Langevin}} dynamics using {{Stratonovich}} perturbations.
\newblock {\em Comptes Rendus Mathematique}, 357(4):349--354, April 2019.

\bibitem{helfferRemarksDecayCorrelations1998}
Bernard Helffer.
\newblock Remarks on {{Decay}} of {{Correlations}} and {{Witten Laplacians
  Brascamp}}\textendash{{Lieb Inequalities}} and {{Semiclassical Limit}}.
\newblock {\em Journal of Functional Analysis}, 155(2):571--586, June 1998.

\bibitem{saumardLogconcavityStrongLogconcavity2014}
Adrien Saumard and Jon~A. Wellner.
\newblock Log-concavity and strong log-concavity: {{A}} review.
\newblock {\em Statistics Surveys}, 8(none):45--114, January 2014.

\bibitem{guillinOptimalLinearDrift2021}
Arnaud Guillin and Pierre Monmarch{\'e}.
\newblock Optimal linear drift for the speed of convergence of an hypoelliptic
  diffusion.
\newblock {\em arXiv:1604.07295 [math]}, October 2021.

\bibitem{maThereAnalogNesterov2019}
Yi-An Ma, Niladri Chatterji, Xiang Cheng, Nicolas Flammarion, Peter Bartlett,
  and Michael~I. Jordan.
\newblock Is {{There}} an {{Analog}} of {{Nesterov Acceleration}} for {{MCMC}}?
\newblock {\em arXiv:1902.00996 [cs, math, stat]}, October 2019.

\bibitem{chakOptimalFrictionMatrix2021}
Martin Chak, Nikolas Kantas, Tony Leli{\`e}vre, and Grigorios~A Pavliotis.
\newblock Optimal friction matrix for underdamped {{Langevin}} sampling.
\newblock November 2021.

\bibitem{duncanUsingPerturbedUnderdamped2017}
A.~B. Duncan, N.~N{\"u}sken, and G.~A. Pavliotis.
\newblock Using {{Perturbed Underdamped Langevin Dynamics}} to {{Efficiently
  Sample}} from {{Probability Distributions}}.
\newblock {\em Journal of Statistical Physics}, 169(6):1098--1131, December
  2017.

\bibitem{mattinglyErgodicitySDEsApproximations2002}
J.~C. Mattingly, A.~M. Stuart, and D.~J. Higham.
\newblock Ergodicity for {{SDEs}} and approximations: Locally {{Lipschitz}}
  vector fields and degenerate noise.
\newblock {\em Stochastic Processes and their Applications}, 101(2):185--232,
  October 2002.

\bibitem{katsoulakisMeasuringIrreversibilityNumerical2014}
Markos Katsoulakis, Yannis Pantazis, and Luc {Rey-Bellet}.
\newblock {Measuring the Irreversibility of Numerical Schemes for Reversible
  Stochastic Differential Equations}.
\newblock {\em ESAIM: Mathematical Modelling and Numerical Analysis -
  Mod\'elisation Math\'ematique et Analyse Num\'erique}, 48(5):1351--1379,
  2014.

\bibitem{mattinglyConvergenceNumericalTimeAveraging2010}
Jonathan~C. Mattingly, Andrew~M. Stuart, and M.~V. Tretyakov.
\newblock Convergence of {{Numerical Time-Averaging}} and {{Stationary
  Measures}} via {{Poisson Equations}}.
\newblock {\em SIAM Journal on Numerical Analysis}, 48(2):552--577, January
  2010.

\bibitem{radivojevic2018multi}
T.~Radivojevi{\'c}, M.~Fern{\'a}ndez-Pend{\'a}s, J.~M. Sanz-Serna, and
  E.~Akhmatskaya.
\newblock Multi-stage splitting integrators for sampling with modified
  hamiltonian monte carlo methods.
\newblock {\em Journal of Computational Physics}, 373:900--916, 2018.

\bibitem{neal1992bayesian}
R.~M. Neal.
\newblock Bayesian training of backpropagation networks by the hybrid monte
  carlo method.
\newblock Technical report, Citeseer, 1992.

\bibitem{cances2007theoretical}
Eric Cances, Fr{\'e}d{\'e}ric Legoll, and Gabriel Stoltz.
\newblock Theoretical and numerical comparison of some sampling methods for
  molecular dynamics.
\newblock {\em ESAIM: Mathematical Modelling and Numerical Analysis},
  41(2):351--389, 2007.

\bibitem{fang2014compressible}
Youhan Fang, Jesus-Maria Sanz-Serna, and Robert~D Skeel.
\newblock Compressible generalized hybrid monte carlo.
\newblock {\em The Journal of chemical physics}, 140(17):174108, 2014.

\bibitem{barp2020bracket}
A.~Barp.
\newblock The bracket geometry of statistics.
\newblock 2020.

\bibitem{weinstein1997modular}
A.~Weinstein.
\newblock The modular automorphism group of a poisson manifold.
\newblock {\em Journal of Geometry and Physics}, 23(3-4):379--394, 1997.

\bibitem{holm1998euler}
D.~D. Holm, J.~E. Marsden, and T.~S. Ratiu.
\newblock The euler--poincar{\'e} equations and semidirect products with
  applications to continuum theories.
\newblock {\em Advances in Mathematics}, 137(1):1--81, 1998.

\bibitem{modin2010geodesics}
K.~Modin, M.~Perlmutter, S.~Marsland, and R.~McLachlan.
\newblock Geodesics on lie groups: Euler equations and totally geodesic
  subgroup.
\newblock 2010.

\bibitem{barp2019hamiltonianB}
A.~Barp.
\newblock Hamiltonian monte carlo on lie groups and constrained mechanics on
  homogeneous manifolds.
\newblock In {\em International Conference on Geometric Science of
  Information}, pages 665--675. Springer, 2019.

\bibitem{holbrook2018geodesic}
A.~Holbrook, S.~Lan, A.~Vandenberg-Rodes, and B.~Shahbaba.
\newblock Geodesic lagrangian monte carlo over the space of positive definite
  matrices: with application to bayesian spectral density estimation.
\newblock {\em Journal of statistical computation and simulation},
  88(5):982--1002, 2018.

\bibitem{holbrook2016bayesian}
A.~Holbrook, A.~Vandenberg-Rodes, and B.~Shahbaba.
\newblock Bayesian inference on matrix manifolds for linear dimensionality
  reduction.
\newblock {\em arXiv preprint arXiv:1606.04478}, 2016.

\bibitem{lelievre2019hybrid}
T.~Leli{\`e}vre, M.~Rousset, and G.~Stoltz.
\newblock Hybrid {M}onte {C}arlo methods for sampling probability measures on
  submanifolds.
\newblock {\em Numerische Mathematik}, 143(2):379--421, 2019.

\bibitem{lelievre2020multiple}
T.~Leli{\`e}vre, G.~Stoltz, and W.~Zhang.
\newblock Multiple projection mcmc algorithms on submanifolds.
\newblock {\em arXiv preprint arXiv:2003.09402}, 2020.

\bibitem{graham2019manifold}
M.~M. Graham, A.~H. Thiery, and A.~Beskos.
\newblock Manifold markov chain monte carlo methods for bayesian inference in a
  wide class of diffusion models.
\newblock {\em arXiv preprint arXiv:1912.02982}, 2019.

\bibitem{au2020manifold}
K.~X. Au, M.~M. Graham, and A.~H. Thiery.
\newblock Manifold lifting: scaling mcmc to the vanishing noise regime.
\newblock {\em arXiv preprint arXiv:2003.03950}, 2020.

\bibitem{livingstone2014information}
S.~Livingstone and M.~Girolami.
\newblock Information-geometric markov chain monte carlo methods using
  diffusions.
\newblock {\em Entropy}, 16(6):3074--3102, 2014.

\bibitem{tao2016explicit}
M.~Tao.
\newblock Explicit symplectic approximation of nonseparable hamiltonians:
  Algorithm and long time performance.
\newblock {\em Physical Review E}, 94(4):043303, 2016.

\bibitem{cobb2019introducing}
.~D. Cobb, A.~G. Baydin, A.~Markham, and S.~J. Roberts.
\newblock Introducing an explicit symplectic integration scheme for riemannian
  manifold hamiltonian monte carlo.
\newblock {\em arXiv preprint arXiv:1910.06243}, 2019.

\bibitem{predescu2012computationally}
C.~Predescu, R.~A. Lippert, M.~P. Eastwood, D.~Ierardi, H.~Xu, M.~Jensen, K.~J.
  Bowers, J.~Gullingsrud, C.~A. Rendleman, R.~O. Dror, et~al.
\newblock Computationally efficient molecular dynamics integrators with
  improved sampling accuracy.
\newblock {\em Molecular Physics}, 110(9-10):967--983, 2012.

\bibitem{blanes2014numerical}
S.~Blanes, F.~Casas, and J.~M. Sanz-Serna.
\newblock Numerical integrators for the hybrid monte carlo method.
\newblock {\em SIAM Journal on Scientific Computing}, 36(4):A1556--A1580, 2014.

\bibitem{fernandez2016adaptive}
M.~Fern{\'a}ndez-Pend{\'a}s, E.~Akhmatskaya, and J.~M. Sanz-Serna.
\newblock Adaptive multi-stage integrators for optimal energy conservation in
  molecular simulations.
\newblock {\em Journal of Computational Physics}, 327:434--449, 2016.

\bibitem{campos2017palindromic}
C.~M. Campos and J.~M. Sanz-Serna.
\newblock Palindromic 3-stage splitting integrators, a roadmap.
\newblock {\em Journal of Computational Physics}, 346:340--355, 2017.

\bibitem{bou2018geometric}
N.~Bou-Rabee and J.~M. Sanz-Serna.
\newblock Geometric integrators and the hamiltonian monte carlo method.
\newblock {\em Acta Numerica}, 27:113--206, 2018.

\bibitem{clark2011improving}
M.~A. Clark, B.~Jo{\'o}, A.~D. Kennedy, and P.~J. Silva.
\newblock Improving dynamical lattice qcd simulations through integrator tuning
  using poisson brackets and a force-gradient integrator.
\newblock {\em Physical Review D}, 84(7):071502, 2011.

\bibitem{tuckerman1992reversible}
M.B.B.J.M. Tuckerman, B.~J. Berne, and G.~J. Martyna.
\newblock Reversible multiple time scale molecular dynamics.
\newblock {\em The Journal of chemical physics}, 97(3):1990--2001, 1992.

\bibitem{sexton1992hamiltonian}
J.C. Sexton and D.H. Weingarten.
\newblock Hamiltonian evolution for the hybrid monte carlo algorithm.
\newblock {\em Nuclear Physics B}, 380(3):665--677, 1992.

\bibitem{shahbaba2014split}
B.~Shahbaba, S.~Lan, W.~O. Johnson, and R.~M. Neal.
\newblock Split hamiltonian monte carlo.
\newblock {\em Statistics and Computing}, 24(3):339--349, 2014.

\bibitem{mackenze1989improved}
P.~B. Mackenze.
\newblock An improved hybrid monte carlo method.
\newblock {\em Physics Letters B}, 226(3-4):369--371, 1989.

\bibitem{betancourt2016identifying}
M.~Betancourt.
\newblock Identifying the optimal integration time in hamiltonian monte carlo.
\newblock {\em arXiv preprint arXiv:1601.00225}, 2016.

\bibitem{wang2013adaptive}
Z.~Wang, S.~Mohamed, and N.~Freitas.
\newblock Adaptive hamiltonian and riemann manifold monte carlo.
\newblock In {\em International conference on machine learning}, pages
  1462--1470. PMLR, 2013.

\bibitem{durmus2017convergence}
A.~Durmus, E.~Moulines, and E.~Saksman.
\newblock On the convergence of hamiltonian monte carlo.
\newblock {\em arXiv preprint arXiv:1705.00166}, 2017.

\bibitem{campos2015extra}
C.~M. Campos and J.~M. Sanz-Serna.
\newblock Extra chance generalized hybrid monte carlo.
\newblock {\em Journal of Computational Physics}, 281:365--374, 2015.

\bibitem{dick2014}
J.~Sohl-Dickstein, M.~Mudigonda, and M.~DeWeese.
\newblock Hamiltonian monte carlo without detailed balance.
\newblock {\em International Conference on Machine Learning}, pages 719--726,
  2014.

\bibitem{hoffman2014no}
M.~D. Hoffman, A.~Gelman, et~al.
\newblock The no-u-turn sampler: adaptively setting path lengths in hamiltonian
  monte carlo.
\newblock {\em J. Mach. Learn. Res.}, 15(1):1593--1623, 2014.

\bibitem{ottobre2016function}
M.~Ottobre, N.~S. Pillai, F.~J. Pinski, and A.~M. Stuart.
\newblock A function space hmc algorithm with second order langevin diffusion
  limit.
\newblock {\em Bernoulli}, 22(1):60--106, 2016.

\bibitem{heber2020posterior}
F.~Heber, Z.~Trst’anov{\'a}, and B.~Leimkuhler.
\newblock Posterior sampling strategies based on discretized stochastic
  differential equations for machine learning applications.
\newblock {\em Journal of Machine Learning Research}, 21(228):1--33, 2020.

\bibitem{horowitz1991generalized}
A.~M. Horowitz.
\newblock A generalized guided monte carlo algorithm.
\newblock {\em Physics Letters B}, 268(2):247--252, 1991.

\bibitem{izaguirre2004shadow}
J.~A. Izaguirre and S.~S. Hampton.
\newblock Shadow hybrid monte carlo: an efficient propagator in phase space of
  macromolecules.
\newblock {\em Journal of Computational Physics}, 200(2):581--604, 2004.

\bibitem{radivojevic2020modified}
T.~Radivojevi{\'c} and E.~Akhmatskaya.
\newblock Modified hamiltonian monte carlo for bayesian inference.
\newblock {\em Statistics and Computing}, 30(2):377--404, 2020.

\bibitem{strathmann2015gradient}
H.~Strathmann, D.~Sejdinovic, S.~Livingstone, Z.~Szabo, and A.~Gretton.
\newblock Gradient-free hamiltonian monte carlo with efficient kernel
  exponential families.
\newblock {\em arXiv preprint arXiv:1506.02564}, 2015.

\bibitem{zhang2017hamiltonian}
C.~Zhang, B.~Shahbaba, and H.~Zhao.
\newblock Hamiltonian monte carlo acceleration using surrogate functions with
  random bases.
\newblock {\em Statistics and computing}, 27(6):1473--1490, 2017.

\bibitem{chen2014stochastic}
T.~Chen, E.~Fox, and C.~Guestrin.
\newblock Stochastic gradient hamiltonian monte carlo.
\newblock In {\em International conference on machine learning}, pages
  1683--1691. PMLR, 2014.

\bibitem{betancourt2015fundamental}
M.~Betancourt.
\newblock The fundamental incompatibility of scalable hamiltonian monte carlo
  and naive data subsampling.
\newblock In {\em International Conference on Machine Learning}, pages
  533--540. PMLR, 2015.

\bibitem{muller1997integral}
A.~M{\"u}ller.
\newblock Integral probability metrics and their generating classes of
  functions.
\newblock {\em Advances in Applied Probability}, 29(2):429--443, 1997.

\bibitem{aronszajn1950theory}
N.~Aronszajn.
\newblock Theory of reproducing kernels.
\newblock {\em Transactions of the American mathematical society},
  68(3):337--404, 1950.

\bibitem{berlinet2011reproducing}
A.~Berlinet and C.~Thomas-Agnan.
\newblock {\em Reproducing kernel Hilbert spaces in probability and
  statistics}.
\newblock Springer Science \& Business Media, 2011.

\bibitem{steinwart2008support}
I.~Steinwart and A.~Christmann.
\newblock {\em Support vector machines}.
\newblock Springer Science \& Business Media, 2008.

\bibitem{barp2022simonlester}
A.~Barp, C.J. Simon-Gabriel, and L.~Mackey.
\newblock Targeted convergence characteristics of maximum mean discrepancies
  and kernel {S}tein discrepancies.
\newblock {\em In preparation}.

\bibitem{sriperumbudur2010hilbert}
B.~K. Sriperumbudur, A.~Gretton, K.~Fukumizu, B.~Sch{\"o}lkopf, and G.~R.G.
  Lanckriet.
\newblock Hilbert space embeddings and metrics on probability measures.
\newblock {\em The Journal of Machine Learning Research}, 11:1517--1561, 2010.

\bibitem{muandet2016kernel}
K.~Muandet, K.~Fukumizu, B.~Sriperumbudur, and B.~Sch{\"o}lkopf.
\newblock Kernel mean embedding of distributions: A review and beyond.
\newblock {\em arXiv preprint arXiv:1605.09522}, 2016.

\bibitem{schwartz1964sous}
L.~Schwartz.
\newblock Sous-espaces hilbertiens d’espaces vectoriels topologiques et
  noyaux associ{\'e}s (noyaux reproduisants).
\newblock {\em Journal d’analyse math{\'e}matique}, 13(1):115--256, 1964.

\bibitem{simon2018kernel}
Carl-Johann Simon-Gabriel and Bernhard Sch{\"o}lkopf.
\newblock Kernel distribution embeddings: Universal kernels, characteristic
  kernels and kernel metrics on distributions.
\newblock {\em The Journal of Machine Learning Research}, 19(1):1708--1736,
  2018.

\bibitem{sriperumbudur2011universality}
B.~K. Sriperumbudur, K.~Fukumizu, and G.~R.G. Lanckriet.
\newblock Universality, characteristic kernels and rkhs embedding of measures.
\newblock {\em Journal of Machine Learning Research}, 12(7), 2011.

\bibitem{carmeli2010vector}
C.~Carmeli, E.~De~Vito, A.~Toigo, and V.~Umanit{\'a}.
\newblock Vector valued reproducing kernel hilbert spaces and universality.
\newblock {\em Analysis and Applications}, 8(01):19--61, 2010.

\bibitem{simon2020metrizing}
C.J. Simon-Gabriel, A.~Barp, B.~Sch{\"o}lkopf, and L.~Mackey.
\newblock Metrizing weak convergence with maximum mean discrepancies.
\newblock {\em arXiv preprint arXiv:2006.09268}, 2020.

\bibitem{ethier2009markov}
Stewart~N Ethier and Thomas~G Kurtz.
\newblock {\em Markov processes: characterization and convergence}, volume 282.
\newblock John Wiley \& Sons, 2009.

\bibitem{gorham2017measuring}
Jackson Gorham and Lester Mackey.
\newblock Measuring sample quality with kernels.
\newblock In {\em International Conference on Machine Learning}, pages
  1292--1301. PMLR, 2017.

\bibitem{stein1972bound}
C.~Stein.
\newblock A bound for the error in the normal approximation to the distribution
  of a sum of dependent random variables.
\newblock In {\em Proceedings of the sixth Berkeley symposium on mathematical
  statistics and probability, volume 2: Probability theory}, volume~6, pages
  583--603. University of California Press, 1972.

\bibitem{anastasiou2021stein}
A.~Anastasiou, A.~Barp, F.~Briol, B.~Ebner, R.~E. Gaunt, F.~Ghaderinezhad,
  J.~Gorham, A.~Gretton, C.~Ley, Q.~Liu, et~al.
\newblock Stein's method meets statistics: A review of some recent
  developments.
\newblock {\em arXiv preprint arXiv:2105.03481}, 2021.

\bibitem{lee2013smooth}
J.~M. Lee.
\newblock Smooth manifolds.
\newblock In {\em Introduction to Smooth Manifolds}, pages 1--31. Springer,
  2013.

\bibitem{liu2018riemannian}
C.~Liu and J.~Zhu.
\newblock Riemannian stein variational gradient descent for bayesian inference.
\newblock In {\em Proceedings of the AAAI Conference on Artificial
  Intelligence}, volume~32, 2018.

\bibitem{hodgkinson2020reproducing}
L.~Hodgkinson, R.~Salomone, and F.~Roosta.
\newblock The reproducing stein kernel approach for post-hoc corrected
  sampling.
\newblock {\em arXiv preprint arXiv:2001.09266}, 2020.

\bibitem{barbour1988stein}
A.~D. Barbour.
\newblock Stein's method and poisson process convergence.
\newblock {\em Journal of Applied Probability}, 25(A):175--184, 1988.

\bibitem{gorham2019measuring}
J.~Gorham, A.~B. Duncan, S.~J Vollmer, and L.~Mackey.
\newblock Measuring sample quality with diffusions.
\newblock {\em The Annals of Applied Probability}, 29(5):2884--2928, 2019.

\bibitem{liu2016stein}
Q.~Liu and D.~Wang.
\newblock Stein variational gradient descent: A general purpose bayesian
  inference algorithm.
\newblock {\em Advances in neural information processing systems}, 29, 2016.

\bibitem{oates2017control}
C.~J. Oates, M.~Girolami, and N.~Chopin.
\newblock Control functionals for monte carlo integration.
\newblock {\em Journal of the Royal Statistical Society: Series B (Statistical
  Methodology)}, 79(3):695--718, 2017.

\bibitem{NEURIPS2019_ba7609ee}
A.~Barp, F.~Briol, .~Duncan, M.~Girolami, and L.~Mackey.
\newblock Minimum stein discrepancy estimators.
\newblock In H.~Wallach, H.~Larochelle, A.~Beygelzimer, F.~d\textquotesingle
  Alch\'{e}-Buc, E.~Fox, and R.~Garnett, editors, {\em Advances in Neural
  Information Processing Systems}, volume~32. Curran Associates, Inc., 2019.

\bibitem{chen2019stein}
W.~Y. Chen, A.~Barp, F.~Briol, J.~Gorham, M.~Girolami, L~Mackey, and C.~Oates.
\newblock Stein point markov chain monte carlo.
\newblock In {\em International Conference on Machine Learning}, pages
  1011--1021. PMLR, 2019.

\bibitem{chwialkowski2016kernel}
K.~Chwialkowski, H.~Strathmann, and A.~Gretton.
\newblock A kernel test of goodness of fit.
\newblock In {\em International conference on machine learning}, pages
  2606--2615. PMLR, 2016.

\bibitem{liu2016kernelized}
Q.~Liu, J.~Lee, and M.~Jordan.
\newblock A kernelized stein discrepancy for goodness-of-fit tests.
\newblock In {\em International conference on machine learning}, pages
  276--284. PMLR, 2016.

\bibitem{gretton2012kernel}
A.~Gretton, K.~M. Borgwardt, M.~J. Rasch, B.~Sch{\"o}lkopf, and A.~Smola.
\newblock A kernel two-sample test.
\newblock {\em The Journal of Machine Learning Research}, 13(1):723--773, 2012.

\bibitem{dziugaite2015training}
G.~K. Dziugaite, D.~M. Roy, and Z.~Ghahramani.
\newblock Training generative neural networks via maximum mean discrepancy
  optimization.
\newblock {\em arXiv preprint arXiv:1505.03906}, 2015.

\bibitem{park2000adaptive}
H.~Park, S.~Amari, and K.~Fukumizu.
\newblock Adaptive natural gradient learning algorithms for various stochastic
  models.
\newblock {\em Neural Networks}, 13(7):755--764, 2000.

\bibitem{chen2018natural}
Y.~Chen and W.~Li.
\newblock Natural gradient in wasserstein statistical manifold.
\newblock {\em arXiv preprint arXiv:1805.08380}, 2018.

\bibitem{karakida2016adaptive}
R.~Karakida, M.~Okada, and S.~Amari.
\newblock Adaptive natural gradient learning algorithms for unnormalized
  statistical models.
\newblock In {\em International Conference on Artificial Neural Networks},
  pages 427--434. Springer, 2016.

\bibitem{kakade2001natural}
S.~M. Kakade.
\newblock A natural policy gradient.
\newblock {\em Advances in neural information processing systems}, 14, 2001.

\bibitem{bonnabel2013stochastic}
S.~Bonnabel.
\newblock Stochastic gradient descent on riemannian manifolds.
\newblock {\em IEEE Transactions on Automatic Control}, 58(9):2217--2229, 2013.

\bibitem{leok2017connecting}
M.~Leok and J.~Zhang.
\newblock Connecting information geometry and geometric mechanics.
\newblock {\em Entropy}, 19(10):518, 2017.

\bibitem{briol2019statistical}
F.~Briol, A.~Barp, A.~B. Duncan, and M.~Girolami.
\newblock Statistical inference for generative models with maximum mean
  discrepancy.
\newblock 2019.
\newblock arXiv:1906.05944.

\bibitem{gretton2009fast}
A.~Gretton, K.~Fukumizu, Z.~Harchaoui, and B.~K. Sriperumbudur.
\newblock A fast, consistent kernel two-sample test.
\newblock In {\em NIPS}, volume~23, pages 673--681, 2009.

\bibitem{garreau2017large}
Damien Garreau, Wittawat Jitkrittum, and Motonobu Kanagawa.
\newblock Large sample analysis of the median heuristic.
\newblock {\em arXiv preprint arXiv:1707.07269}, 2017.

\bibitem{sutherland2016generative}
Danica~J Sutherland, Hsiao-Yu Tung, Heiko Strathmann, Soumyajit De, Aaditya
  Ramdas, Alex Smola, and Arthur Gretton.
\newblock Generative models and model criticism via optimized maximum mean
  discrepancy.
\newblock {\em arXiv preprint arXiv:1611.04488}, 2016.

\bibitem{ramdas2015adaptivity}
Aaditya Ramdas, Sashank~J Reddi, Barnabas Poczos, Aarti Singh, and Larry
  Wasserman.
\newblock Adaptivity and computation-statistics tradeoffs for kernel and
  distance based high dimensional two sample testing.
\newblock {\em arXiv preprint arXiv:1508.00655}, 2015.

\bibitem{li2017mmd}
Chun-Liang Li, Wei-Cheng Chang, Yu~Cheng, Yiming Yang, and Barnab{\'a}s
  P{\'o}czos.
\newblock Mmd gan: Towards deeper understanding of moment matching network.
\newblock {\em arXiv preprint arXiv:1705.08584}, 2017.

\bibitem{fristonFreeEnergyPrinciple2006}
K.~Friston, J.~Kilner, and L.~Harrison.
\newblock A free energy principle for the brain.
\newblock {\em J. Physiology-Paris}, 100(1-3):70--87, 2006.

\bibitem{parrComputationalNeurologyActive2019}
Thomas Parr.
\newblock {\em The Computational Neurology of Active Vision}.
\newblock PhD thesis, University College London, {London}, 2019.

\bibitem{fountasDeepActiveInference2020}
Zafeirios Fountas, Noor Sajid, Pedro A.~M. Mediano, and Karl Friston.
\newblock Deep active inference agents using {{Monte-Carlo}} methods.
\newblock {\em arXiv:2006.04176 [cs, q-bio, stat]}, June 2020.

\bibitem{fristonStochasticChaosMarkov2021}
Karl Friston, Conor Heins, Kai Ueltzh{\"o}ffer, Lancelot Da~Costa, and Thomas
  Parr.
\newblock Stochastic {{Chaos}} and {{Markov Blankets}}.
\newblock {\em Entropy}, 23(9):1220, September 2021.

\bibitem{fristonFreeEnergyPrinciple2022}
Karl Friston, Lancelot Da~Costa, Noor Sajid, Conor Heins, Kai Ueltzh{\"o}ffer,
  Grigorios~A. Pavliotis, and Thomas Parr.
\newblock The free energy principle made simpler but not too simple.
\newblock {\em arXiv:2201.06387 [cond-mat, physics:nlin, physics:physics,
  q-bio]}, January 2022.

\bibitem{dacostaBayesianMechanicsStationary2021a}
Lancelot Da~Costa, Karl Friston, Conor Heins, and Grigorios~A. Pavliotis.
\newblock Bayesian mechanics for stationary processes.
\newblock {\em Proceedings of the Royal Society A: Mathematical, Physical and
  Engineering Sciences}, 477(2256):20210518, December 2021.

\bibitem{fristonSophisticatedInference2021}
Karl Friston, Lancelot Da~Costa, Danijar Hafner, Casper Hesp, and Thomas Parr.
\newblock Sophisticated {{Inference}}.
\newblock {\em Neural Computation}, 33(3):713--763, February 2021.

\bibitem{parrMemoryMarkovBlankets2021}
Thomas Parr, Lancelot Da~Costa, Conor Heins, Maxwell James~D. Ramstead, and
  Karl~J. Friston.
\newblock Memory and {{Markov Blankets}}.
\newblock {\em Entropy}, 23(9):1105, September 2021.

\bibitem{tenka}
Samuel Tenka.
\newblock personal communication.

\bibitem{schwartenbeckComputationalMechanismsCuriosity2019}
Philipp Schwartenbeck, Johannes Passecker, Tobias~U Hauser, Thomas~HB
  FitzGerald, Martin Kronbichler, and Karl~J Friston.
\newblock Computational mechanisms of curiosity and goal-directed exploration.
\newblock {\em eLife}, page~45, 2019.

\bibitem{bleiVariationalInferenceReview2017}
David~M. Blei, Alp Kucukelbir, and Jon~D. McAuliffe.
\newblock Variational {{Inference}}: {{A Review}} for {{Statisticians}}.
\newblock {\em Journal of the American Statistical Association},
  112(518):859--877, April 2017.

\bibitem{kahnemanProspectTheoryAnalysis1979}
Daniel Kahneman and Amos Tversky.
\newblock Prospect {{Theory}}: {{An Analysis}} of {{Decision}} under {{Risk}}.
\newblock {\em Econometrica}, 47(2):263--291, 1979.

\bibitem{levineReinforcementLearningControl2018}
Sergey Levine.
\newblock Reinforcement {{Learning}} and {{Control}} as {{Probabilistic
  Inference}}: {{Tutorial}} and {{Review}}.
\newblock {\em arXiv:1805.00909 [cs, stat]}, May 2018.

\bibitem{rawlikStochasticOptimalControl2013}
Konrad Rawlik, Marc Toussaint, and Sethu Vijayakumar.
\newblock On {{Stochastic Optimal Control}} and {{Reinforcement Learning}} by
  {{Approximate Inference}}.
\newblock In {\em Twenty-{{Third International Joint Conference}} on
  {{Artificial Intelligence}}}, June 2013.

\bibitem{toussaintRobotTrajectoryOptimization2009}
Marc Toussaint.
\newblock Robot trajectory optimization using approximate inference.
\newblock In {\em Proceedings of the 26th {{Annual International Conference}}
  on {{Machine Learning}}}, {{ICML}} '09, pages 1049--1056, {Montreal, Quebec,
  Canada}, June 2009. {Association for Computing Machinery}.

\bibitem{kalmanNewApproachLinear1960}
R.~E. Kalman.
\newblock A {{New Approach}} to {{Linear Filtering}} and {{Prediction
  Problems}}.
\newblock {\em Journal of Basic Engineering}, 82(1):35--45, March 1960.

\bibitem{todorovGeneralDualityOptimal2008a}
Emanuel Todorov.
\newblock General duality between optimal control and estimation.
\newblock In {\em 2008 47th {{IEEE Conference}} on {{Decision}} and
  {{Control}}}, pages 4286--4292, December 2008.

\bibitem{kappenOptimalControlGraphical2012}
Hilbert~J. Kappen, Vicen{\c c} G{\'o}mez, and Manfred Opper.
\newblock Optimal control as a graphical model inference problem.
\newblock {\em Machine Learning}, 87(2):159--182, May 2012.

\bibitem{ziebartModelingPurposefulAdaptive2010}
B.~Ziebart.
\newblock {\em Modeling Purposeful Adaptive Behavior with the Principle of
  Maximum Causal Entropy.}
\newblock PhD thesis, Carnegie Mellon University, {Pittsburgh}, 2010.

\bibitem{jaynesInformationTheoryStatistical1957}
E.~T. Jaynes.
\newblock Information {{Theory}} and {{Statistical Mechanics}}.
\newblock {\em Physical Review}, 106(4):620--630, May 1957.

\bibitem{lasotaChaosFractalsNoise1994}
Andrzej Lasota and Michael~C. MacKey.
\newblock {\em Chaos, {{Fractals}}, and {{Noise}}: {{Stochastic Aspects}} of
  {{Dynamics}}}.
\newblock {Springer-Verlag}, 1994.

\bibitem{mackayInformationTheoryInference2003}
David J.~C. MacKay.
\newblock {\em Information {{Theory}}, {{Inference}} and {{Learning
  Algorithms}}}.
\newblock {Cambridge University Press}, {Cambridge, UK ; New York}, sixth
  printing 2007 edition edition, September 2003.

\bibitem{lindleyMeasureInformationProvided1956}
D.~V. Lindley.
\newblock On a {{Measure}} of the {{Information Provided}} by an
  {{Experiment}}.
\newblock {\em The Annals of Mathematical Statistics}, 27(4):986--1005, 1956.

\bibitem{mackayInformationBasedObjectiveFunctions1992}
David J.~C. MacKay.
\newblock Information-{{Based Objective Functions}} for {{Active Data
  Selection}}.
\newblock {\em Neural Computation}, 4(4):590--604, July 1992.

\bibitem{oudeyerWhatIntrinsicMotivation2007}
Pierre-Yves Oudeyer and Frederic Kaplan.
\newblock What is {{Intrinsic Motivation}}? {{A Typology}} of {{Computational
  Approaches}}.
\newblock {\em Frontiers in Neurorobotics}, 1:6, November 2007.

\bibitem{schmidhuberFormalTheoryCreativity2010}
J{\"u}rgen Schmidhuber.
\newblock Formal {{Theory}} of {{Creativity}}, {{Fun}}, and {{Intrinsic
  Motivation}} (1990\textendash 2010).
\newblock {\em IEEE Transactions on Autonomous Mental Development},
  2(3):230--247, September 2010.

\bibitem{bartoNoveltySurprise2013}
A.~Barto, M.~Mirolli, and G.~Baldassarre.
\newblock Novelty or {{Surprise}}?
\newblock {\em Frontiers in Psychology}, 4, 2013.

\bibitem{sunPlanningBeSurprised2011}
Yi~Sun, Faustino Gomez, and Juergen Schmidhuber.
\newblock Planning to {{Be Surprised}}: {{Optimal Bayesian Exploration}} in
  {{Dynamic Environments}}.
\newblock {\em arXiv:1103.5708 [cs, stat]}, March 2011.

\bibitem{deciIntrinsicMotivationSelfDetermination1985}
Edward Deci and Richard~M. Ryan.
\newblock {\em Intrinsic {{Motivation}} and {{Self-Determination}} in {{Human
  Behavior}}}.
\newblock Perspectives in {{Social Psychology}}. {Springer US}, {New York},
  1985.

\bibitem{ittiBayesianSurpriseAttracts2009}
Laurent Itti and Pierre Baldi.
\newblock Bayesian surprise attracts human attention.
\newblock {\em Vision research}, 49(10):1295--1306, May 2009.

\bibitem{parrGenerativeModelsActive2021}
Thomas Parr, Noor Sajid, Lancelot Da~Costa, M.~Berk Mirza, and Karl~J. Friston.
\newblock Generative {{Models}} for {{Active Vision}}.
\newblock {\em Frontiers in Neurorobotics}, 15, 2021.

\bibitem{barlowPossiblePrinciplesUnderlying1961}
H.~B. Barlow.
\newblock {\em Possible {{Principles Underlying}} the {{Transformations}} of
  {{Sensory Messages}}}.
\newblock {The MIT Press}, 1961.

\bibitem{linskerPerceptualNeuralOrganization1990}
R~Linsker.
\newblock Perceptual {{Neural Organization}}: {{Some Approaches Based}} on
  {{Network Models}} and {{Information Theory}}.
\newblock {\em Annual Review of Neuroscience}, 13(1):257--281, 1990.

\bibitem{opticanTemporalEncodingTwodimensional1987a}
L.~M. Optican and B.~J. Richmond.
\newblock Temporal encoding of two-dimensional patterns by single units in
  primate inferior temporal cortex. {{III}}. {{Information}} theoretic
  analysis.
\newblock {\em Journal of Neurophysiology}, 57(1):162--178, January 1987.

\bibitem{bellmanDynamicProgramming1957}
Richard~E. Bellman.
\newblock {\em Dynamic {{Programming}}}.
\newblock {Princeton University Press}, {Princeton, NJ, US}, 1957.

\bibitem{astromOptimalControlMarkov1965a}
K.~J {\AA}str{\"o}m.
\newblock Optimal control of {{Markov}} processes with incomplete state
  information.
\newblock {\em Journal of Mathematical Analysis and Applications},
  10(1):174--205, February 1965.

\bibitem{kaplanConductInquiry1973}
Abraham Kaplan.
\newblock {\em The Conduct of Inquiry}.
\newblock {Transaction Publishers}, 1973.

\bibitem{sajidActiveInferenceBayesian2021}
Noor Sajid, Lancelot Da~Costa, Thomas Parr, and Karl Friston.
\newblock Active inference, {{Bayesian}} optimal design, and expected utility.
\newblock {\em arXiv:2110.04074 [cs, math, stat]}, September 2021.

\bibitem{berger-talExplorationExploitationDilemmaMultidisciplinary2014}
Oded {Berger-Tal}, Jonathan Nathan, Ehud Meron, and David Saltz.
\newblock The {{Exploration-Exploitation Dilemma}}: {{A Multidisciplinary
  Framework}}.
\newblock {\em PLOS ONE}, 9(4):e95693, April 2014.

\bibitem{heinsPymdpPythonLibrary2022}
Conor Heins, Beren Millidge, Daphne Demekas, Brennan Klein, Karl Friston, Iain
  Couzin, and Alexander Tschantz.
\newblock Pymdp: {{A Python}} library for active inference in discrete state
  spaces.
\newblock {\em arXiv:2201.03904 [cs, q-bio]}, January 2022.

\bibitem{smithStepbyStepTutorialActive2022}
Ryan Smith, Karl~J. Friston, and Christopher~J. Whyte.
\newblock A step-by-step tutorial on active inference and its application to
  empirical data.
\newblock {\em Journal of Mathematical Psychology}, 107:102632, April 2022.

\bibitem{fristonActiveInferenceCuriosity2017}
Karl~J. Friston, Marco Lin, Christopher~D. Frith, Giovanni Pezzulo, J.~Allan
  Hobson, and Sasha Ondobaka.
\newblock Active {{Inference}}, {{Curiosity}} and {{Insight}}.
\newblock {\em Neural Computation}, 29(10):2633--2683, October 2017.

\bibitem{catalRobotNavigationHierarchical2021}
Ozan {\c C}atal, Tim Verbelen, Toon {Van de Maele}, Bart Dhoedt, and Adam
  Safron.
\newblock Robot navigation as hierarchical active inference.
\newblock {\em Neural Networks}, 142:192--204, October 2021.

\bibitem{fristonDeepTemporalModels2018}
Karl~J. Friston, Richard Rosch, Thomas Parr, Cathy Price, and Howard Bowman.
\newblock Deep temporal models and active inference.
\newblock {\em Neuroscience \& Biobehavioral Reviews}, 90:486--501, July 2018.

\bibitem{bishopPatternRecognitionMachine2006}
Christopher~M. Bishop.
\newblock {\em Pattern Recognition and Machine Learning}.
\newblock Information Science and Statistics. {Springer}, {New York}, 2006.

\bibitem{fristonActiveInferenceProcess2017}
Karl Friston, Thomas FitzGerald, Francesco Rigoli, Philipp Schwartenbeck, and
  Giovanni Pezzulo.
\newblock Active {{Inference}}: {{A Process Theory}}.
\newblock {\em Neural Computation}, 29(1):1--49, January 2017.

\bibitem{fristonActiveInferenceLearning2016}
Karl Friston, Thomas FitzGerald, Francesco Rigoli, Philipp Schwartenbeck, John
  O'Doherty, and Giovanni Pezzulo.
\newblock Active inference and learning.
\newblock {\em Neuroscience \& Biobehavioral Reviews}, 68:862--879, September
  2016.

\bibitem{smithActiveInferenceApproach2020}
Ryan Smith, Philipp Schwartenbeck, Thomas Parr, and Karl~J. Friston.
\newblock An {{Active Inference Approach}} to {{Modeling Structure Learning}}:
  {{Concept Learning}} as an {{Example Case}}.
\newblock {\em Frontiers in Computational Neuroscience}, 14, May 2020.

\bibitem{fristonWorldModelLearning2021}
Karl Friston, Rosalyn~J. Moran, Yukie Nagai, Tadahiro Taniguchi, Hiroaki Gomi,
  and Josh Tenenbaum.
\newblock World model learning and inference.
\newblock {\em Neural Networks}, 144:573--590, December 2021.

\bibitem{wauthierSleepModelReduction2020}
Samuel~T Wauthier, Ozan {\c C}atal, Tim Verbelen, and Bart Dhoedt.
\newblock Sleep: {{Model Reduction}} in {{Deep Active Inference}}.
\newblock page~13, 2020.

\bibitem{tschantzLearningActionorientedModels2020}
Alexander Tschantz, Anil~K. Seth, and Christopher~L. Buckley.
\newblock Learning action-oriented models through active inference.
\newblock {\em PLOS Computational Biology}, 16(4):e1007805, April 2020.

\bibitem{fristonBayesianModelReduction2019}
Karl Friston, Thomas Parr, and Peter Zeidman.
\newblock Bayesian model reduction.
\newblock {\em arXiv:1805.07092 [stat]}, October 2019.

\bibitem{championBranchingTimeActive2021}
Th{\'e}ophile Champion, Howard Bowman, and Marek Grze{\'s}.
\newblock Branching {{Time Active Inference}}: Empirical study and complexity
  class analysis.
\newblock {\em arXiv:2111.11276 [cs]}, November 2021.

\bibitem{championBranchingTimeActive2021a}
Th{\'e}ophile Champion, Lancelot Da~Costa, Howard Bowman, and Marek Grze{\'s}.
\newblock Branching {{Time Active Inference}}: The theory and its generality.
\newblock {\em arXiv:2111.11107 [cs]}, November 2021.

\bibitem{maistoActiveTreeSearch2021}
Domenico Maisto, Francesco Gregoretti, Karl Friston, and Giovanni Pezzulo.
\newblock Active {{Tree Search}} in {{Large POMDPs}}.
\newblock {\em arXiv:2103.13860 [cs, math, q-bio]}, March 2021.

\bibitem{dacostaRelationshipDynamicProgramming2020a}
Lancelot Da~Costa, Noor Sajid, Thomas Parr, Karl Friston, and Ryan Smith.
\newblock The relationship between dynamic programming and active inference:
  The discrete, finite-horizon case.
\newblock {\em arXiv:2009.08111 [cs, math, q-bio]}, September 2020.

\bibitem{paulActiveInferenceStochastic}
Aswin Paul, Lancelot Da~Costa, Manoj Gopalkrishnan, and Adeel Razi.
\newblock Active {{Inference}} for {{Stochastic}} and {{Adaptive Control}} in a
  {{Partially Observable Environment}}.

\bibitem{zhangAdvancesVariationalInference2017}
Cheng Zhang, Judith Butepage, Hedvig Kjellstrom, and Stephan Mandt.
\newblock Advances in {{Variational Inference}}.
\newblock {\em arXiv:1711.05597 [cs, stat]}, November 2017.

\bibitem{vandelaarSimulatingActiveInference2019}
Thijs~W. {van de Laar} and Bert {de Vries}.
\newblock Simulating {{Active Inference Processes}} by {{Message Passing}}.
\newblock {\em Frontiers in Robotics and AI}, 6, 2019.

\bibitem{sajidMixedGenerativeModel2022}
Noor Sajid, Emma Holmes, Lancelot~Da Costa, Cathy Price, and Karl Friston.
\newblock A mixed generative model of auditory word repetition, January 2022.

\bibitem{tschantzControlHybridInference2020}
Alexander Tschantz, Beren Millidge, Anil~K. Seth, and Christopher~L. Buckley.
\newblock Control as {{Hybrid Inference}}.
\newblock {\em arXiv:2007.05838 [cs, stat]}, July 2020.

\bibitem{winnVariationalMessagePassing2005}
John Winn and Christopher~M Bishop.
\newblock Variational {{Message Passing}}.
\newblock {\em Journal of Machine Learning Research}, page~34, 2005.

\bibitem{wainwrightGraphicalModelsExponential2007}
M.~J. Wainwright and M.~I. Jordan.
\newblock Graphical {{Models}}, {{Exponential Families}}, and {{Variational
  Inference}}.
\newblock {\em Found. Trends in Mach. Learn.}, 1(1--2):1--305, 2007.

\bibitem{parrNeuronalMessagePassing2019}
Thomas Parr, Dimitrije Markovic, Stefan~J. Kiebel, and Karl~J. Friston.
\newblock Neuronal message passing using {{Mean-field}}, {{Bethe}}, and
  {{Marginal}} approximations.
\newblock {\em Scientific Reports}, 9(1):1889, December 2019.

\bibitem{schwobelActiveInferenceBelief2018}
Sarah Schw{\"o}bel, Stefan Kiebel, and Dimitrije Markovi{\'c}.
\newblock Active {{Inference}}, {{Belief Propagation}}, and the {{Bethe
  Approximation}}.
\newblock {\em Neural Computation}, 30(9):2530--2567, September 2018.

\bibitem{championRealizingActiveInference2021}
Th{\'e}ophile Champion, Marek Grze{\'s}, and Howard Bowman.
\newblock Realizing {{Active Inference}} in {{Variational Message Passing}}:
  {{The Outcome-Blind Certainty Seeker}}.
\newblock {\em Neural Computation}, 33(10):2762--2826, September 2021.

\bibitem{fristonGraphicalBrainBelief2017}
Karl~J. Friston, Thomas Parr, and Bert {de Vries}.
\newblock The graphical brain: {{Belief}} propagation and active inference.
\newblock {\em Network Neuroscience}, 1(4):381--414, December 2017.

\bibitem{parrComputationalNeurologyMovement2021}
Thomas Parr, Jakub Limanowski, Vishal Rawji, and Karl Friston.
\newblock The computational neurology of movement under active inference.
\newblock {\em Brain}, March 2021.

\end{thebibliography}

\end{document}